\pdfoutput=1

\PassOptionsToPackage{table}{xcolor}

\documentclass[11pt]{article}

\usepackage[preprint]{acl}

\usepackage[T1]{fontenc}
\usepackage{xcolor}

\definecolor{boxbg}{HTML}{F8F8F8}
\definecolor{boxframe}{HTML}{2E86AB}
\definecolor{friendly-bg}{HTML}{F0F0F0}
\definecolor{friendly-keyword}{rgb}{0.00,0.50,0.00}
\definecolor{friendly-string}{HTML}{BB4444}
\definecolor{friendly-comment}{HTML}{408090}

\usepackage{newtxtext}
\usepackage{latexsym}
\usepackage{amsmath}
\usepackage{amssymb}
\usepackage{pifont}
\usepackage{xspace}

\usepackage{multirow}
\usepackage{booktabs}

\usepackage{geometry}
\usepackage{setspace}
\usepackage{adjustbox}
\usepackage{cuted}

\usepackage{hyperref}
\usepackage{xurl}
\usepackage{cleveref}

\usepackage{listings}
\lstdefinestyle{pythonstyle}{
    language=Python,
    basicstyle=\scriptsize\ttfamily,
    keywordstyle=\color{friendly-keyword}\bfseries,
    stringstyle=\color{friendly-string},
    commentstyle=\color{friendly-comment}\itshape,
    backgroundcolor=\color{friendly-bg},
    frame=single,
    rulecolor=\color{boxframe},
    framesep=6pt,
    showstringspaces=false,
    tabsize=4,
    breaklines=true,
    breakatwhitespace=false,
    breakindent=0pt,
    postbreak=\mbox{\textcolor{gray}{$\hookrightarrow$}\space},
    columns=flexible,
    keepspaces=true,
}

\usepackage[breakable,most]{tcolorbox}
\usepackage[framemethod=TikZ]{mdframed}
\usepackage{caption}

\usepackage[colorinlistoftodos,textsize=scriptsize]{todonotes}

\usepackage{array}
\newcolumntype{L}[1]{>{\raggedright\arraybackslash}p{#1}}

\newcommand{\dataset}[1]{\textsc{#1}\xspace}
\newcommand{\ourdataset}{\dataset{EngTrace}}

\usepackage[T1]{fontenc}

\usepackage[utf8]{inputenc}

\usepackage{microtype}

\usepackage{inconsolata}

\usepackage{arydshln}

\usepackage{fontawesome}

\usepackage{graphicx}

\usepackage{booktabs}
\usepackage{array}
\usepackage{xcolor}
\usepackage{colortbl}
\usepackage{caption}

\usepackage{makecell}
\usepackage[table]{xcolor}

\usepackage{enumitem}

\definecolor{aclblue}{RGB}{0,85,160}

\title{ \ourdataset: A Symbolic Benchmark \\ for Verifiable Process Evaluation of Engineering Reasoning
}

\author{
\textbf{Ayesha Gull\textsuperscript{1}$^*$} \ 
\textbf{Muhammad Usman Safder\textsuperscript{1}$^*$} \
\textbf{Rania Hossam\textsuperscript{2}} \
\textbf{Fan Zhang\textsuperscript{3}} \\
\textbf{Veselin Stoyanov\textsuperscript{2}} \
\textbf{Preslav Nakov\textsuperscript{2}} \ 
\textbf{Zhuohan Xie\textsuperscript{2}} \
\\
\textsuperscript{1}Namal University
 \textsuperscript{2}MBZUAI \
  \textsuperscript{3}The University of Tokyo \\
 \faGlobe\ \href{https://usmansafdarktk.github.io/EngTrace/}{\textcolor{aclblue}{Project}}
\quad
\faGithub\ \href{https://github.com/usmansafdarktk/EngTrace}{\textcolor{aclblue}{Code}}
}

\begin{document}
\maketitle

\begin{abstract}

Most engineering problems involve complex multi-step derivations through checkable intermediate
quantities.
Grading only the final answer misjudges LLMs, crediting correct answers reached through flawed steps
and penalizing wrong answers built on mostly sound ones.
Yet, existing benchmarks such as \textsc{SuperGPQA} and \textsc{EEE-Bench} score only final answers,
and step-level ones such as \textsc{TPS-CalcBench} rely on LLM judges or human graders, or cover a
single domain.
We therefore introduce \ourdataset, a symbolic benchmark for verifiable process evaluation of
engineering reasoning.
Its 150 expert-certified templates, spanning five branches, 15 domains, and 42 areas, generate each
problem with a gold reasoning trace, so that every intermediate quantity is known exactly.
We design and validate an evaluator that checks final answers, coverage of the gold trace's
intermediate quantities (\textit{milestones}), and displayed arithmetic, mostly without an LLM judge.
Across 11 LLMs, final answers barely separate the strongest models, while their derivations differ.
Wrong answers still state about half of the gold milestones, and domain experts trace most wrong
answers to calculation slips, with conceptual errors rarer on easy problems than on harder ones.
Overall, as LLMs near the final-answer ceiling, checking derivations, as \ourdataset does, reveals
differences that final answers miss.


\end{abstract}

\section{Introduction}
\label{sec:intro}

As LLMs enter specialized domains such as engineering, it becomes essential to evaluate how they
reason, not only what they answer.
Engineering problems are solved through \emph{derivations}, multi-step chains of reasoning that are
governed by physical laws and pass through checkable intermediate quantities to a final answer.
A correct final answer can still follow flawed reasoning, more so on harder problems: in competition
mathematics, as many as half of LLMs' correct answers rest on erroneous
steps~\citep{zheng2024processbench, zhang2025prmlessons}.

\begin{figure}[t]
    \centering
    \includegraphics[width=1\linewidth]{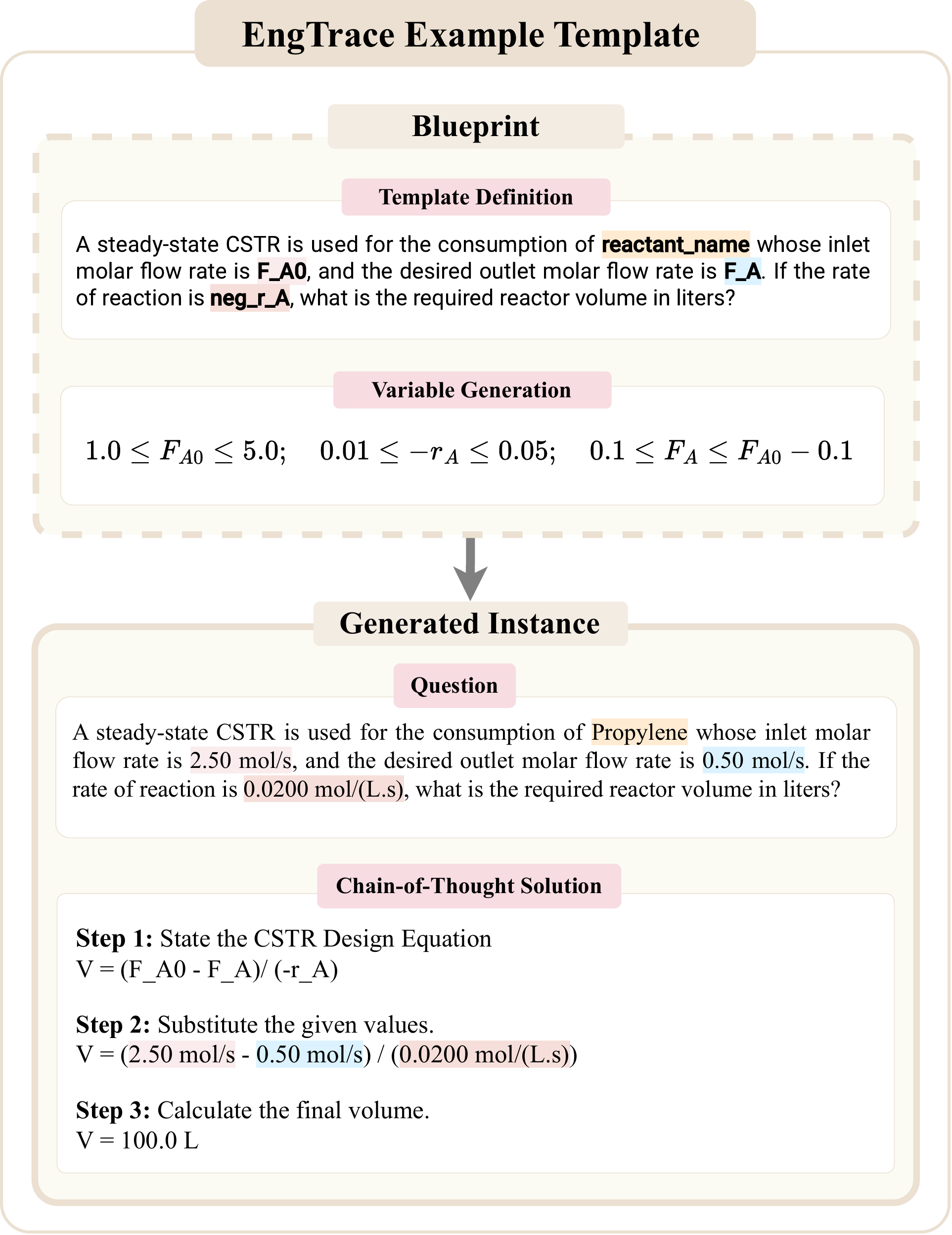}
    \caption{\textbf{An example from \ourdataset: continuous stirred-tank reactor (CSTR) volume
        calculation.}
        A symbolic template generates a unique problem instance paired with a gold reasoning trace
        to enable verifiable process evaluation.}
    \label{fig:engtrace-template-cstr}
\end{figure}

In engineering, the mismatch between answer and derivation runs both ways: LLMs reach correct
answers through flawed reasoning, and sound derivations miss the answer through a minor arithmetic
slip~\citep{syed2024transportbench, zheng2026tpscalcbench}.
Yet, most engineering benchmarks grade only the final
answer~\citep{du2025supergpqa, li2025eeebench, zhou2025engibench}, and those that assess
intermediate steps rely on LLM judges or human graders, or are limited to a single
domain~\citep{syed2024transportbench, zheng2026tpscalcbench, duzkar2026thermoqa}.
What the final answer cannot show is therefore how much of the correct derivation a wrong final
answer keeps, which intermediate quantities it misses, which slips a correct answer hides, and how
models with tied answers differ.

To address these gaps, we introduce \ourdataset, a symbolic benchmark for verifiable process
evaluation of engineering reasoning across five branches.
Following symbolic-template benchmarks in mathematics~\citep{mirzadeh2024gsmsymbolic} and
finance~\citep{xie2025finchain}, \ourdataset represents each problem as a template: Python code
that samples physically grounded parameters and computes the answer together with every
intermediate quantity of its gold reasoning trace (\autoref{fig:engtrace-template-cstr}).
The benchmark covers a depth-oriented subset of engineering, three core domains per branch, and its
problems are examination-style by construction, since a gold trace must be computable.
Our expert-validated evaluator deterministically checks a response's final answer, its coverage of
the gold trace's intermediate quantities (\emph{milestones}), and its displayed arithmetic, leaving
the milestones it cannot match to an LLM judge from outside the evaluated model families.

In summary, our contributions are as follows:
\begin{itemize}
    \item We present and release \ourdataset, whose 150 expert-certified templates across five
          branches generate each problem with a gold reasoning trace.
    \item We design an expert-validated evaluator that deterministically checks final answers,
          milestones, and displayed arithmetic, and leaves to an LLM judge only what these checks
          cannot decide.
    \item We evaluate eleven LLMs and show that final answers barely separate the strongest models,
          while their derivations differ.
          Domain experts trace most wrong answers to calculation slips, with conceptual errors
          rarer on easy problems than on harder ones.
\end{itemize}

\section{Related Work}
\label{sec:related}

\paragraph{Engineering and Science Benchmarks.} Engineering and science benchmarks mostly score the
final answer or the whole
response~\citep{zhou2025engibench, naser2026eri, li2025eeebench, qiu2025phybench}.
Those that examine the reasoning have it rated by domain experts or LLM
judges~\citep{syed2024transportbench, zheng2026tpscalcbench, wasiq2026engvqa}, or check steps
against references on fixed problems~\citep{duzkar2026thermoqa, zhao2026prismphysics}.
\ourdataset evaluates the final answer as well as the derivation behind it against a gold reasoning
trace, rather than relying on a rater.
Because this reference is exact, most checks are deterministic and do not depend on whether a rater
could solve the problem.

\paragraph{Symbolic and Dynamic Benchmarks.} Fixed problems can also leak into training
data~\citep{zhang2024gsm1k}, so symbolic and dynamic benchmarks generate fresh instances from
templates, code, or perturbations~\citep{mirzadeh2024gsmsymbolic, zhu2024dyval, li2024gsmplus}.
Most of these generators, including those for
engineering~\citep{skelic2025circuit, chen2026optengine}, record only the final answer, and the few
that also write out step-by-step solutions do so outside
engineering~\citep{fan2024hardmath, imani2025sympybench, xie2025finchain}.
\ourdataset also uses templates, but each computes the full derivation alongside the problem, so
every intermediate quantity is known exactly.
Moreover, their parameters are physically grounded and realistic.

\paragraph{Step-Level Evaluators.} Step-level evaluation mostly relies on learned or judged scores.
Process reward models (PRMs) learn step correctness from human labels~\citep{lightman2023verify} or
rollout-estimated labels~\citep{wang2023mathshepherd}, which are noisy~\citep{zhang2025prmlessons}.
Trained and tested mostly on mathematics~\citep{zheng2024processbench, song2025prmbench}, they
generalize poorly to other domains~\citep{zeng2025versaprm}.
Reference-free evaluators rate each step's validity, informativeness, or
redundancy~\citep{golovneva2022roscoe, prasad2023receval, xia2024reasoneval}, and benchmarks with
generated solutions align steps or grade solutions with an
LLM~\citep{xie2025finchain, fan2024hardmath}.
LLM judges agree with experts mainly on questions they can answer themselves, unless they are given
a correct reference~\citep{krumdick2026nofreelabels}.
\ourdataset matches each response against the problem's known milestones and recomputes its
displayed arithmetic.
An LLM judge rules on the unmatched milestones, given their correct values, so it never has to
solve the problem itself.

\section{\ourdataset Design and Methodology}
\label{sec:benchmark}

\subsection{Taxonomy and Content Selection}
\label{subsec:taxonomy}

\begin{figure*}[t]
    \centering
    \includegraphics[width=0.85\textwidth]{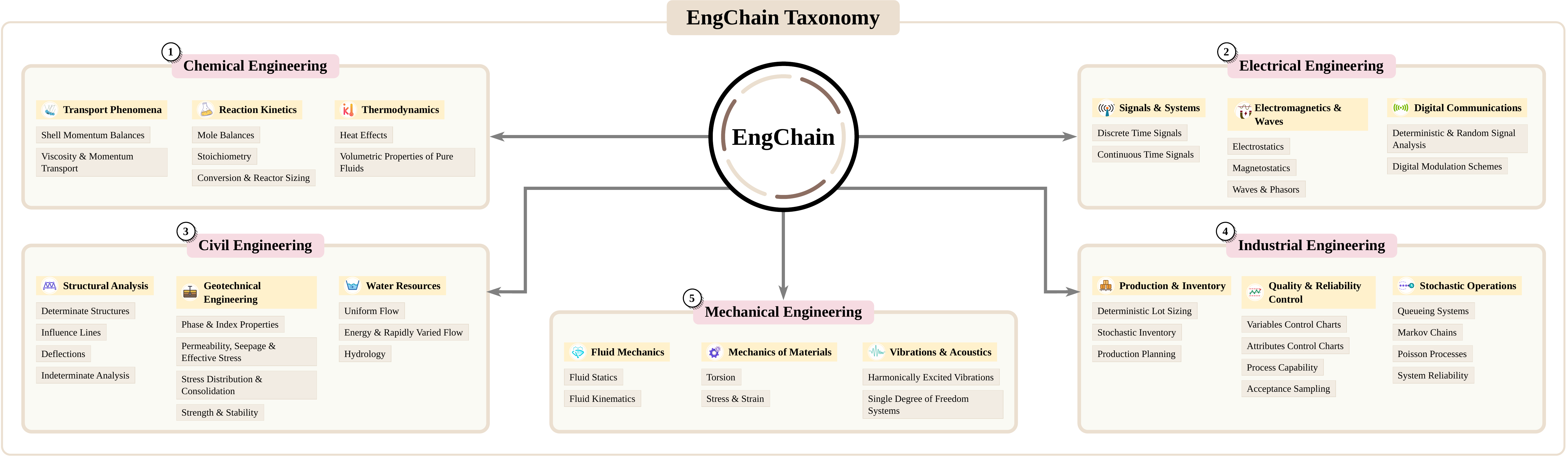}
    \caption{\ourdataset's \textbf{Hierarchical Taxonomy} spanning 15 domains and 42 areas across
    five branches:
    process equilibrium and dynamics (Chemical); field theory and signal transmission (Electrical);
    structures, soils, and water resources (Civil); production, quality, and stochastic operations
    (Industrial); and physical mechanics and material properties (Mechanical).}
    \label{fig:engtrace-overview}
\end{figure*}

The taxonomy of \ourdataset, illustrated in~\autoref{fig:engtrace-overview}, follows formal
engineering curricula.

\paragraph{Domain and Area Selection.}
We drew the candidate domains and areas from the accreditation criteria of
ABET~\citep{abet2024criteria}, the Fundamentals of Engineering exam specifications of
NCEES~\citep{ncees2020fechemical, ncees2020fecivil, ncees2020feelectrical, ncees2020feindustrial, ncees2020femechanical},
the bodies of knowledge of ASCE~\citep{asce2019cebok} and IISE~\citep{iise2021bok}, and the standard
textbooks of each branch.
A panel of three LLMs prompted as curriculum designers named the cornerstone domains and their
fundamental areas, and we finalized the 15 domains and 42 areas by majority vote (prompts
in~\autoref{sec:appendix_domain_validation} and~\autoref{sec:appendix_area_validation}).

\paragraph{Template Selection.}
Domain experts hand-authored all symbolic Python templates, grounding them directly in foundational
textbooks and authoritative engineering
handbooks~\citep[e.g.,][]{fogler2020elements, ulaby2015fundamentals, hibbeler2020, montgomery2013}
(see~\autoref{sec:appendix_books} for the complete list of source texts).
The template distribution per area is non-uniform and follows a pedagogical significance score
(1--5) that the same three LLMs assigned to each area within its domain
(\autoref{sec:appendix_significance_scoring}); \autoref{tab:area} gives the templates per area and
level.

\paragraph{Difficulty Levels.}
Domain experts also assigned each template one of three levels, Easy, Intermediate, or Advanced,
following the protocol in~\autoref{sec:appendix_difficulty} and scoring three dimensions:
conceptual complexity (isolated principles vs.\ multi-domain synthesis), mathematical
sophistication (direct algebraic substitution vs.\ differential equations), and procedural depth
(the length of interdependent reasoning chains).
\autoref{appendix:statistics} details how the templates are distributed across branches and levels.

\subsection{Template Generation Pipeline}
\label{subsec:templates}

\begin{figure*}[t]
    \centering
    \includegraphics[width=0.75\linewidth]{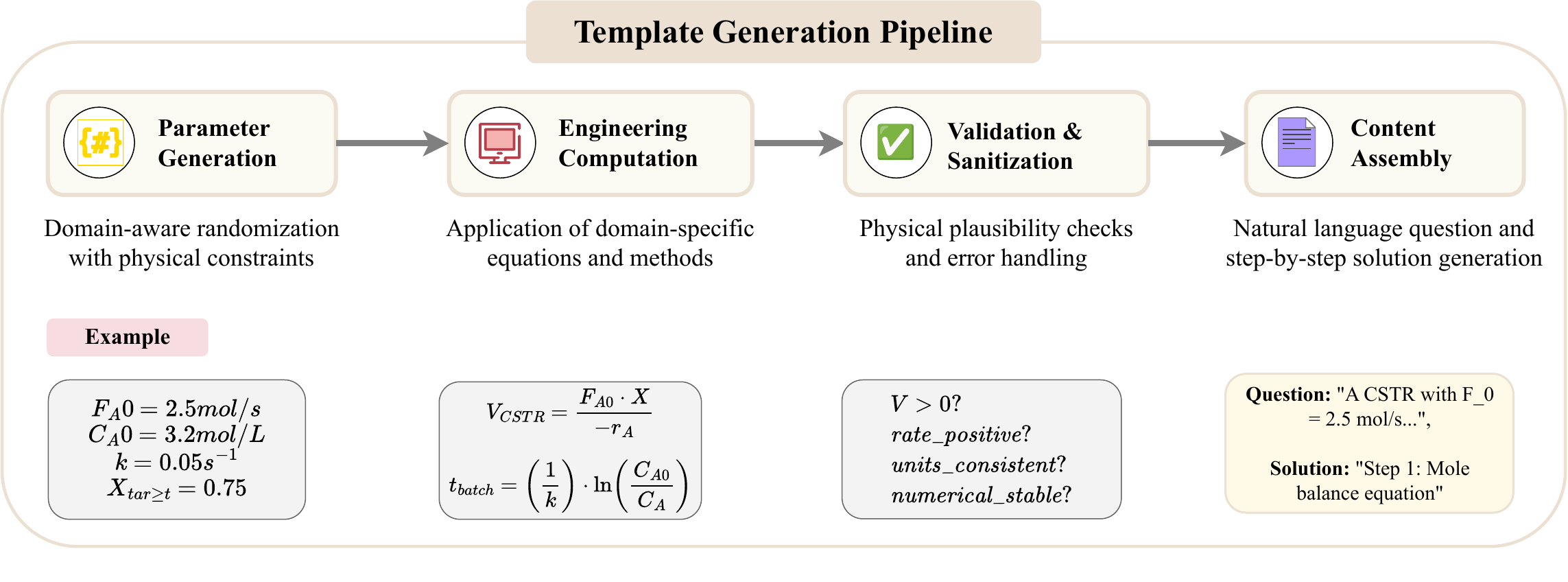}
    \caption{The \ourdataset \textbf{Template Generation Pipeline} maps domain-aware constraints to
    serialized question-solution pairs via physically validated computation, illustrated here using
    a CSTR volume calculation example.}
    \label{fig:template-generation}
\end{figure*}

The core of \ourdataset is the symbolic template, a Python blueprint for generating physically
grounded engineering problems.
As shown in~\autoref{fig:template-generation}, it samples domain-aware parameters from
authoritative data sources (\autoref{sec:appendix_data_sources}) and fundamental principles
(\autoref{sec:appendix_param_details}), redraws them until the scenario is physically valid,
computes every quantity with the governing equations, and writes the question and the gold
reasoning trace.
Because both come from one computation, every intermediate quantity is known exactly, with its
unit; these are the milestones whose coverage our evaluator measures.

Templates go beyond variable substitution: sampled parameters can change the physical regime, the
governing equation, and the reasoning path (\autoref{sec:appendix_template_examples}).
This diversity is physical rather than linguistic, since the question wording comes from the
template code; a paraphrase test on 275 instances of 114 templates (\autoref{appendix:paraphrase})
bounds the change in a model's score within $\pm 0.05$ for nine of the eleven models, and no change
is significant after correction.

\subsection{Template Certification}
\label{subsec:certification}

Because an error in a template recurs in every instance and gold trace, each template passes three
checks before it enters the benchmark (\autoref{appendix:certification}):


\paragraph{Automated Integrity Checks.}
At 500 seeds per template, automated checks verify that every printed calculation follows from its
printed operands, that generation is deterministic, and that every instance keeps the output format.
All 150 templates pass.

\paragraph{LLM Screen.}
Three LLM judges from families outside the evaluated models then score each template's code and
instances, and we revise the templates they flag, after which 147 of the 150 pass and none fails
critically.
The screen only guides revision, and the experts decide certification, since in their first round a
majority of experts reject 15 of the 147 templates that the screen passes.

\paragraph{Expert Certification.}
Three domain experts per branch review its 30 templates mixed with four planted defects, without
knowing which items are planted, and solve one instance of each template by hand before seeing its
gold trace.
Every expert rejects all four planted defects of their branch (60 of 60 readings), 459 of the 504
hand checks that yield a comparable number match the gold value within 1\%, and agreement on
approval is high (Gwet's AC1 0.914~\citep{gwet2008}; Fleiss'~$\kappa$~\citep{fleiss1971} 0.589,
deflated because nearly every template is approved~\citep{feinstein1990high}).
After six rounds, all 150 templates are certified by a unanimous vote of their three experts.

\section{Evaluation Framework}
\label{sec:evaluation}

Because each template computes every quantity of its gold trace, we check deterministically what can
be checked: the final value, the presence of the gold trace's intermediate values, and displayed
arithmetic; a judge decides what these checks leave open.~\autoref{box:worked_example} scores one response step by step.

\paragraph{Preliminaries.} An instance comprises a question, a gold trace, and milestones
$\mathcal{M}$: the values that the template computes, the gold trace states, and the question does
not give.
A response $r$ is the text a model returns, a sequence of steps ending in a final answer; no check
reads the reasoning text that some endpoints return separately.

\paragraph{Final Answer Accuracy.} The \emph{targets} of an instance are the milestones that its
gold final answer states or, where it states none, the answer's last value that the question does
not give.
A classification answer also targets the label that the gold answer states, such as a flow regime.
Let $u(x)$ be one unit in the last digit that a value $x$ displays, $\mathcal{C}$ a fixed set of
unit factors (e.g., m to mm), and $\epsilon = 0.002$, the check's one fitted constant.
A value $\hat{y}$ in the final answer matches a numerical target $y$ if, for some
$c \in \mathcal{C}$,
\begin{equation}
    |c\,\hat{y} - y| \le \max\bigl(\epsilon\,|y|,\; \min(c\,u(\hat{y}),\, 0.01\,|y|),\; u(y)\bigr),
    \label{eq:match}
\end{equation}
so a value matches within one unit of the last digit that either value displays, the response's
unit counting only up to 1\% of the target.
Symbolic answers are checked for equivalence on five templates and by their stated numbers on four
(\autoref{appendix:scoring}).
A response scores $v(r) = 1$ if every target matches, $0.5$ if only some do, and $0$ if none does or
no final answer is readable, since stating one within the output limit is part of the task.
\emph{Final Answer Accuracy} (FAC) is the mean of $v(r)$ over instances.

\paragraph{Milestone Coverage.} A milestone is reached if the response states its value within
0.5\% under a unit factor; 0.5\% is the tolerance at which the gold trace counts as stating a
milestone, not a fitted constant.
Matching compares values only, not the quantity's name, the unit written beside it, the step it
appears in, or the order and dependencies among steps, so a coincidental value counts, which the
chance floor of~\autoref{tab:coverage} measures.
Milestones that matching misses go to an LLM judge from outside the evaluated families,
\texttt{MiMo-V2.5-Pro}~\citep{mimov25pro}, which sees the question, the response, and each
milestone's name and value, but not the gold trace, and rules each reached, not needed, or missing.
We credit only reached, since the judge also excuses unstated values as not needed
(\autoref{appendix:scoring}).
With $\mathcal{R}(r)$ the milestones matched or ruled reached,
\begin{equation}
    m(r) = |\mathcal{R}(r)| \,/\, |\mathcal{M}|,
    \label{eq:coverage}
\end{equation}
and \emph{Milestone Coverage} (MC) is the mean of $m(r)$ over instances with milestones.
The judge decides 11\% to 24\% of milestones, depending on the model.
MC credits the gold route: a valid route that bypasses a milestone is not credited (case A
of~\autoref{box:coverage_cases}), so MC measures how much of the gold route a derivation states,
not whether it is free of error (case B).

\paragraph{Step-Level Diagnostics.} Two diagnostics look for flawed steps behind a correct answer.
The arithmetic check recomputes each displayed calculation it can parse from its printed operands
and flags a result that is not a correct rounding at its printed precision.
The judged step check sends every step the arithmetic check does not flag to the same judge, at the
provider's default settings, with the question and the gold steps, and flags a step labeled a
conceptual or calculation error; its flags are judged, not verified.
\emph{Arithmetic flags} and \emph{judged step flags} are the shares of correct-answer responses with
at least one such flag; MC and the two flag rates together are the process measures of this paper.
Both are flags with measured precision, not scores, since neither check sees every flawed step: on
60 planted misstated rules behind correct answers, none of our deterministic checks flags any, and
four LLM judges label at most about a third of them an error.

\paragraph{Validation.} An expert study guides and validates these checks: on 300 responses from
five LLMs outside the eleven we evaluate, each labeled step by step by three domain experts, the
final-answer check matches the experts' three-way verdict on 0.930 of responses, and milestone
matching reaches F1 0.926 alone and 0.958 with the judge (\autoref{appendix:validation}).
By answer kind, the final-answer check matches the experts' verdict on every scalar, array and
symbolic response of the study, and multipart responses agree least (\autoref{tab:answer_kinds}).
The study also informed the checks' settings, so these figures are in-sample, except that
$\epsilon$ is also fitted on one half of the responses and measured on the other
(\autoref{tab:scoring_settings}).
On the evaluated models, domain experts confirm 160 of 166 readings of correct verdicts, 75 of 86
of incorrect ones, and 83 of 98 readings of reached rulings (\autoref{appendix:readings}).

\section{Experiments}
\label{sec:experiments}

\subsection{Evaluated Model Suite}
\label{subsec:models}

We evaluate eleven LLMs: eight open-weights models,
\texttt{DeepSeek V4.1 Flash}~\citep{deepseekv41}, \texttt{Gemma 4 26B}~\citep{gemma4},
\texttt{GLM-5.3}~\citep{glm5}, \texttt{GLM-5.3-Flash}~\citep{glm5},
\texttt{gpt-oss-20b}~\citep{gptoss}, \texttt{Kimi K3}~\citep{kimik3},
\texttt{Muse Glimmer 30B}~\citep{museglimmer}, and \texttt{Qwen3-235B-2507}~\citep{qwen3}; and three
closed models, \texttt{Claude Sonnet 5}~\citep{claudesonnet5},
\texttt{Gemini 3.1 Flash-Lite}~\citep{gemini31flashlite}, and \texttt{GPT-5.4 mini}~\citep{gpt54}.
None of them wrote responses that we used to develop or validate the evaluator
(\autoref{sec:evaluation}), and none serves as a judge.
Two flagships, \texttt{GPT-5.4}~\citep{gpt54} and \texttt{DeepSeek V4 Pro}~\citep{deepseekv4}, run
only as anchors on a fixed subset of 450 instances (three per template), which the further
experiments also use.
\autoref{appendix:models} gives the selection rule and each model's size and identifiers.

\subsection{Experimental Setup}
\label{subsec:setup}

We evaluate the models at matched settings: the same zero-shot prompt, no tools or retrieval, and
reasoning on wherever the endpoint offers it, so \texttt{Gemini 3.1 Flash-Lite},
\texttt{Gemma 4 26B}, and \texttt{GPT-5.4 mini} answer with reasoning at medium effort and the
other eight models at their providers' defaults (\autoref{appendix:setup}; the three models'
default responses are the comparison of~\autoref{tab:matched}).
Each model answers each instance once (24,750 responses); 263 (1.1\%) have no readable final
answer and score 0.
The template is the unit of analysis, because a template's 15 instances and their gold traces come
from one procedure~\citep{miller2024errorbars}; every 95\% interval is a bootstrap over
templates~\citep{efron1993bootstrap}.
We compare two models with a sign-flip permutation test on their per-template
differences~\citep{dror2018hitchhiker} and correct each family of tests, such as the 55 pairs of
models at one setting, with Holm's method~\citep{holm1979simple}.
\autoref{appendix:setup} gives the decoding settings, the prompt, and the statistical protocol.

\subsection{Results}
\label{subsec:results}

\subsubsection{Overall Model Performance}
\label{subsec:overall_performance}

\begin{table*}[t]
\centering
\small
\setlength{\tabcolsep}{4pt}
\renewcommand{\arraystretch}{1.1}
\adjustbox{max width=\textwidth}{%
\begin{tabular}{l c c r c c r}
\toprule
 & \multicolumn{3}{c}{\textbf{Final answer}} & \multicolumn{3}{c}{\textbf{Derivation}} \\
\cmidrule(lr){2-4}\cmidrule(lr){5-7}
\textbf{Model} & \textbf{FAC $\uparrow$} & \textbf{\makecell{All 15 instances\\solved $\uparrow$}} & \textbf{\makecell{No readable\\answer}} & \textbf{MC $\uparrow$} & \textbf{\makecell{Arithmetic flags,\\correct answers}} & \textbf{\makecell{Calculations parsed\\per response}} \\
\midrule
\rowcolor{gray!10}\multicolumn{7}{l}{\textit{Open-weights LLMs}} \\
\texttt{DeepSeek V4.1 Flash} & 0.988 {\scriptsize\textcolor{gray}{(0.979--0.995)}} & 0.900 & 0.4\% & 0.899 {\scriptsize\textcolor{gray}{(0.874--0.922)}} & 0.006 {\scriptsize\textcolor{gray}{(0.003--0.010)}} & 3.3 \\
\texttt{Kimi K3} & 0.980 {\scriptsize\textcolor{gray}{(0.967--0.990)}} & 0.833 & 0.7\% & 0.912 {\scriptsize\textcolor{gray}{(0.888--0.934)}} & 0.007 {\scriptsize\textcolor{gray}{(0.003--0.011)}} & 2.8 \\
\texttt{GLM-5.3-Flash} & 0.972 {\scriptsize\textcolor{gray}{(0.953--0.987)}} & 0.853 & 2.0\% & 0.912 {\scriptsize\textcolor{gray}{(0.886--0.936)}} & 0.019 {\scriptsize\textcolor{gray}{(0.012--0.026)}} & 4.7 \\
\texttt{Muse Glimmer 30B} & 0.969 {\scriptsize\textcolor{gray}{(0.949--0.985)}} & 0.807 & 1.4\% & 0.897 {\scriptsize\textcolor{gray}{(0.871--0.922)}} & 0.058 {\scriptsize\textcolor{gray}{(0.044--0.074)}} & 3.3 \\
\texttt{GLM-5.3}$^{\dagger}$ & 0.950 {\scriptsize\textcolor{gray}{(0.921--0.975)}} & 0.813 & 4.4\% & 0.900 {\scriptsize\textcolor{gray}{(0.866--0.930)}} & 0.018 {\scriptsize\textcolor{gray}{(0.012--0.025)}} & 5.7 \\
\texttt{Gemma 4 26B}$^{\ddagger}$ & 0.942 {\scriptsize\textcolor{gray}{(0.916--0.964)}} & 0.733 & 0.1\% & 0.884 {\scriptsize\textcolor{gray}{(0.856--0.910)}} & 0.099 {\scriptsize\textcolor{gray}{(0.066--0.135)}} & 3.3 \\
\texttt{Qwen3-235B-2507}$^{\ast}$ & 0.899 {\scriptsize\textcolor{gray}{(0.869--0.926)}} & 0.513 & 0.0\% & 0.892 {\scriptsize\textcolor{gray}{(0.867--0.916)}} & 0.232 {\scriptsize\textcolor{gray}{(0.187--0.279)}} & 10.2 \\
\texttt{gpt-oss-20b} & 0.808 {\scriptsize\textcolor{gray}{(0.764--0.849)}} & 0.313 & 2.6\% & 0.812 {\scriptsize\textcolor{gray}{(0.775--0.847)}} & 0.147 {\scriptsize\textcolor{gray}{(0.116--0.181)}} & 3.8 \\
\rowcolor{gray!10}\multicolumn{7}{l}{\textit{Closed LLMs}} \\
\texttt{Claude Sonnet 5} & 0.987 {\scriptsize\textcolor{gray}{(0.980--0.993)}} & 0.867 & 0.1\% & 0.922 {\scriptsize\textcolor{gray}{(0.902--0.942)}} & 0.143 {\scriptsize\textcolor{gray}{(0.111--0.177)}} & 5.3 \\
\texttt{GPT-5.4 mini}$^{\ddagger}$ & 0.969 {\scriptsize\textcolor{gray}{(0.958--0.980)}} & 0.720 & 0.0\% & 0.892 {\scriptsize\textcolor{gray}{(0.866--0.917)}} & 0.075 {\scriptsize\textcolor{gray}{(0.057--0.096)}} & 3.3 \\
\texttt{Gemini 3.1 Flash-Lite}$^{\ddagger}$ & 0.917 {\scriptsize\textcolor{gray}{(0.884--0.945)}} & 0.693 & 0.0\% & 0.881 {\scriptsize\textcolor{gray}{(0.851--0.908)}} & 0.069 {\scriptsize\textcolor{gray}{(0.047--0.094)}} & 3.5 \\
\bottomrule
\end{tabular}
}
\caption{\textbf{Overall performance of the evaluated LLMs on \ourdataset.} Final Answer Accuracy
(FAC) and Milestone Coverage (MC) at matched settings, with 95\% intervals over templates in gray;
the share of templates solved on every instance; the share of responses that are empty or state no
answer, scored 0; the share of correct answers with a step the arithmetic check flags (flags, not
error counts); and the calculations the check parses per response. $^{\ddagger}$Run with reasoning
at medium effort. $^{\ast}$No reasoning setting offered. $^{\dagger}$99 responses empty at the
output ceiling.}
\label{tab:main_results}
\end{table*}

\autoref{tab:main_results} places the answer measures beside the derivation measures; we read the
final answer first and then what the derivation shows that the answer cannot.

\paragraph{A Ceiling and a Graded Order.} The final answer barely ranks the strongest models: the
first five are near the ceiling of the evaluation set, scoring 0.94 to 0.98 even on Advanced
templates, and only three of their ten pairs differ after Holm's correction.
The models differ in reliability more than in reach: no model fails every instance of more than
five of the 150 templates, yet the share of templates solved on every instance runs from 31\% to
90\%, a difference that one instance per problem would hide.
Across the eleven models the order is graded rather than tiered: 36 of the 55 pairs differ, and
only \texttt{gpt-oss-20b} differs from every other model (\autoref{tab:matched}).
The three models run with reasoning at medium effort all gain, \texttt{GPT-5.4 mini} by 0.118 and
into the first five; at their providers' defaults, the order splits into two tiers.

\paragraph{Derivations Behind Correct Answers.} The derivation shows what the final answer cannot:
how closely a model follows the gold route, and the slips a correct answer passes.
Neither the final answer nor MC as scored separates \texttt{Claude Sonnet 5} from
\texttt{DeepSeek V4.1 Flash}; matching alone does, but not once the milestones the judge rules
not needed for each model's route are set aside (\autoref{tab:coverage_variants}).
What separates them is how closely each follows the gold route, so MC ranks how much of that route a
model shows, not how soundly it reasons.
Correct answers can still contain a wrong calculation: the arithmetic check flags a displayed
calculation that does not follow from its own operands in up to 23.2\% of a model's correct
answers, and of the 189 flags a domain expert decided on, 171 are real slips (precision 0.905;
\autoref{tab:coverage}).
These flags miss most flawed steps: in the expert study, the arithmetic and judged step checks
together find only 0.36 of those the domain experts marked inside correct answers.

\paragraph{Derivations Behind Wrong Answers.} A wrong final answer hides how much of the derivation
is right: the last four models' wrong answers still state 46\% to 59\% of the gold milestones,
against a chance floor of at most 17.5\% (\autoref{tab:coverage}), and up to 25\% of them reach
every milestone yet end in an incorrect answer.
Where a derivation breaks, the judge's rulings name the gold quantity it fails to reach: a milestone
ruled missing marks most of these models' wrong answers (63\% to 78\%) but only 20\% of
\texttt{Claude Sonnet 5}'s, whose wrong answers mostly reach every milestone and still miss the
answer.
Weaker models more often lose answers on derivations they can carry out: on the 58 templates whose
instances all follow one derivation, the first five solve some but not all instances of 7\% to 34\%
of the templates and the other six of 14\% to 59\% (\autoref{appendix:results}), consistent with
the expert reading, in which calculation errors dominate (\autoref{subsec:error_analysis}).

\paragraph{Stability of the Order.} Every pair of models that differs holds under the small changes
we test: rewording, a looser or stricter answer tolerance, and repeated decoding reverse none of
them; only the reasoning setting, or how empty responses count, moves a model past one it differs
from.
Under an expert-checked paraphrase that keeps every number, unit, and symbol, no model's FAC
changes significantly, and for nine of the eleven models the change is bounded within $\pm 0.05$
at 90\% confidence; halving or doubling the tolerance swaps no pair of models that differs
(\autoref{appendix:results}), and decoding the four lowest models three more times moves FAC by a
standard deviation of at most 0.008 (\autoref{appendix:paraphrase}).
The paraphrase bounds memorized wording only, since the numbers and the problem structure stay.
Counting only the responses that state an answer moves \texttt{GLM-5.3}, whose 99 responses are
empty at the output ceiling, from seventh to first, so its place in~\autoref{tab:main_results}
reflects the output ceiling.

\subsubsection{Performance by Engineering Branch and Domain}
\label{subsec:branch_domain}

No branch is hard for every model: the lowest branch differs from model to model, and only one of
the 110 within-model branch comparisons holds after correction (\texttt{gpt-oss-20b}, electrical
above civil); with 30 templates per branch, differences below 3 to 19 points cannot be told from
sampling noise.
Weaker models are less even: the first five stay within 0.018 to 0.053 of their best branch and
above 0.88 in every domain, while the last five spread 0.071 to 0.226 and each falls below 0.81 in
some domain, though these spreads carry no test.
Thermodynamics is the lowest domain for four of the eleven models, but the domain means carry no
test, so this ranks domains, not fields (\autoref{appendix:branch_domain}).

\subsubsection{Performance by Difficulty Level}
\label{subsec:difficulty_level}

\begin{figure}[t]
    \centering
    \includegraphics[width=\columnwidth]{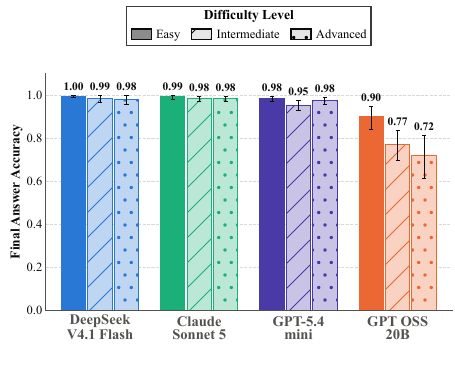}
    \caption{\textbf{Final Answer Accuracy by difficulty level for four representative models.} The mean
    over the 58 Easy, 58 Intermediate, and 34 Advanced templates, with 95\% intervals, for
    \texttt{DeepSeek V4.1 Flash} (first on FAC), \texttt{Claude Sonnet 5} (first on MC),
    \texttt{GPT-5.4 mini} (closed, fifth on FAC), and \texttt{gpt-oss-20b} (GPT OSS 20B in the figure).
    \autoref{tab:level_gap} gives every model's gap with its tests.}
    \label{fig:level_bars}
\end{figure}

Harder problems cost the weaker models most.
Every model scores lower on Advanced than on Easy templates (\autoref{fig:level_bars}), by 0.006 to
0.181, but the gap holds after Holm's correction only for \texttt{gpt-oss-20b} under Welch's
$t$-test, the test we read because scores vary more across Advanced templates, and for four models
under the permutation of level labels (\autoref{tab:level_gap}).
The length of the derivation splits the models more cleanly: on instances with six or more gold
milestones, the share of responses scored 0 reaches 0.104 to 0.254 for the last five against at
most 0.061 for the first five, and over readable responses each further milestone raises the odds
of a wrong answer by about a fifth when pooled over the models (odds ratio 1.21, 95\% interval 1.09
to 1.33), a slope that holds per model for three of the eleven (\autoref{tab:level_gap}).

\subsubsection{Reasoning, Equations, and Tools}
\label{subsec:reasoning_equations_tools}

Three experiments on a 450-instance subset test whether a larger model, the governing equations, or
a Python interpreter changes these results, and a fourth adds reasoning at medium effort
(\autoref{appendix:experiments}); each change starts from the providers' defaults and is paired with
the default responses on the same instances.

\paragraph{Flagship Anchors and Reasoning.} A larger model does not clearly raise the ceiling, but
reasoning helps a flagship as it helps a small model: of the two flagships run on the subset as
anchors, the best run, \texttt{GPT-5.4} with
reasoning at medium effort (0.997), still lies inside the interval of \texttt{DeepSeek V4.1 Flash}
on the same 450 instances.
\texttt{GPT-5.4}, which returns no reasoning tokens at its default, gains 0.046 with reasoning and
sheds most of its arithmetic flags, and \texttt{GPT-5.4 mini} with the same setting gains 0.057 in
MC while its flags on correct answers roughly halve: reasoning changes the derivation, not only the
answer.

\paragraph{Open Book.} Of the three models tested, recalling the governing equations holds back only
the weakest. Given the equations of the templates that state them, without data tables, \texttt{gpt-oss-20b} answers more correctly ($+0.068$), partly because fewer of its responses are empty at the output ceiling, while \texttt{Claude Sonnet 5} and \texttt{GPT-5.4 mini} stay within $\pm 0.05$.
Yet all three state more of the gold milestones: the equations shape the derivation even where they
leave the answer unchanged.

\paragraph{Open Tool.} An offered interpreter changes how the closed models compute but not how
well they score: \texttt{Claude Sonnet 5}, already near the ceiling, calls it on two thirds of the
instances and \texttt{GPT-5.4 mini}'s arithmetic flags on correct answers fall, yet neither changes
in FAC or MC beyond $\pm 0.05$.
The open-weights model is left out, since no endpoint serving it at 8-bit precision or higher
handles its tool calls reliably.
Because each model chooses when to call the interpreter, the result bounds what an offered tool
adds, not what required use would.

\subsection{Error Analysis}
\label{subsec:error_analysis}

\begin{figure}[t]
    \centering
    \includegraphics[width=0.85\columnwidth]{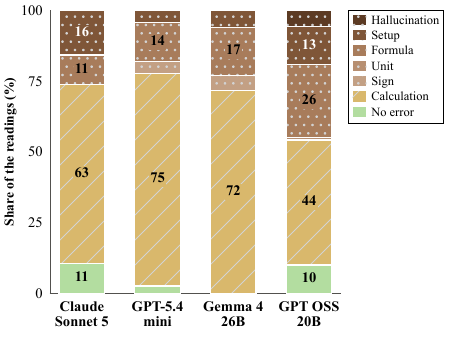}
    \caption{\textbf{Error categories of the wrong answers read.} Each bar gives one model's readings by
    category, from three domain experts per wrong answer (57 to 120 readings per model) on responses at
    the providers' default settings. \autoref{tab:error_by_level} gives them by level.}
    \label{fig:error_categories}
\end{figure}


Most wrong answers are calculation slips, and conceptual errors grow with difficulty.
Three domain experts read the full response of 137 wrong answers from four models that span the
FAC order, all at the providers' default settings, and assigned the first category that applies in a
six-category hierarchy from hallucination to calculation, or ``no error'' (Fleiss'~$\kappa$
0.912~\citep{fleiss1971}; \autoref{fig:error_categories}).
Calculation is the majority label for \texttt{GPT-5.4 mini}, \texttt{Gemma 4 26B}, and
\texttt{Claude Sonnet 5} and the largest for \texttt{gpt-oss-20b}, the model with the most
conceptual errors: a hallucination, a wrong setup, or a wrong formula.
Such conceptual errors are rarer on Easy problems than on Intermediate and Advanced ones
(\autoref{tab:error_by_level}).
The reading agrees with the reasoning setting and the experiments, each agreement resting on one
or two models: \texttt{GPT-5.4 mini} and \texttt{Gemma 4 26B}, whose wrong answers are mostly
calculation slips, gain the most from reasoning, and the equations lift only \texttt{gpt-oss-20b},
the model with the most conceptual errors. The first five lose few answer points, and the largest share goes to incorrect verdicts.
Over 11,250 responses, the first five models lose 231 answer points.
Of these, 71 go to responses with no readable final answer and 34 to 68 partial answers at half
credit.
Of their 126 incorrect verdicts, 38 lie within 0.2\% of a target whose digits the question
prescribes, which makes them incorrect by its terms, 5 are symbolic answers on one template checked
for equivalence, and 83 fall in neither group (\autoref{appendix:error_analysis}).
On the final answer, then, the evaluation set is near ceiling for the first five, and most of what
still differs among them lies in the derivation: how closely it follows the gold route, and how
often its displayed arithmetic slips, which the arithmetic check flags in 0.6\% to 14.3\% of their
correct answers, at rates that depend on how much arithmetic each displays.
Since a response that misstates the rule it applies but computes the right answer passes every
deterministic check, the process measures mark where to read, and only a reader can confirm such a
rule.

\section{Conclusion and Future Work}
\label{sec:conclusion}
We introduced \ourdataset, a symbolic benchmark that pairs each engineering problem with an exact
gold reasoning trace, and an evaluator that checks responses against this trace, calling an LLM
judge only where matching cannot decide.
The strongest LLMs were hard to tell apart by their final answers, yet their derivations differed in how closely they followed the gold route and how often their displayed arithmetic was flagged.
Most failures were calculation slips rather than conceptual errors, which were rarer on easy
problems. Enabling reasoning raised final-answer accuracy for every model compared over the full evaluation,
and no pair of models that differs reversed under rewording, a changed answer tolerance, or repeated
decoding of the lowest-scoring models.
These results suggest that, near the ceiling, progress is better measured in the derivation than in the final answer.
Future work will check the principles and assumptions a solution applies, not only its numbers, since a misstated rule behind a correct result escapes every numerical test, and will turn the gold reasoning traces into feedback for training and selecting LLMs. The approach could also extend to open-ended design and tool-based workflows.

\section*{Limitations}
\label{sec:limitations}





This work has several limitations, which future research can address.

First, \ourdataset is written in English. Engineering is taught and practiced in many languages, and models may perform differently when the same problems are posed in them. Since the question text of every template is generated from code, the benchmark can be extended to other languages by translating its templates and adapting units and notation to regional
conventions.

Second, the templates span several core engineering branches, and adding disciplines such as aerospace, biomedical, or environmental engineering would broaden the benchmark further.
Because all templates share one format, new branches can be written and certified in the same way as the existing ones.

Third, the problems are presented as text, while engineering work often involves diagrams,
schematics, and plots. Extending the benchmark to problems that include such material would allow multimodal models to be
evaluated on it.

Finally, we evaluate a selection of LLMs available at the time of our experiments, and the results
reflect those model versions. New models and versions are released frequently, and the benchmark and its evaluator can be applied
to them as they appear.

\section*{Ethics Statement}

This work uses synthetic data generated from symbolic templates and standard engineering principles;
no private, sensitive, or proprietary industrial data was used.
The domain experts who certified the templates and read the models' responses are not named, and
their individual labels are reported only as counts and agreement.
The results describe what LLMs compute on examination-style problems and do not support deploying
them in safety-critical engineering systems.
\ourdataset and its evaluator will be released under the MIT License.

\paragraph{Data License}

The \ourdataset dataset and associated code will be released under the MIT License.




\bibliography{custom}

@inproceedings{zheng2024processbench,
  title={{ProcessBench}: Identifying Process Errors in Mathematical Reasoning},
  author={Zheng, Chujie and Zhang, Zhenru and Zhang, Beichen and Lin, Runji and Lu, Keming and Yu, Bowen and Liu, Dayiheng and Zhou, Jingren and Lin, Junyang},
  booktitle={Proceedings of the 63rd Annual Meeting of the Association for Computational Linguistics (Volume 1: Long Papers)},
  series={ACL~'25},
  address={Vienna, Austria},
  publisher={Association for Computational Linguistics},
  pages={1009--1024},
  year={2025},
  doi={10.18653/v1/2025.acl-long.50},
  url={https://aclanthology.org/2025.acl-long.50/}
}

@inproceedings{du2025supergpqa,
  title={{SuperGPQA}: Scaling {LLM} Evaluation across 285 Graduate Disciplines},
  author={{M-A-P Team} and Du, Xinrun and Yao, Yifan and Ma, Kaijing and Wang, Bingli and Zheng, Tianyu and Zhu, King and Liu, Minghao and Liang, Yiming and Jin, Xiaolong and others},
  booktitle={Advances in Neural Information Processing Systems: Datasets and Benchmarks Track},
  series={NeurIPS~'25},
  address={San Diego, CA, USA},
  year={2025},
  url={https://arxiv.org/abs/2502.14739}
}

@inproceedings{zhou2025engibench,
  title={{EngiBench}: A Benchmark for Evaluating Large Language Models on Engineering Problem Solving},
  author={Zhou, Xiyuan and Wang, Xinlei and He, Yirui and Wu, Yang and Zou, Ruixi and Cheng, Yuheng and Xie, Yulu and Liu, Wenxuan and Zhao, Huan and Xu, Yan and Gu, Jinjin and Zhao, Junhua},
  booktitle={Findings of the Association for Computational Linguistics: ACL 2026},
  series={ACL~'26},
  address={San Diego, California, United States},
  publisher={Association for Computational Linguistics},
  pages={36308--36334},
  year={2026},
  doi={10.18653/v1/2026.findings-acl.1810},
  url={https://aclanthology.org/2026.findings-acl.1810/}
}

@article{syed2024transportbench,
  title={Benchmarking the Capabilities of Large Language Models in Transportation System Engineering: Accuracy, Consistency, and Reasoning Behaviors},
  author={Syed, Usman and Light, Ethan and Guo, Xingang and Zhang, Huan and Qin, Lianhui and Ouyang, Yanfeng and Hu, Bin},
  journal={ArXiv preprint},
  volume={arXiv:2408.08302},
  year={2024},
  url={https://arxiv.org/abs/2408.08302}
}

@article{zheng2026tpscalcbench,
  title={{TPS-CalcBench}: A Benchmark and Diagnostic Evaluation Framework for {LLM} Analytical Calculation Competence in Hypersonic Thermal Protection System Engineering},
  author={Zheng, Jinglai and Qiao, Chuhan and Huang, Haiming},
  journal={ArXiv preprint},
  volume={arXiv:2604.17966},
  year={2026},
  url={https://arxiv.org/abs/2604.17966}
}

@article{duzkar2026thermoqa,
  title={{ThermoQA}: A Three-Tier Benchmark for Evaluating Thermodynamic Reasoning in Large Language Models},
  author={D{\"u}zkar, Kemal},
  journal={ArXiv preprint},
  volume={arXiv:2604.19758},
  year={2026},
  url={https://arxiv.org/abs/2604.19758}
}

@inproceedings{mirzadeh2024gsmsymbolic,
  title={{GSM-Symbolic}: Understanding the Limitations of Mathematical Reasoning in Large Language Models},
  author={Mirzadeh, Iman and Alizadeh, Keivan and Shahrokhi, Hooman and Tuzel, Oncel and Bengio, Samy and Farajtabar, Mehrdad},
  booktitle={Proceedings of the Thirteenth International Conference on Learning Representations},
  series={ICLR~'25},
  address={Singapore},
  pages={94743--94765},
  year={2025},
  url={https://proceedings.iclr.cc/paper_files/paper/2025/hash/ec2e7a896f8250986b3907f57621ce94-Abstract-Conference.html}
}

@inproceedings{xie2025finchain,
  title={{FinChain}: A Symbolic Benchmark for Verifiable Chain-of-Thought Financial Reasoning},
  author={Xie, Zhuohan and Orel, Daniil and Thareja, Rushil and Sahnan, Dhruv and Madmoun, Hachem and Zhang, Fan and Banerjee, Debopriyo and Georgiev, Georgi and Peng, Xueqing and Qian, Lingfei and Huang, Jimin and Su, Jinyan and Singh, Aaryamonvikram and Xing, Rui and Elbadry, Rania and Xu, Chen and Li, Haonan and Koto, Fajri and Koychev, Ivan and Chakraborty, Tanmoy and Wang, Yuxia and Lahlou, Salem and Stoyanov, Veselin and Ananiadou, Sophia and Nakov, Preslav},
  booktitle={Proceedings of the 64th Annual Meeting of the Association for Computational Linguistics (Volume 1: Long Papers)},
  series={ACL~'26},
  address={San Diego, California, United States},
  publisher={Association for Computational Linguistics},
  pages={14529--14553},
  year={2026},
  doi={10.18653/v1/2026.acl-long.662},
  url={https://aclanthology.org/2026.acl-long.662/}
}

@inproceedings{panickssery2024selfrecognition,
  title={{LLM} Evaluators Recognize and Favor Their Own Generations},
  author={Panickssery, Arjun and Bowman, Samuel R. and Feng, Shi},
  booktitle={Advances in Neural Information Processing Systems},
  series={NeurIPS~'24},
  address={Vancouver, BC, Canada},
  volume={37},
  year={2024},
  url={https://arxiv.org/abs/2404.13076}
}

@inproceedings{li2026preferenceleakage,
  title={Preference Leakage: A Contamination Problem in {LLM}-as-a-judge},
  author={Li, Dawei and Sun, Renliang and Huang, Yue and Zhong, Ming and Jiang, Bohan and Han, Jiawei and Zhang, Xiangliang and Wang, Wei and Liu, Huan},
  booktitle={Proceedings of the Fourteenth International Conference on Learning Representations},
  series={ICLR~'26},
  address={Rio de Janeiro, Brazil},
  pages={60486--60511},
  year={2026},
  url={https://proceedings.iclr.cc/paper_files/paper/2026/hash/6297baf3f5c98146b62dd3a1bffe068e-Abstract-Conference.html}
}

@inproceedings{zhao2026prismphysics,
  title={{PRISM-Physics}: Causal {DAG}-Based Process Evaluation for Physics Reasoning},
  author={Zhao, Wanjia and Ma, Qinwei and Shi, Jingzhe and Wu, Shirley and Han, Jiaqi and Xiao, Yijia and Chen, Si-Yuan and Luo, Xiao and Schmidt, Ludwig and Zou, James},
  booktitle={Proceedings of the Fourteenth International Conference on Learning Representations},
  series={ICLR~'26},
  address={Rio de Janeiro, Brazil},
  year={2026},
  url={https://arxiv.org/abs/2510.03185}
}

@inproceedings{li2025eeebench,
  title={{EEE-Bench}: A Comprehensive Multimodal Electrical and Electronics Engineering Benchmark},
  author={Li, Ming and Zhong, Jike and Chen, Tianle and Lai, Yuxiang and Psounis, Konstantinos},
  booktitle={Proceedings of the IEEE/CVF Conference on Computer Vision and Pattern Recognition},
  series={CVPR~'25},
  address={Nashville, TN, USA},
  year={2025},
  url={https://arxiv.org/abs/2411.01492}
}

@article{naser2026eri,
  title={Engineering Reasoning and Instruction ({ERI}) Benchmark: A Large Taxonomy-driven Dataset for Foundation Models and Agents},
  author={Naser, M. Z. and Bani Awwad, Ahmad and McCreery, Zoie and Eissa, Radwa and Naser, Ahmad and Cusatis, Gianluca and Metcalf, Andrew and Madathil, Kapil and Abdalla, Jamal and Kodur, Venkatesh and others},
  journal={Computers \& Industrial Engineering},
  volume={221},
  year={2026},
  doi={10.1016/j.cie.2026.112333},
  url={https://doi.org/10.1016/j.cie.2026.112333}
}

@article{skelic2025circuit,
  title={{CIRCUIT}: A Benchmark for Circuit Interpretation and Reasoning Capabilities of {LLMs}},
  author={Skelic, Lejla and Xu, Yan and Cox, Matthew and Lu, Wenjie and Yu, Tao and Han, Ruonan},
  journal={ArXiv preprint},
  volume={arXiv:2502.07980},
  year={2025},
  url={https://arxiv.org/abs/2502.07980}
}

@inproceedings{lightman2023verify,
  title={Let's Verify Step by Step},
  author={Lightman, Hunter and Kosaraju, Vineet and Burda, Yura and Edwards, Harri and Baker, Bowen and Lee, Teddy and Leike, Jan and Schulman, John and Sutskever, Ilya and Cobbe, Karl},
  booktitle={Proceedings of the Twelfth International Conference on Learning Representations},
  series={ICLR~'24},
  address={Vienna, Austria},
  year={2024},
  url={https://arxiv.org/abs/2305.20050}
}

@inproceedings{zeng2025versaprm,
  title={{VersaPRM}: Multi-Domain Process Reward Model via Synthetic Reasoning Data},
  author={Zeng, Thomas and Zhang, Shuibai and Wu, Shutong and Classen, Christian and Chae, Daewon and Ewer, Ethan and Lee, Minjae and Kim, Heeju and Kang, Wonjun and Kunde, Jackson and others},
  booktitle={Proceedings of the 42nd International Conference on Machine Learning},
  series={ICML~'25},
  address={Vancouver, BC, Canada},
  year={2025},
  url={https://arxiv.org/abs/2502.06737}
}

@inproceedings{krumdick2026nofreelabels,
  title={No Free Labels: Limitations of {LLM}-as-a-Judge Without Human Grounding},
  author={Krumdick, Michael and Lovering, Charles and Reddy, Varshini and Ebner, Seth and Tanner, Chris},
  booktitle={Proceedings of the Third Conference on Language Modeling},
  series={COLM~'26},
  address={San Francisco, CA, USA},
  year={2026},
  url={https://arxiv.org/abs/2503.05061}
}

@inproceedings{fan2024hardmath,
  title={{HARDMath}: A Benchmark Dataset for Challenging Problems in Applied Mathematics},
  author={Fan, Jingxuan and Martinson, Sarah and Wang, Erik Y. and Hausknecht, Kaylie and Brenner, Jonah and Liu, Danxian and Peng, Nianli and Wang, Corey and Brenner, Michael P.},
  booktitle={Proceedings of the Thirteenth International Conference on Learning Representations},
  series={ICLR~'25},
  address={Singapore},
  year={2025},
  url={https://arxiv.org/abs/2410.09988}
}

@inproceedings{zhang2024gsm1k,
  title={A Careful Examination of Large Language Model Performance on Grade School Arithmetic},
  author={Zhang, Hugh and Da, Jeff and Lee, Dean and Robinson, Vaughn and Wu, Catherine and Song, Will and Zhao, Tiffany and Raja, Pranav and Zhuang, Charlotte and Slack, Dylan and others},
  booktitle={Advances in Neural Information Processing Systems: Datasets and Benchmarks Track},
  series={NeurIPS~'24},
  address={Vancouver, BC, Canada},
  volume={37},
  year={2024},
  url={https://arxiv.org/abs/2405.00332}
}

@inproceedings{imani2025sympybench,
  title={{SymPyBench}: A Dynamic Benchmark for Scientific Reasoning with Executable {Python} Code},
  author={Imani, Shima and Moon, Seungwhan and Ahmadyan, Adel and Zhang, Lu and Ahmed, Kirmani and Damavandi, Babak},
  booktitle={Proceedings of the 19th Conference of the European Chapter of the Association for Computational Linguistics (Volume 5: Industry Track)},
  series={EACL~'26},
  address={Rabat, Morocco},
  publisher={Association for Computational Linguistics},
  pages={105--118},
  year={2026},
  doi={10.18653/v1/2026.eacl-industry.8},
  url={https://aclanthology.org/2026.eacl-industry.8/}
}

@inproceedings{chen2026optengine,
  title={{OPT-Engine}: Benchmarking the Limits of {LLMs} in Optimization Modeling via Complexity Scaling},
  author={Chen, Yitian and Cheng, Cheng and Sun, Yinan and Ling, Zi and Ge, Dongdong},
  booktitle={Proceedings of the 43rd International Conference on Machine Learning},
  series={ICML~'26},
  address={Seoul, South Korea},
  year={2026},
  url={https://arxiv.org/abs/2601.19924}
}

@article{wasiq2026engvqa,
  title={Do {VLMs} Reason Like Engineers? A Benchmark and a Stage-wise Evaluation},
  author={Wasiq, Syed and Tawseeq, Syed Mohamad and Bangde, Yashwant Pravinrao and Roy, Debaditya},
  journal={ArXiv preprint},
  volume={arXiv:2606.10833},
  year={2026},
  url={https://arxiv.org/abs/2606.10833}
}

@inproceedings{wang2023mathshepherd,
  title={{Math-Shepherd}: Verify and Reinforce {LLMs} Step-by-step without Human Annotations},
  author={Wang, Peiyi and Li, Lei and Shao, Zhihong and Xu, R. X. and Dai, Damai and Li, Yifei and Chen, Deli and Wu, Y. and Sui, Zhifang},
  booktitle={Proceedings of the 62nd Annual Meeting of the Association for Computational Linguistics (Volume 1: Long Papers)},
  series={ACL~'24},
  address={Bangkok, Thailand},
  publisher={Association for Computational Linguistics},
  pages={9426--9439},
  year={2024},
  doi={10.18653/v1/2024.acl-long.510},
  url={https://aclanthology.org/2024.acl-long.510/}
}

@inproceedings{zhang2025prmlessons,
  title={The Lessons of Developing Process Reward Models in Mathematical Reasoning},
  author={Zhang, Zhenru and Zheng, Chujie and Wu, Yangzhen and Zhang, Beichen and Lin, Runji and Yu, Bowen and Liu, Dayiheng and Zhou, Jingren and Lin, Junyang},
  booktitle={Findings of the Association for Computational Linguistics: ACL 2025},
  series={ACL~'25},
  address={Vienna, Austria},
  publisher={Association for Computational Linguistics},
  pages={10495--10516},
  year={2025},
  doi={10.18653/v1/2025.findings-acl.547},
  url={https://aclanthology.org/2025.findings-acl.547/}
}

@misc{abet2024criteria,
  title={Criteria for Accrediting Engineering Programs, 2024--2025},
  author={{ABET}},
  year={2024}
}

@article{gwet2008,
  title={Computing Inter-Rater Reliability and Its Variance in the Presence of High Agreement},
  author={Gwet, Kilem Li},
  journal={British Journal of Mathematical and Statistical Psychology},
  volume={61},
  number={1},
  pages={29--48},
  year={2008},
  doi={10.1348/000711006X126600}
}

@article{fleiss1971,
  title={Measuring Nominal Scale Agreement Among Many Raters},
  author={Fleiss, Joseph L.},
  journal={Psychological Bulletin},
  volume={76},
  number={5},
  pages={378--382},
  year={1971},
  doi={10.1037/h0031619}
}

@article{feinstein1990high,
  title={High Agreement but Low Kappa: {I}. {T}he Problems of Two Paradoxes},
  author={Feinstein, Alvan R. and Cicchetti, Domenic V.},
  journal={Journal of Clinical Epidemiology},
  volume={43},
  number={6},
  pages={543--549},
  year={1990},
  doi={10.1016/0895-4356(90)90158-L}
}

@book{fogler2020elements,
  title={Elements of Chemical Reaction Engineering},
  author={Fogler, H. Scott},
  edition={6th},
  publisher={Pearson},
  address={Boston, MA, USA},
  year={2020}
}

@book{ulaby2015fundamentals,
  title={Fundamentals of Applied Electromagnetics},
  author={Ulaby, Fawwaz T. and Ravaioli, Umberto},
  edition={7th},
  publisher={Pearson},
  address={Upper Saddle River, NJ, USA},
  year={2015}
}

@book{hibbeler2020,
  title={Structural Analysis},
  author={Hibbeler, Russell C.},
  edition={10th {SI}},
  publisher={Pearson},
  address={Harlow, England},
  year={2020}
}

@book{montgomery2013,
  title={Introduction to Statistical Quality Control},
  author={Montgomery, Douglas C.},
  edition={7th},
  publisher={John Wiley \& Sons},
  address={Hoboken, NJ, USA},
  year={2013}
}

@book{asce2019cebok,
  title={Civil Engineering Body of Knowledge: Preparing the Future Civil Engineer},
  author={{Civil Engineering Body of Knowledge 3 Task Committee}},
  edition={3rd},
  publisher={American Society of Civil Engineers},
  address={Reston, VA, USA},
  year={2019},
  doi={10.1061/9780784415221}
}

@misc{iise2021bok,
  title={Industrial and Systems Engineering Body of Knowledge},
  author={{IISE}},
  year={2021},
  url={https://www.iise.org/uploadedFiles/IIE/Technical_Resources/Publications/Industrial_and_Systems_Engineering_Body_of_Knowledge_2021.pdf}
}

@article{cohen1960,
  title={A Coefficient of Agreement for Nominal Scales},
  author={Cohen, Jacob},
  journal={Educational and Psychological Measurement},
  volume={20},
  number={1},
  pages={37--46},
  year={1960},
  doi={10.1177/001316446002000104}
}

@article{meurer2017sympy,
  title={{SymPy}: Symbolic Computing in {Python}},
  author={Meurer, Aaron and Smith, Christopher P. and Paprocki, Mateusz and {\v{C}}ert{\'\i}k, Ond{\v{r}}ej and Kirpichev, Sergey B. and Rocklin, Matthew and Kumar, Amit and Ivanov, Sergiu and Moore, Jason K. and Singh, Sartaj and others},
  journal={PeerJ Computer Science},
  volume={3},
  pages={e103},
  year={2017},
  doi={10.7717/peerj-cs.103}
}

@misc{mimov25pro,
  title={{MiMo-V2.5-Pro}},
  author={{Xiaomi MiMo Team}},
  year={2026},
  url={https://huggingface.co/XiaomiMiMo/MiMo-V2.5-Pro}
}

@misc{grok46,
  title={Model Card: {Grok 4.6}},
  author={{xAI}},
  year={2026},
  url={https://media.x.ai/v1/website/card-4p6-4cd2dc57.pdf}
}

@article{minimaxm3,
  title={{MiniMax} Sparse Attention},
  author={Lai, Xunhao and Xu, Weiqi and Yang, Yufeng and Chen, Qiaorui and Xu, Yang and others},
  journal={ArXiv preprint},
  volume={arXiv:2606.13392},
  year={2026},
  url={https://arxiv.org/abs/2606.13392}
}

@misc{gpt5,
  title={{GPT-5} System Card},
  author={{OpenAI}},
  year={2025},
  url={https://openai.com/index/gpt-5-system-card/}
}

@misc{claudeopus47,
  title={System Card: {Claude Opus 4.7}},
  author={{Anthropic}},
  year={2026},
  url={https://www.anthropic.com/claude-opus-4-7-system-card}
}

@misc{claudeopus45,
  title={System Card: {Claude Opus 4.5}},
  author={{Anthropic}},
  year={2025},
  url={https://www.anthropic.com/claude-opus-4-5-system-card}
}

@misc{gemini31pro,
  title={{Gemini 3.1 Pro} Model Card},
  author={{Google DeepMind}},
  year={2026},
  url={https://deepmind.google/models/model-cards/gemini-3-1-pro/}
}

@article{guo2025deepseekr1,
  title={{DeepSeek-R1} Incentivizes Reasoning in {LLMs} through Reinforcement Learning},
  author={Guo, Daya and Yang, Dejian and Zhang, Haowei and Song, Junxiao and Wang, Peiyi and Zhu, Qihao and Xu, Runxin and Zhang, Ruoyu and others},
  journal={Nature},
  volume={645},
  number={8081},
  pages={633--638},
  year={2025},
  doi={10.1038/s41586-025-09422-z}
}

@article{grattafiori2024llama3,
  title={The {Llama 3} Herd of Models},
  author={Grattafiori, Aaron and Dubey, Abhimanyu and Jauhri, Abhinav and Pandey, Abhinav and Kadian, Abhishek and Al-Dahle, Ahmad and others},
  journal={ArXiv preprint},
  volume={arXiv:2407.21783},
  year={2024},
  url={https://arxiv.org/abs/2407.21783}
}

@article{miller2024errorbars,
  title={Adding Error Bars to Evals: A Statistical Approach to Language Model Evaluations},
  author={Miller, Evan},
  journal={ArXiv preprint},
  volume={arXiv:2411.00640},
  year={2024},
  url={https://arxiv.org/abs/2411.00640}
}

@book{efron1993bootstrap,
  title={An Introduction to the Bootstrap},
  author={Efron, Bradley and Tibshirani, Robert J.},
  publisher={Chapman and Hall},
  address={New York, NY, USA},
  year={1993},
  doi={10.1007/978-1-4899-4541-9}
}

@inproceedings{dror2018hitchhiker,
  title={The Hitchhiker's Guide to Testing Statistical Significance in Natural Language Processing},
  author={Dror, Rotem and Baumer, Gili and Shlomov, Segev and Reichart, Roi},
  booktitle={Proceedings of the 56th Annual Meeting of the Association for Computational Linguistics (Volume 1: Long Papers)},
  series={ACL~'18},
  address={Melbourne, Australia},
  publisher={Association for Computational Linguistics},
  pages={1383--1392},
  year={2018},
  doi={10.18653/v1/P18-1128},
  url={https://aclanthology.org/P18-1128/}
}

@article{welch1947generalization,
  title={The Generalization of ``{S}tudent's'' Problem When Several Different Population Variances Are Involved},
  author={Welch, Bernard L.},
  journal={Biometrika},
  volume={34},
  number={1--2},
  pages={28--35},
  year={1947},
  doi={10.1093/biomet/34.1-2.28}
}

@article{holm1979simple,
  title={A Simple Sequentially Rejective Multiple Test Procedure},
  author={Holm, Sture},
  journal={Scandinavian Journal of Statistics},
  volume={6},
  number={2},
  pages={65--70},
  year={1979},
  url={https://www.jstor.org/stable/4615733}
}

@article{deepseekv41,
  title={{DeepSeek-V4.1-Flash}: Pushing the Limits of {KV} Cache Compression},
  author={{DeepSeek-AI}},
  journal={ArXiv preprint},
  volume={arXiv:2609.19969},
  year={2026},
  url={https://arxiv.org/abs/2609.19969}
}

@article{gemma4,
  title={{Gemma 4} Technical Report},
  author={{Gemma Team}},
  journal={ArXiv preprint},
  volume={arXiv:2607.02770},
  year={2026},
  url={https://arxiv.org/abs/2607.02770}
}

@article{glm5,
  title={{GLM-5}: from Vibe Coding to Agentic Engineering},
  author={{GLM-5-Team}},
  journal={ArXiv preprint},
  volume={arXiv:2602.15763},
  year={2026},
  url={https://arxiv.org/abs/2602.15763}
}

@article{gptoss,
  title={{gpt-oss-120b} \& {gpt-oss-20b} Model Card},
  author={{OpenAI}},
  journal={ArXiv preprint},
  volume={arXiv:2508.10925},
  year={2025},
  url={https://arxiv.org/abs/2508.10925}
}

@article{kimik3,
  title={{Kimi K3}: Open Frontier Intelligence},
  author={{Kimi Team}},
  journal={ArXiv preprint},
  volume={arXiv:2607.24653},
  year={2026},
  url={https://arxiv.org/abs/2607.24653}
}

@misc{museglimmer,
  title={{Muse Glimmer} Model Card},
  author={{Meta}},
  year={2026},
  url={https://huggingface.co/meta-models/Muse-Glimmer-30B}
}

@article{qwen3,
  title={{Qwen3} Technical Report},
  author={{Qwen Team}},
  journal={ArXiv preprint},
  volume={arXiv:2505.09388},
  year={2025},
  url={https://arxiv.org/abs/2505.09388}
}

@misc{claudesonnet5,
  title={System Card: {Claude Sonnet 5}},
  author={{Anthropic}},
  year={2026},
  url={https://www.anthropic.com/claude-sonnet-5-system-card}
}

@misc{gemini31flashlite,
  title={{Gemini 3.1 Flash-Lite} Model Card},
  author={{Google DeepMind}},
  year={2026},
  url={https://deepmind.google/models/model-cards/gemini-3-1-flash-lite/}
}

@misc{gpt54,
  title={{GPT-5.4 Thinking} System Card},
  author={{OpenAI}},
  year={2026},
  url={https://deploymentsafety.openai.com/gpt-5-4-thinking}
}

@article{deepseekv4,
  title={{DeepSeek-V4}: Towards Highly Efficient Million-Token Context Intelligence},
  author={{DeepSeek-AI}},
  journal={ArXiv preprint},
  volume={arXiv:2606.19348},
  year={2026},
  url={https://arxiv.org/abs/2606.19348}
}

@article{kendall1938,
  title={A New Measure of Rank Correlation},
  author={Kendall, Maurice G.},
  journal={Biometrika},
  volume={30},
  number={1--2},
  pages={81--93},
  year={1938},
  doi={10.1093/biomet/30.1-2.81}
}

@article{spearman1904,
  title={The Proof and Measurement of Association between Two Things},
  author={Spearman, Charles},
  journal={The American Journal of Psychology},
  volume={15},
  number={1},
  pages={72--101},
  year={1904},
  doi={10.2307/1412159}
}

@article{wilcoxon1945,
  title={Individual Comparisons by Ranking Methods},
  author={Wilcoxon, Frank},
  journal={Biometrics Bulletin},
  volume={1},
  number={6},
  pages={80--83},
  year={1945},
  doi={10.2307/3001968}
}

@article{mcnemar1947,
  title={Note on the Sampling Error of the Difference between Correlated Proportions or Percentages},
  author={McNemar, Quinn},
  journal={Psychometrika},
  volume={12},
  number={2},
  pages={153--157},
  year={1947},
  doi={10.1007/BF02295996}
}

@inproceedings{zhu2024dyval,
  title={{DyVal}: Dynamic Evaluation of Large Language Models for Reasoning Tasks},
  author={Zhu, Kaijie and Chen, Jiaao and Wang, Jindong and Gong, Neil Zhenqiang and Yang, Diyi and Xie, Xing},
  booktitle={Proceedings of the Twelfth International Conference on Learning Representations},
  series={ICLR~'24},
  address={Vienna, Austria},
  pages={18091--18128},
  year={2024},
  url={https://proceedings.iclr.cc/paper_files/paper/2024/hash/4ea4a1ea4d9ff273688c8e92bd087112-Abstract-Conference.html}
}

@inproceedings{prasad2023receval,
  title={{ReCEval}: Evaluating Reasoning Chains via Correctness and Informativeness},
  author={Prasad, Archiki and Saha, Swarnadeep and Zhou, Xiang and Bansal, Mohit},
  booktitle={Proceedings of the 2023 Conference on Empirical Methods in Natural Language Processing},
  series={EMNLP~'23},
  address={Singapore},
  publisher={Association for Computational Linguistics},
  pages={10066--10086},
  year={2023},
  doi={10.18653/v1/2023.emnlp-main.622},
  url={https://aclanthology.org/2023.emnlp-main.622/}
}

@inproceedings{li2024gsmplus,
  title={{GSM-Plus}: A Comprehensive Benchmark for Evaluating the Robustness of {LLMs} as Mathematical Problem Solvers},
  author={Li, Qintong and Cui, Leyang and Zhao, Xueliang and Kong, Lingpeng and Bi, Wei},
  booktitle={Proceedings of the 62nd Annual Meeting of the Association for Computational Linguistics (Volume 1: Long Papers)},
  series={ACL~'24},
  address={Bangkok, Thailand},
  publisher={Association for Computational Linguistics},
  pages={2961--2984},
  year={2024},
  doi={10.18653/v1/2024.acl-long.163},
  url={https://aclanthology.org/2024.acl-long.163/}
}

@inproceedings{golovneva2022roscoe,
  title={{ROSCOE}: A Suite of Metrics for Scoring Step-by-Step Reasoning},
  author={Golovneva, Olga and Chen, Moya and Poff, Spencer and Corredor, Martin and Zettlemoyer, Luke and Fazel-Zarandi, Maryam and Celikyilmaz, Asli},
  booktitle={Proceedings of the Eleventh International Conference on Learning Representations},
  series={ICLR~'23},
  address={Kigali, Rwanda},
  year={2023},
  url={https://openreview.net/forum?id=xYlJRpzZtsY}
}

@inproceedings{xia2024reasoneval,
  title={Evaluating Mathematical Reasoning Beyond Accuracy},
  author={Xia, Shijie and Li, Xuefeng and Liu, Yixin and Wu, Tongshuang and Liu, Pengfei},
  booktitle={Proceedings of the AAAI Conference on Artificial Intelligence},
  series={AAAI~'25},
  address={Philadelphia, PA, USA},
  publisher={AAAI Press},
  volume={39},
  number={26},
  pages={27723--27730},
  year={2025},
  doi={10.1609/aaai.v39i26.34987},
  url={https://ojs.aaai.org/index.php/AAAI/article/view/34987}
}

@inproceedings{song2025prmbench,
  title={{PRMBench}: A Fine-grained and Challenging Benchmark for Process-Level Reward Models},
  author={Song, Mingyang and Su, Zhaochen and Qu, Xiaoye and Zhou, Jiawei and Cheng, Yu},
  booktitle={Proceedings of the 63rd Annual Meeting of the Association for Computational Linguistics (Volume 1: Long Papers)},
  series={ACL~'25},
  address={Vienna, Austria},
  publisher={Association for Computational Linguistics},
  pages={25299--25346},
  year={2025},
  doi={10.18653/v1/2025.acl-long.1230},
  url={https://aclanthology.org/2025.acl-long.1230/}
}

@inproceedings{qiu2025phybench,
  title={{PHYBench}: Holistic Evaluation of Physical Perception and Reasoning in Large Language Models},
  author={Qiu, Shi and Guo, Shaoyang and Song, Zhuo-Yang and Sun, Yunbo and Cai, Zeyu and Wei, Jiashen and Luo, Tianyu and Yin, Yixuan and Zhang, Haoxu and Hu, Yi and Wang, Chenyang and Tang, Chencheng and Chang, Haoling and Liu, Qi and Zhou, Ziheng and Zhang, Tianyu and Zhang, Jingtian and Liu, Zhangyi and Li, Minghao and Zhang, Yuku and Jing, Boxuan and Yin, Xianqi and Ren, Yutong and Fu, Zizhuo and Ji, Jiaming and Wang, Weike and Tian, Xudong and Lv, Anqi and Man, Laifu and Li, Jianxiang and Tao, Feiyu and Sun, Qihua and Liang, Zhou and Mu, Yushu and Li, Zhongxuan and Zhang, Jing-Jun and Zhang, Shutao and Li, Xiaotian and Xia, Xingqi and Lin, Jiawei and Shen, Zheyu and Chen, Jiahang and Xiong, Qiuhao and Wang, Binran and Wang, Fengyuan and Ni, Ziyang and Zhang, Bohan and Cui, Fan and Shao, Changkun and Cao, Qing-Hong and Luo, Ming-xing and Yang, Yaodong and Zhang, Muhan and Zhu, Hua Xing},
  booktitle={Advances in Neural Information Processing Systems: Datasets and Benchmarks Track},
  series={NeurIPS~'25},
  address={San Diego, CA, USA},
  publisher={Curran Associates, Inc.},
  volume={38},
  pages={1345--1385},
  year={2025},
  doi={10.52202/085713-0044},
  url={https://proceedings.neurips.cc/paper_files/paper/2025/hash/01e793f8cf5689aa0ea46a1b01071bea-Abstract-Datasets_and_Benchmarks_Track.html}
}

@misc{ncees2020fechemical,
  title={Fundamentals of Engineering ({FE}) Chemical {CBT} Exam Specifications},
  author={{National Council of Examiners for Engineering and Surveying}},
  note={Effective beginning with the July 2020 examinations},
  year={2020},
  url={https://ncees.org/wp-content/uploads/2022/09/FE-Che-CBT-specs.pdf}
}

@misc{ncees2020fecivil,
  title={Fundamentals of Engineering ({FE}) Civil {CBT} Exam Specifications},
  author={{National Council of Examiners for Engineering and Surveying}},
  note={Effective beginning with the July 2020 examinations},
  year={2020},
  url={https://ncees.org/wp-content/uploads/2022/09/FE-Civil-CBT-specs-1.pdf}
}

@misc{ncees2020feelectrical,
  title={Fundamentals of Engineering ({FE}) Electrical and Computer {CBT} Exam Specifications},
  author={{National Council of Examiners for Engineering and Surveying}},
  note={Effective beginning with the July 2020 examinations},
  year={2020},
  url={https://ncees.org/wp-content/uploads/FE-Electrical-and-Computer-CBT-specs.pdf}
}

@misc{ncees2020feindustrial,
  title={Fundamentals of Engineering ({FE}) Industrial and Systems {CBT} Exam Specifications},
  author={{National Council of Examiners for Engineering and Surveying}},
  note={Effective beginning with the July 2020 examinations},
  year={2020},
  url={https://ncees.org/wp-content/uploads/FE-Industrial-and-Systems-CBT-specs.pdf}
}

@misc{ncees2020femechanical,
  title={Fundamentals of Engineering ({FE}) Mechanical {CBT} Exam Specifications},
  author={{National Council of Examiners for Engineering and Surveying}},
  note={Effective beginning with the July 2020 examinations},
  year={2020},
  url={https://ncees.org/wp-content/uploads/FE-Mechanical-CBT-specs.pdf}
}

@misc{mistral2025mistral3,
  title={Introducing {Mistral 3}},
  author={{Mistral AI}},
  year={2025},
  url={https://mistral.ai/news/mistral-3}
}

\newpage

\appendix

\appendix


\section{LLM Prompt for Domain Validation}


\label{sec:appendix_domain_validation}

We draw the candidate domains and areas of each branch from its curricular documents
(\autoref{subsec:taxonomy}) and the standard textbooks of~\autoref{tab:foundational_textbooks}.
A panel of three LLMs from families outside the evaluated models, \texttt{Grok 4.6}~\citep{grok46},
\texttt{MiniMax M3}~\citep{minimaxm3}, and \texttt{MiMo-V2.5-Pro}, each prompted individually with
the prompt below, names the cornerstone domains of each branch, and we finalize the 15 domains,
three per branch, by majority vote.
The same panel names the fundamental areas of each domain (\autoref{sec:appendix_area_validation})
and scores each area's pedagogical significance (\autoref{sec:appendix_significance_scoring}).

\begin{tcolorbox}[
    colback=gray!5,
    colframe=gray!40,
    boxrule=0.4pt,
    arc=2pt,
    left=6pt,
    right=6pt,
    top=6pt,
    bottom=6pt,
    breakable
]
\ttfamily\small
You are an expert academic and senior curriculum designer at a top-tier engineering university (like MIT or Caltech). You have decades of experience structuring undergraduate programs for ABET accreditation. \\

Your task is to identify and rank the most fundamental, cornerstone
domains for the five primary branches of engineering. The goal is to identify the non-overlapping, core subject areas that are prerequisites for almost all other advanced topics. \\

For example, for Chemical Engineering, "Reaction Kinetics" would be a fundamental domain, while a niche topic like "Polymer Rheology" would be a sub-specialization, not a cornerstone. \\

Please provide a ranked list of the top 5-7 most fundamental domains for each of the following five branches: \\
1. Chemical Engineering \\
2. Electrical Engineering \\
3. Mechanical Engineering \\
4. Civil Engineering \\
5. Industrial Engineering \\

Return your answer \textbf{only} in a strict JSON format as given below.

\vspace{8pt}
\noindent\bgroup\ttfamily
\texttt{\char`\{}

\vspace{8pt}
\hspace*{1em}"chemical\_engineering": [\\
\hspace*{2em}"Rank 1 Domain",\\
\hspace*{2em}"Rank 2 Domain",\\
\hspace*{2em}"Rank 3 Domain",\\
\hspace*{2em}"Rank 4 Domain",\\
\hspace*{2em}"Rank 5 Domain"\\
\hspace*{1em}],\\

\vspace{2pt}
\hspace*{1em}"electrical\_engineering": [\\
\hspace*{2em}"Rank 1 Domain",\\
\hspace*{2em}"Rank 2 Domain",\\
\hspace*{2em}"Rank 3 Domain",\\
\hspace*{2em}"Rank 4 Domain",\\
\hspace*{2em}"Rank 5 Domain"\\
\hspace*{1em}],\\

\vspace{2pt}
\hspace*{1em}"mechanical\_engineering": [\\
\hspace*{2em}"Rank 1 Domain",\\
\hspace*{2em}"Rank 2 Domain",\\
\hspace*{2em}"Rank 3 Domain",\\
\hspace*{2em}"Rank 4 Domain",\\
\hspace*{2em}"Rank 5 Domain"\\
\hspace*{1em}],\\

\vspace{2pt}
\hspace*{1em}"civil\_engineering": [\\
\hspace*{2em}"Rank 1 Domain",\\
\hspace*{2em}"Rank 2 Domain",\\
\hspace*{2em}"Rank 3 Domain",\\
\hspace*{2em}"Rank 4 Domain",\\
\hspace*{2em}"Rank 5 Domain"\\
\hspace*{1em}],\\

\vspace{2pt}
\hspace*{1em}"industrial\_engineering": [\\
\hspace*{2em}"Rank 1 Domain",\\
\hspace*{2em}"Rank 2 Domain",\\
\hspace*{2em}"Rank 3 Domain",\\
\hspace*{2em}"Rank 4 Domain",\\
\hspace*{2em}"Rank 5 Domain"\\
\hspace*{1em}]\\

\vspace{2pt}
\texttt{\char`\}}
\egroup
\vspace{8pt}

Do not include any preamble, conversational text, explanations, or markdown formatting around the JSON block.
\end{tcolorbox}

We keep three domains per branch to cover each in depth, across all its areas and all three levels
(\autoref{tab:area}).
Domains outside this selection, such as circuit analysis in electrical engineering, thermodynamics
and heat transfer in mechanical engineering, and transportation and environmental engineering in
civil engineering, can be added with templates of the same form.

\section{LLM Prompt for Area Validation}
\label{sec:appendix_area_validation}

The panel of~\autoref{sec:appendix_domain_validation} answers the prompt below for each domain, and
we finalize the 42 areas by majority vote, as for the domains.

\begin{tcolorbox}[
    colback=gray!5,
    colframe=gray!40,
    boxrule=0.4pt,
    arc=2pt,
    left=6pt,
    right=6pt,
    top=6pt,
    bottom=6pt,
    breakable
]
\ttfamily\small
You are an expert academic and senior curriculum designer at a top-tier engineering university (like MIT or Caltech). You have decades of experience structuring undergraduate programs for ABET accreditation. \\

Your task is to identify and list the most fundamental, non-overlapping "areas" or
"sub-topics" within the provided engineering domain. \\

A "fundamental area" is a major pedagogical unit or chapter within a core course on this domain. \textbf{The areas you list should be distinct pillars of the subject.} For example, if the domain is
"Thermodynamics," a fundamental area would be "The First Law" or
"Properties of Pure Fluids." A concept like "Enthalpy" would be
too specific. Similarly, within a "Circuit Analysis" domain, "Kirchhoff's Voltage Law" and "Kirchhoff's Current Law" are foundational concepts often taught together and would be too granular to list separately. \\

Please provide a list of the top 3-5 most fundamental areas for
the following engineering domain:
\texttt{[DOMAIN\_NAME\_HERE]} \\

Return your answer \textbf{only} in a strict JSON format as given below.

\vspace{8pt}
\noindent\bgroup\ttfamily
\texttt{\char`\{}

\vspace{4pt}
\hspace*{1em}"domain": "[DOMAIN\_NAME\_HERE]",\\
\hspace*{1em}"areas": [\\
\hspace*{2em}"Area 1",\\
\hspace*{2em}"Area 2",\\
\hspace*{2em}"Area 3",\\
\hspace*{2em}"Area 4"\\
\hspace*{1em}]\\
\texttt{\char`\}}
\egroup
\vspace{8pt}

Do not include any preamble, conversational text, explanations, or markdown formatting around the JSON block.
\end{tcolorbox}

\section{Textbooks and Data Sources}
\label{sec:appendix_books}
\label{sec:appendix_data_sources}

The domain experts who write the templates take their principles, problem types, and parameter
constraints from the standard textbooks of~\autoref{tab:foundational_textbooks}, one or two per
domain, which are also among the sources of the domains and areas
(\autoref{sec:appendix_domain_validation}).

\begin{table}[t]
\small
\centering
\renewcommand{\arraystretch}{1.1}
\setlength{\tabcolsep}{6pt}
\begin{tabular}{p{0.35\linewidth} p{0.55\linewidth}}
\toprule
\rowcolor{gray!10}\textbf{Domain} & \textbf{Textbook} \\
\midrule
Reaction Kinetics & Elements of Chemical Reaction Engineering by H. Scott Fogler \\
Thermodynamics & Introduction to Chemical Engineering Thermodynamics by J. M. Smith, H. C. Van Ness,
and M. M. Abbott \\
Transport Phenomena & Transport Phenomena by R. Byron Bird, Warren E. Stewart, and Edwin N.
Lightfoot \\
\addlinespace[2pt]
Electromagnetics and Waves & Fundamentals of Applied Electromagnetics by F. T. Ulaby and Umberto
Ravaioli \\
Digital Communications & Digital Communications by John G. Proakis and Masoud Salehi \\
Signals and Systems & Discrete-Time Signal Processing by Alan V. Oppenheim and Ronald W. Schafer \\
\addlinespace[2pt]
Mechanics of Materials & Mechanics of Materials by Ferdinand P. Beer, E. Russell Johnston, Jr., and
John T. DeWolf \\
Fluid Mechanics & Fundamentals of Fluid Mechanics by Bruce R. Munson, Donald F. Young, and Theodore
H. Okiishi \\
Vibrations and Acoustics & Mechanical Vibrations by Singiresu S. Rao \\
\addlinespace[2pt]
Structural Analysis & Structural Analysis by R. C. Hibbeler \\
Geotechnical Engineering & Principles of Geotechnical Engineering by Braja M. Das and Khaled
Sobhan \\
Water Resources & Open Channel Hydraulics by Terry W. Sturm \newline
Applied Hydrology by Ven Te Chow, David R. Maidment, and Larry W. Mays \\
\addlinespace[2pt]
Production and Inventory & Production and Operations Analysis by Steven Nahmias and Tava Lennon
Olsen \\
Quality and Reliability Control & Introduction to Statistical Quality Control by Douglas C.
Montgomery \\
Stochastic Operations & Introduction to Probability Models by Sheldon M. Ross \newline
Introduction to Operations Research by Frederick S. Hillier and Gerald J. Lieberman \\
\bottomrule
\end{tabular}
\caption{\textbf{Foundational textbooks.} The textbooks of each domain, from which the templates
take their problems.}
\label{tab:foundational_textbooks}
\end{table}

The templates take their physical constants and material properties from the handbooks and data sources listed in~\autoref{tab:authoritative_sources}.
Each table of constants in the template code names its source.

\begin{table}[t]
\small
\centering
\renewcommand{\arraystretch}{1.1}
\setlength{\tabcolsep}{6pt}
\begin{tabular}{p{0.35\linewidth} p{0.50\linewidth}}
\toprule
\rowcolor{gray!10}\textbf{Engineering Branch} & \textbf{Authoritative Data Sources} \\
\midrule
Chemical Engineering &
Perry's Chemical Engineers' Handbook \newline
NIST Chemistry WebBook \newline
CRC Handbook of Chemistry and Physics \newline
NIST-JANAF Thermochemical Tables \newline
NASA Technical Report R-132 \\
\addlinespace[2pt]
Electrical Engineering &
Standard Handbook for Electrical Engineers \newline
CODATA 2022 Recommended Values \newline
refractiveindex.info Database \\
\addlinespace[2pt]
Mechanical Engineering &
ASM Handbook, Volume 1 \newline
Marks' Standard Handbook for Mechanical Engineers \newline
Shigley's Mechanical Engineering Design \newline
MIL-HDBK-5J \newline
USDA Wood Handbook \newline
NIST Chemistry WebBook \newline
PubChem \newline
NIST SP 811 \\
\addlinespace[2pt]
Civil Engineering &
NAVFAC DM-7.01 and DM-7.02 \newline
FHWA HDS-4 and HEC-22 \newline
NRCS TR-55 \newline
USGS Water-Supply Paper 2339 \newline
AISC Shapes Database v16.0 \newline
NIST Chemistry WebBook \newline
CODATA 2022 Recommended Values \newline
NIST SP 811 \\
\addlinespace[2pt]
Industrial Engineering &
MIL-STD-105E \newline
NIST/SEMATECH e-Handbook of Statistical Methods \\
\bottomrule
\end{tabular}
\caption{\textbf{Authoritative data sources.} The sources of the constants and material properties
the templates sample, by branch.}
\label{tab:authoritative_sources}
\end{table}

\section{LLM Prompt for Pedagogical Scoring}
\label{sec:appendix_significance_scoring}

The panel of~\autoref{sec:appendix_domain_validation} scores the pedagogical significance of each
area within its domain with the prompt below.

\begin{tcolorbox}[
    colback=gray!5,
    colframe=gray!30,
    boxrule=0.4pt,
    arc=2pt,
    left=6pt,
    right=6pt,
    top=6pt,
    bottom=6pt,
    breakable
]
\ttfamily\small
You are an expert academic and senior curriculum designer at a top-tier engineering university with decades of experience structuring undergraduate programs for ABET accreditation. \\

Your task is to evaluate the pedagogical significance of a specific engineering "area" within a broader "domain" on a scale of 1 to 5, based on the rubric below. \\

\textbf{Rubric:}
\par\medskip\noindent
\textbf{Score 5 (Cornerstone):} A
foundational, prerequisite topic for nearly all other concepts in the domain.
\par\medskip\noindent
\textbf{Score 3 (Core Concept):} A standard, important topic that builds upon cornerstone principles but is not necessarily a universal prerequisite.
\par\medskip\noindent
\textbf{Score 1 (Specialized Application):} A focused application or an
integrative topic taught after cornerstones and core concepts are mastered.
\par\medskip

Please evaluate the following area:
\par\medskip\noindent
\textbf{Domain:} [DOMAIN\_NAME\_HERE]
\par\noindent
\textbf{Area:} [AREA\_NAME\_HERE]
\par\bigskip

\noindent\bgroup\ttfamily
\texttt{\char`\{}

\par\medskip
\hspace*{1em}"area": "[AREA\_NAME\_HERE]",\\
\hspace*{1em}"pedagogical\_significance\_score": <integer, a score from 1 to 5>,\\
\hspace*{1em}"justification": <string, a one-sentence explanation for the score>.\\
\vspace{2pt}
\texttt{\char`\}}
\egroup
\par\bigskip

Do not include any preamble, conversational text, explanations, or markdown formatting around the JSON block.
\end{tcolorbox}

\section{Difficulty Labeling}
\label{sec:appendix_difficulty}

Three domain experts of the template's branch set its difficulty level; one is its author and the
other two did not write it.
Each expert first scores the template on their own, from 1 to 3 on conceptual complexity,
mathematical sophistication, and procedural depth, counting only what the solver must supply: a
formula or method that the question states does not count.
Each dimension takes the median of the three scores, and the sum of the three medians sets the
level: 3--4 Easy, 5--6 Intermediate, and 7--9 Advanced.
When a template's instances follow several reasoning paths, the experts score each path, and the
template takes the median level of its paths.
After the experts have scored, the panel of~\autoref{sec:appendix_domain_validation} rates every
template with the prompt below as a second check; a template whose panel level differs from the
experts' level goes back to its three experts for review, and the panel sets no level itself.
The experts fix the levels before any evaluated model runs and never change them on the basis of
evaluation results.

\begin{tcolorbox}[
    colback=gray!5,
    colframe=gray!40,
    boxrule=0.4pt,
    arc=2pt,
    left=6pt,
    right=6pt,
    top=6pt,
    bottom=6pt,
    breakable
]
\ttfamily\small
You are an expert academic and senior engineering educator at a top-tier engineering university (like MIT or Caltech). You have decades of experience designing undergraduate courses and examinations for ABET-accredited programs. \\

Your task is to rate the difficulty of an engineering problem template for a student who has just completed the core undergraduate course in its domain. Rate only what the student must supply: if the question states a formula, method, or principle, do not count it. \\

Rate the template on three dimensions, each on a scale of 1 to 3: \\

Conceptual complexity: \\
1 = one principle, applied directly \\
2 = several principles from one domain, or one modelling choice (e.g., flow regime, configuration, control volume) \\
3 = principles from several domains, or assumptions the student must choose and justify \\

Mathematical sophistication: \\
1 = direct substitution into formulas (arithmetic, powers, roots, logarithms) \\
2 = rearrangement, simultaneous or nonlinear equations, or probability beyond a single formula \\
3 = calculus, differential equations, or an iterative or numerical solution \\

Procedural depth: \\
1 = 1-2 dependent intermediate quantities \\
2 = 3-5 dependent intermediate quantities \\
3 = 6 or more dependent intermediate quantities \\

Sum the three scores and assign a level: 3-4 = Easy, 5-6 = Intermediate, 7-9 = Advanced. \\

If the instances follow different reasoning paths, rate each path separately. \\

Please evaluate the following template: \\
Domain: [DOMAIN\_NAME\_HERE] \\
Area: [AREA\_NAME\_HERE] \\
Template source code: [TEMPLATE\_CODE\_HERE] \\
Instances (one per reasoning path): [INSTANCES\_HERE] \\

Return your answer only in a strict JSON format as given below.

\vspace{8pt}
\noindent\bgroup\ttfamily
\texttt{\char`\{}

\vspace{4pt}
\hspace*{1em}"template": "[TEMPLATE\_NAME\_HERE]",\\
\hspace*{1em}"paths": [\\
\hspace*{2em}\texttt{\char`\{}\\
\hspace*{3em}"path": "<short description of the reasoning path>",\\
\hspace*{3em}"conceptual\_complexity": <integer 1-3>,\\
\hspace*{3em}"mathematical\_sophistication": <integer 1-3>,\\
\hspace*{3em}"procedural\_depth": <integer 1-3>,\\
\hspace*{3em}"level": "<Easy | Intermediate | Advanced>",\\
\hspace*{3em}"justification": "<one sentence explaining the scores>"\\
\hspace*{2em}\texttt{\char`\}}\\
\hspace*{1em}]\\
\texttt{\char`\}}
\egroup
\vspace{8pt}

Do not include any preamble, conversational text, explanations, or markdown formatting around the JSON block.
\end{tcolorbox}

\section{Dataset Statistics}
\label{appendix:statistics}

This section gives the statistics of the templates and the evaluation set
behind~\autoref{subsec:taxonomy} and~\autoref{subsec:templates}.

\paragraph{Structural Balance and Granularity.} Each branch holds 30 templates, so every branch
weighs the same in the aggregate measures.
The number of areas differs by branch (\autoref{tab:difficulty_stats}): Mechanical Engineering
groups its topics into six broad areas, Chemical and Electrical Engineering have seven, and Civil
and Industrial Engineering have eleven.
\autoref{tab:area} gives the templates of every area by level; the number of templates per area
follows the pedagogical significance the panel assigns to the area
(\autoref{sec:appendix_significance_scoring}).

\begin{table*}[t]
\centering
\footnotesize
\renewcommand{\arraystretch}{1.05}
\setlength{\tabcolsep}{3pt}
\begin{minipage}[t]{0.48\textwidth}
\centering
\begin{tabular}{@{}p{118pt} r r r r@{}}
\toprule
\rowcolor{gray!10}\textbf{Domain and area} & \textbf{Easy} & \textbf{Int.} & \textbf{Adv.} & \textbf{Total} \\
\midrule
\rowcolor{gray!10}\textbf{Chemical} (7 areas) & \textbf{13} & \textbf{9} & \textbf{8} & \textbf{30} \\
\multicolumn{5}{@{}l}{\textit{Reaction Kinetics}} \\
\quad Conversion and Reactor Sizing & 0 & 0 & 1 & 1 \\
\quad Mole Balances & 2 & 1 & 2 & 5 \\
\quad Stoichiometry & 1 & 2 & 1 & 4 \\
\multicolumn{5}{@{}l}{\textit{Thermodynamics}} \\
\quad Heat Effects & 3 & 1 & 1 & 5 \\
\quad Volumetric Properties of Pure Fluids & 3 & 2 & 2 & 7 \\
\multicolumn{5}{@{}l}{\textit{Transport Phenomena}} \\
\quad Shell Momentum Balances & 2 & 1 & 0 & 3 \\
\quad Viscosity and Momentum Transport & 2 & 2 & 1 & 5 \\
\rowcolor{gray!10}\textbf{Electrical} (7 areas) & \textbf{11} & \textbf{12} & \textbf{7} & \textbf{30} \\
\multicolumn{5}{@{}l}{\textit{Digital Communications}} \\
\quad Deterministic and Random Signal Analysis & 2 & 2 & 1 & 5 \\
\quad Digital Modulation Schemes & 1 & 2 & 2 & 5 \\
\multicolumn{5}{@{}l}{\textit{Electromagnetics and Waves}} \\
\quad Electrostatics & 1 & 2 & 1 & 4 \\
\quad Magnetostatics & 1 & 0 & 0 & 1 \\
\quad Waves and Phasors & 2 & 2 & 1 & 5 \\
\multicolumn{5}{@{}l}{\textit{Signals and Systems}} \\
\quad Continuous Time Signals & 2 & 2 & 1 & 5 \\
\quad Discrete Time Signals & 2 & 2 & 1 & 5 \\
\rowcolor{gray!10}\textbf{Civil} (11 areas) & \textbf{12} & \textbf{12} & \textbf{6} & \textbf{30} \\
\multicolumn{5}{@{}l}{\textit{Geotechnical Engineering}} \\
\quad Permeability, Seepage, and Effective Stress & 1 & 2 & 0 & 3 \\
\quad Phase and Index Properties & 2 & 1 & 0 & 3 \\
\quad Strength and Stability & 1 & 1 & 0 & 2 \\
\quad Stress Distribution and Consolidation & 0 & 0 & 2 & 2 \\
\multicolumn{5}{@{}l}{\textit{Structural Analysis}} \\
\quad Deflections & 1 & 2 & 0 & 3 \\
\quad Determinate Structures & 3 & 1 & 0 & 4 \\
\quad Indeterminate Analysis & 0 & 0 & 2 & 2 \\
\quad Influence Lines & 0 & 1 & 0 & 1 \\
\multicolumn{5}{@{}l}{\textit{Water Resources}} \\
\quad Energy and Rapidly Varied Flow & 1 & 2 & 0 & 3 \\
\quad Hydrology & 1 & 1 & 1 & 3 \\
\quad Uniform Flow & 2 & 1 & 1 & 4 \\
\bottomrule
\end{tabular}
\end{minipage}
\hfill
\begin{minipage}[t]{0.48\textwidth}
\centering
\begin{tabular}{@{}p{118pt} r r r r@{}}
\toprule
\rowcolor{gray!10}\textbf{Domain and area} & \textbf{Easy} & \textbf{Int.} & \textbf{Adv.} & \textbf{Total} \\
\midrule
\rowcolor{gray!10}\textbf{Industrial} (11 areas) & \textbf{12} & \textbf{12} & \textbf{6} & \textbf{30} \\
\multicolumn{5}{@{}l}{\textit{Production and Inventory}} \\
\quad Deterministic Lot Sizing & 2 & 2 & 0 & 4 \\
\quad Production Planning & 1 & 0 & 2 & 3 \\
\quad Stochastic Inventory & 1 & 2 & 0 & 3 \\
\multicolumn{5}{@{}l}{\textit{Quality and Reliability Control}} \\
\quad Acceptance Sampling & 0 & 1 & 1 & 2 \\
\quad Attributes Control Charts & 1 & 1 & 0 & 2 \\
\quad Process Capability & 1 & 1 & 0 & 2 \\
\quad Variables Control Charts & 2 & 1 & 1 & 4 \\
\multicolumn{5}{@{}l}{\textit{Stochastic Operations}} \\
\quad Markov Chains & 1 & 1 & 1 & 3 \\
\quad Poisson Processes & 1 & 0 & 0 & 1 \\
\quad Queueing Systems & 1 & 2 & 1 & 4 \\
\quad System Reliability & 1 & 1 & 0 & 2 \\
\rowcolor{gray!10}\textbf{Mechanical} (6 areas) & \textbf{10} & \textbf{13} & \textbf{7} & \textbf{30} \\
\multicolumn{5}{@{}l}{\textit{Fluid Mechanics}} \\
\quad Fluid Kinematics & 1 & 2 & 2 & 5 \\
\quad Fluid Statics & 2 & 2 & 1 & 5 \\
\multicolumn{5}{@{}l}{\textit{Mechanics of Materials}} \\
\quad Stress and Strain & 2 & 2 & 1 & 5 \\
\quad Torsion & 2 & 2 & 1 & 5 \\
\multicolumn{5}{@{}l}{\textit{Vibrations and Acoustics}} \\
\quad Harmonically Excited Vibrations & 1 & 2 & 1 & 4 \\
\quad Single Degree of Freedom Systems & 2 & 3 & 1 & 6 \\
\midrule
\rowcolor{gray!10}\textbf{Total} & \textbf{58} & \textbf{58} & \textbf{34} & \textbf{150} \\
\bottomrule
\end{tabular}
\end{minipage}
\caption{\textbf{Templates per area and level.} 42 areas in 15 domains; each branch has 30
templates; the shaded rows give the branch subtotals and the italic rows name the domain.}
\label{tab:area}
\end{table*}

\paragraph{Difficulty Distribution.} The domain experts' levels (\autoref{subsec:taxonomy}) give 58
Easy, 58 Intermediate, and 34 Advanced templates, so 77\% of the templates are Easy or Intermediate
and 23\% Advanced.
Each template contributes 15 instances to the evaluation set, which therefore holds 870 Easy, 870
Intermediate, and 510 Advanced instances, 450 per branch.
\autoref{tab:difficulty_stats} gives the levels by branch.

\begin{table}[t]
    \centering
    \small
    \setlength{\tabcolsep}{3.5pt}
    \begin{tabular}{l|cc|ccc|c}
        \toprule
        \rowcolor{gray!10}
        \multirow{2}{*}{\textbf{Branch}} & \multicolumn{2}{c|}{\textbf{Structure}} & \multicolumn{3}{c|}{\textbf{Difficulty Level}} & \multirow{2}{*}{\textbf{Total}} \\
        & \textbf{Dom.} & \textbf{Areas} & \textbf{Easy} & \textbf{Int.} & \textbf{Adv.} & \\
        \midrule
        Chemical & 3 & 7 & 13 & 9 & 8 & \textbf{30} \\
        Electrical & 3 & 7 & 11 & 12 & 7 & \textbf{30} \\
        Civil & 3 & 11 & 12 & 12 & 6 & \textbf{30} \\
        Industrial & 3 & 11 & 12 & 12 & 6 & \textbf{30} \\
        Mechanical & 3 & 6 & 10 & 13 & 7 & \textbf{30} \\
        \midrule
        \textbf{Total} & \textbf{15} & \textbf{42} & \textbf{58} & \textbf{58} & \textbf{34} & \textbf{150} \\
        \bottomrule
    \end{tabular}
    \caption{\textbf{Dataset statistics by branch.} Domains, areas, and templates per difficulty
    level; each branch holds 30 templates.}
    \label{tab:difficulty_stats}
\end{table}

\paragraph{Answer Kinds.}
The final answers are of six kinds; by template, 89 scalar, 32 multipart, 9 symbolic, 8 vector,
7 array, and 5 classification.

\paragraph{Variation Within a Template.}
We measure how far the 15 evaluation-set instances of a template differ in three respects, each with
numbers masked (\autoref{tab:template_variation}): the question wording, in which a different
material or fluid counts as a different wording; the reasoning path, that is, the lines of the gold
trace that contain an equation, in order, with the names that vary between questions also masked;
and the form of the answer.
In 92 of the 150 templates, the instances follow two or more reasoning paths, and the other 58
follow a single path (chemical 18, electrical 5, civil 14, industrial 10, mechanical 11).
The same 92 templates show more than one path in 500 further draws, so the evaluation set misses no
template's second path; counting a name change inside an equation as a different path raises the
count to 96 templates.
Two instances of a template are near-duplicates when their questions share the wording and every
number lies within 5\% of its counterpart; the evaluation set holds 61 such pairs, in 14 templates.

\begin{table}[t]
    \centering
    \small
    \setlength{\tabcolsep}{6pt}
    \begin{tabular}{c|ccc}
        \toprule
        \rowcolor{gray!10}
        \textbf{Variants} & \textbf{Wording} & \textbf{Reasoning Paths} & \textbf{Answer Forms} \\
        \midrule
        1 & 38 & 58 & 104 \\
        2 & 17 & 31 & 25 \\
        3--5 & 34 & 25 & 14 \\
        6--15 & 61 & 36 & 7 \\
        \bottomrule
    \end{tabular}
    \caption{\textbf{Variation within a template.} Number of templates whose 15 evaluation-set
    instances show the given number of distinct question wordings, reasoning paths, and answer
    forms, with numbers masked; each column sums to 150.}
    \label{tab:template_variation}
\end{table}

\paragraph{Drawing the Evaluation Set.}
For each template, we generate its first 100 draws from the private 128-bit seed, 15,000 draws in
all, and skip 549 repeated questions, 35 rounding ties (instances whose gold trace prints a value
lying exactly halfway between two roundings), and one question already used to validate the
evaluator, so 14,415 remain.
We group the remaining draws of a template by reasoning path and answer form and take the 15
instances round-robin across the groups, in order of first appearance, so that no group is repeated
before every group has been used.
101 templates have more than one group, and 18 templates have more groups than their 15 instances
can cover.
All 2,250 questions are distinct.
We publish the SHA-256 hash of every instance and of the seed, and will release the seed at
publication so that anyone can regenerate the evaluation set and check it against the hashes.

\section{Domain-Aware Parameterization}
\label{sec:appendix_param_details}

Templates constrain the values they sample in four ways, shown below with examples from all five
branches.~\autoref{tab:authoritative_sources} lists the sources of the tabulated values.

\paragraph{Tabulated properties.} A template selects a substance, material, section or scenario
from a curated table and takes the properties the problem needs from that entry, so that they are
mutually consistent.
Examples are the critical properties, heats of formation and heat-capacity coefficients
($C_p/R = A + BT + CT^2 + DT^{-2}$) of chemical species; the Young's modulus, Poisson's ratio and
shear modulus of materials such as 6061-T6 aluminum, in SI or US customary units; the properties
that set the phase velocity in media such as fused silica; the depth, area and moment of inertia of
AISC wide-flange sections; the ranges of permeability and friction angle for each soil type; and the
ranges of arrival and service rates for each queueing scenario.

\paragraph{Laws, correlations and standards.} Other values follow the relations and design tables
used in practice.
Reaction problems use balanced chemical equations, compression indices follow Skempton's
correlation with the liquid limit, and bearing-capacity factors are Terzaghi's values for the
sampled friction angle.
Manning's coefficients follow the channel lining, runoff coefficients the land use, and SCS curve
numbers the land use and hydrologic soil group.
Acceptance-sampling plans come from the tables of MIL-STD-105E, control-chart factors from their
statistical definitions, and safety factors from standard normal quantiles at typical
cycle-service levels.

\paragraph{Regime-aware ranges.}
Templates set sampling ranges relative to the quantity that
decides the regime, so that each instance stays in the regime its problem assumes or lands in the
one it tests.
Sampling frequencies are drawn relative to the Nyquist rate, to model ideal conversion or to induce
aliasing, and forcing frequencies relative to the natural frequency, so that instances fall below,
near and above resonance.
Channel slopes keep the flow in the regime the problem assumes, such as subcritical uniform flow;
beam loads keep deflections within serviceability limits without overstressing the section; queue
utilizations stay below one, so that a steady state exists; and production rates are 1.5 to 6 times
the demand rate, so that finite-rate production is feasible.
Some templates also check each drawn scenario and redraw it when it leaves the regime: the
falling-film template, for example, redraws until the film Reynolds number is below 1,500 and the
maximum velocity below 10\,m/s.

\paragraph{Exact values.} Templates draw some values from structured sets, so that they have the
form the problem requires and the arithmetic stays exact.
Probabilities of discrete random variables sum to exactly one as printed, rows of Markov transition
matrices sum exactly to one, modulation orders are powers of two, carrier frequencies are integer
multiples of the bit rate, and discrete-time frequencies are rational multiples of $\pi$.

\section{Template Implementation Examples}
\label{sec:appendix_template_examples}

The following examples illustrate both parametric and \textit{structural} diversity across
templates.
Structural variation refers to differences in the type of physical reasoning chain required, not
merely in numerical values.
The Easy template (\texttt{Template Rational Method Peak Flow}, civil engineering) applies the
rational formula directly to a homogeneous catchment, or first computes an area-weighted runoff
coefficient when the catchment combines two land uses.
The Intermediate template (\texttt{Template Exponential MTTF Topology}, industrial engineering)
samples the system topology: for pumps in series, the lifetime of the system is the minimum of
independent exponential lifetimes, whereas for two pumps in parallel, the memoryless property splits
the lifetime into two phases.
The Advanced template (\texttt{Template Equivalent Stiffness Frequency}, mechanical engineering)
reduces a series or parallel spring arrangement to an equivalent stiffness before computing the
natural frequency.
In each of the three, a sampled parameter changes the reasoning chain, not only the numbers.
The listings show the template functions without their documentation strings.

\vspace{6pt}

\subsection{Easy}

\vspace{6pt}

\begin{lstlisting}[style=pythonstyle]
def template_rational_method_peak_flow():
    form = random.choice(["single", "composite"])

    if form == "single":
        use = random.choice(list(_RATIONAL_USES.keys()))
        phrase = _RATIONAL_USES[use]
        c_lo, c_hi = RATIONAL_C[use]
        C = round(random.uniform(c_lo, c_hi), 2)
        A = round(random.uniform(max(2.0, 0.52 / C), 20.0), 1)
        i = random.randint(max(25, math.ceil(43.2 / (C * A))), 90)
        Q = round(C * i * A / 360.0, 2)
        setup_text = (
            f"a {A:.1f} ha catchment consisting of a {phrase} with a "
            f"runoff coefficient of C = {C:.2f} (per HEC-22)")
        steps = (
            f"**Step 1:** State the rational formula in SI form.\n"
            f"Q = C * i * A / 360, with Q in m^3/s, i in mm/h, and A in "
            f"ha.\n\n"
            f"**Step 2:** Substitute the given values.\n"
            f"Q = {C:.2f} * {i} * {A:.1f} / 360\n\n"
            f"**Step 3:** Evaluate the peak discharge.\n"
            f"Q = {Q:.2f} m^3/s\n\n"
        )
    else:
        use1, use2 = random.sample(list(_RATIONAL_USES.keys()), 2)
        p1, p2 = _RATIONAL_USES[use1], _RATIONAL_USES[use2]
        c1 = round(random.uniform(*RATIONAL_C[use1]), 2)
        c2 = round(random.uniform(*RATIONAL_C[use2]), 2)
        A1 = round(random.uniform(3.0, 12.0), 1)
        A2 = round(random.uniform(3.0, 12.0), 1)
        A = round(A1 + A2, 1)
        Cw = round((c1 * A1 + c2 * A2) / A, 3)
        i = random.randint(max(25, math.ceil(43.2 / (Cw * A))), 90)
        Q = round(Cw * i * A / 360.0, 2)
        C = Cw
        setup_text = (
            f"a catchment with two land uses: {A1:.1f} ha of {p1} "
            f"(C = {c1:.2f}) and {A2:.1f} ha of {p2} (C = {c2:.2f}), the "
            f"runoff coefficients taken from HEC-22 Table 3-1")
        steps = (
            f"**Step 1:** Compute the area-weighted composite runoff "
            f"coefficient.\n"
            f"C_w = (C1*A1 + C2*A2) / (A1 + A2) "
            f"= ({c1:.2f} * {A1:.1f} + {c2:.2f} * {A2:.1f}) / {A:.1f} "
            f"= {Cw:.3f}\n\n"
            f"**Step 2:** State the rational formula in SI form.\n"
            f"Q = C_w * i * A / 360, with Q in m^3/s, i in mm/h, and A "
            f"in ha.\n\n"
            f"**Step 3:** Substitute and evaluate.\n"
            f"Q = {Cw:.3f} * {i} * {A:.1f} / 360 = {Q:.2f} m^3/s\n\n"
        )

    assert 0.1 <= Q <= 6.0, f"peak flow out of bounds: {Q}"

    question = (
        f"During a design storm with rainfall intensity i = {i} mm/h, "
        f"determine the peak discharge in m^3/s from {setup_text}, using "
        f"the rational method in SI form (Q = C*i*A/360 with i in mm/h "
        f"and A in hectares)."
    )

    solution = (
        f"**Given:**\n"
        f"Rainfall intensity (i): {i} mm/h\n"
        f"Catchment: {setup_text}\n\n"
        f"{steps}"
        f"**Answer:** The peak discharge is {Q:.2f} m^3/s"
    )

    return question, solution
\end{lstlisting}

\vspace{6pt}

\subsection{Intermediate}

\vspace{6pt}

\begin{lstlisting}[style=pythonstyle]
def template_exponential_mttf_topology():
    configuration = random.choice(["series", "parallel"])

    if configuration == "series":
        lam1 = random.randint(*_T9_SERIES_RATE_CENTS) / 100
        lam2 = random.randint(*_T9_SERIES_RATE_CENTS) / 100
        lam3 = random.randint(*_T9_SERIES_RATE_CENTS) / 100
        lams = round(lam1 + lam2 + lam3, 2)          # exact 2-dp sum
        mttf = float((Decimal("1000") / Decimal(f"{lams:.2f}"))
                     .quantize(Decimal("0.1"), rounding=ROUND_HALF_UP))
        assert 333.0 <= mttf <= 1667.0, f"series MTTF out of bounds: {mttf}"

        question = (
            f"A booster station moves slurry through three pumps "
            f"connected in series along one line, so the station fails "
            f"as soon as ANY one pump fails. The pumps fail "
            f"independently, with exponentially distributed lifetimes "
            f"and constant failure rates of {lam1:.2f}, {lam2:.2f}, and "
            f"{lam3:.2f} failures per 1000 hours. Determine the mean "
            f"time to failure (MTTF -- the expected operating time until "
            f"the station first fails) of the station, in hours, to one "
            f"decimal place (round half up)."
        )
        solution = (
            f"**Given:**\n"
            f"Three pumps in series; independent exponential lifetimes "
            f"with failure rates lam1 = {lam1:.2f}, lam2 = {lam2:.2f}, "
            f"lam3 = {lam3:.2f} per 1000 hours.\n\n"
            f"**Step 1:** Characterize the station lifetime. The station "
            f"fails at the FIRST pump failure, i.e., at the minimum of "
            f"the three lifetimes; the minimum of independent "
            f"exponential lifetimes is exponential with the SUM of the "
            f"rates.\n\n"
            f"**Step 2:** Sum the failure rates.\n"
            f"lams = {lam1:.2f} + {lam2:.2f} + {lam3:.2f} = {lams:.2f} "
            f"failures per 1000 hours\n\n"
            f"**Step 3:** Invert the rate and convert to hours. A rate "
            f"of {lams:.2f} per 1000 hours means a mean lifetime of\n"
            f"MTTF = 1000 / {lams:.2f} = {mttf:.1f} hours (rounded half "
            f"up to one decimal)\n\n"
            f"**Answer:** The mean time to failure of the booster "
            f"station is {mttf:.1f} hours"
        )

    else:
        theta = random.randint(*_T9_THETA_RANGE)
        t_first = theta / 2                          # exact (x.0 or x.5)
        mttf = theta + t_first                       # 1.5*theta, exact
        assert 1500.0 <= mttf <= 7500.0, f"parallel MTTF out of bounds: {mttf}"

        question = (
            f"A cooling loop is served by two identical pumps in "
            f"parallel; the loop keeps operating as long as at least one "
            f"pump operates, and fails only when BOTH pumps have "
            f"failed. The pumps fail independently, with exponentially "
            f"distributed lifetimes averaging {theta} hours each. Using "
            f"the memoryless property, determine the mean time to "
            f"failure (MTTF -- the expected operating time until the "
            f"loop fails) of the cooling loop, in hours, to one decimal "
            f"place."
        )
        solution = (
            f"**Given:**\n"
            f"Two identical pumps in parallel; independent exponential "
            f"lifetimes with mean theta = {theta} hours (failure rate "
            f"1/{theta} per hour each).\n\n"
            f"**Step 1:** Split the loop lifetime into two phases: the "
            f"time until the FIRST pump failure, then the remaining "
            f"lifetime of the surviving pump.\n\n"
            f"**Step 2:** Expected time to the first failure. While both "
            f"pumps run, failures compete: the minimum of two "
            f"independent exponentials is exponential with doubled rate "
            f"2/{theta}, so\n"
            f"E[first failure] = theta / 2 = {theta} / 2 = {t_first:.1f} "
            f"hours\n\n"
            f"**Step 3:** Expected remaining lifetime of the survivor. "
            f"By the memoryless property, the surviving pump is "
            f"statistically good as new at the moment the other fails, "
            f"so its expected remaining lifetime is the full mean:\n"
            f"E[remaining] = theta = {theta} hours\n\n"
            f"**Step 4:** Add the phases.\n"
            f"MTTF = {t_first:.1f} + {theta} = {mttf:.1f} hours\n\n"
            f"**Answer:** The mean time to failure of the cooling loop "
            f"is {mttf:.1f} hours"
        )

    return question, solution
\end{lstlisting}

\vspace{6pt}

\subsection{Advanced}

\vspace{6pt}

\begin{lstlisting}[style=pythonstyle]
def template_equivalent_stiffness_frequency():
    # 1. Parameterize the inputs with random values
    mass = round(random.uniform(5.0, 100.0), 2)       # in kg
    stiffness_1 = random.randint(1000, 50000)         # in N/m
    stiffness_2 = random.randint(1000, 50000)         # in N/m

    # Randomly choose the spring configuration
    configuration = random.choice(['series', 'parallel'])

    # Standardize precision for final outputs
    precision = 3

    # 2. Perform the core calculations for the solution

    # Step A: Calculate the equivalent stiffness (k_eq) based on the configuration
    if configuration == 'parallel':
        k_eq = stiffness_1 + stiffness_2
        formula_str = "k_eq = k1 + k2"
        calculation_str = f"k_eq = {stiffness_1} + {stiffness_2} = {k_eq} N/m"
    else: # configuration == 'series'
        k_eq = (stiffness_1 * stiffness_2) / (stiffness_1 + stiffness_2)
        formula_str = "k_eq = (k1 * k2) / (k1 + k2)"
        calculation_str = (
            f"k_eq = ({stiffness_1} * {stiffness_2}) / ({stiffness_1} + {stiffness_2}) = "
            f"{round(k_eq, 2)} N/m"
        )

    # Step B: Calculate the natural frequency using the equivalent stiffness
    omega_n = math.sqrt(k_eq / mass)

    # 3. Generate the question and solution strings

    question = (
        f"A mass of {mass} kg is attached to a system of two springs. The first spring has a "
        f"stiffness (k1) of {stiffness_1} N/m, and the second spring has a stiffness (k2) of {stiffness_2} N/m. "
        f"The two springs are arranged in {configuration}. "
        f"Determine the equivalent stiffness of the spring system and the resulting undamped "
        f"natural frequency (omega_n) in rad/s."
    )

    solution = (
        f"**Given:**\n"
        f"Mass (m) = {mass} kg\n"
        f"Stiffness 1 (k1) = {stiffness_1} N/m\n"
        f"Stiffness 2 (k2) = {stiffness_2} N/m\n"
        f"Configuration = {configuration.capitalize()}\n\n"

        f"**Step 1:** Calculate the equivalent stiffness (k_eq) for the {configuration} configuration.\n"
        f"The formula for springs in {configuration} is: {formula_str}\n"
        f"{calculation_str}\n\n"

        f"**Step 2:** Calculate the undamped natural frequency (omega_n) using the equivalent stiffness.\n"
        f"The formula is: omega_n = sqrt(k_eq / m)\n"
        f"omega_n = sqrt({round(k_eq, 2)} / {mass})\n"
        f"omega_n = {round(omega_n, precision)} rad/s\n\n"

        f"**Answer:**\n"
        f"The equivalent stiffness of the system is {round(k_eq, 2)} N/m.\n"
        f"The undamped natural frequency is {round(omega_n, precision)} rad/s."
    )

    return question, solution
\end{lstlisting}

\section{Template Certification}
\label{appendix:certification}

This section details the three checks of~\autoref{subsec:certification} (automated integrity
checks, an LLM screen, and expert certification) and their results.

\paragraph{Automated Integrity Checks.} We run every template at 500 seeds.
Every seed must generate an instance without error, regenerate it identically in a separate
process, keep the output format (numbered steps and one answer marker), print no malformed
expression, and pass the closure check: every printed calculation, recomputed from its printed
operands, gives its printed result at the printed precision.
We check by hand the lines that the closure check cannot parse (six line patterns in four
templates), and all 150 templates pass.

\paragraph{LLM Screen.} Three LLM judges from families outside the evaluated models,
\texttt{Grok 4.6}~\citep{grok46}, \texttt{MiniMax M3}~\citep{minimaxm3}, and
\texttt{MiMo-V2.5-Pro}, score each template's code and three of its instances with the prompt
below, at temperature 0, on physical plausibility, mathematical correctness, and pedagogical
clarity.
The screen gives each template one of three outcomes:
\begin{itemize}[nosep, leftmargin=*]
\item \textbf{Critical failure:} At least two judges flag the template for human review, or its
median score on any of the three is below 4.
\item \textbf{Controversial:} Not a critical failure, but the judges' scores on one of the three
have a population standard deviation above 0.5.
\item \textbf{Pass:} Neither of the above.
\end{itemize}

In the first pass, 126 templates pass, 13 are controversial, and 11 are critical failures, and the
judges agree on the review flag with Gwet's AC1 of 0.84~\citep{gwet2008}.
We check each claim in the judges' explanations against the code and the instances, and 34 of the
45 claims hold; after we revise the 24 flagged templates, 147 templates pass, 3 are controversial,
and none is a critical failure in the second pass (AC1 of 0.93).

Against the experts' first-round decisions, 15 of the 147 templates the screen passes are rejected
by a majority of their experts, so 10.2\% of the screen's passes are false positives.
On average, the median scores of experts and judges differ by 0.587 for physical plausibility,
0.187 for mathematical correctness, and 0.787 for pedagogical clarity, on the scale of 1 to 5.
The screen therefore guides revision, and the experts decide certification.

\begin{tcolorbox}[colback=gray!5, colframe=gray!40, boxrule=0.4pt, arc=2pt, left=5pt, right=5pt,
    top=5pt, bottom=5pt, breakable]
\ttfamily\small
You are an expert engineering professor and a senior Python developer acting as a peer reviewer for the EngTrace benchmark.\\
Your task is to meticulously evaluate a new problem template based on its source code and several example outputs.\\
Analyze the provided information and then respond ONLY with a single, valid JSON object that strictly adheres to the schema described in the user prompt. Do not add any explanatory text or markdown formatting around the JSON object.\\[3pt]
Please evaluate the following engineering problem template.\\[3pt]
\textbf{1. Template Source Code:}\\
python \{template\_code\}\\[3pt]
\textbf{2. Generated Instances from the Template:}\\
Instance 1: - Question: "\{q1\}" - Solution: "\{s1\}"\\
Instance 2: - Question: "\{q2\}" - Solution: "\{s2\}"\\
Instance 3: - Question: "\{q3\}" - Solution: "\{s3\}"\\[3pt]
\textbf{3. Evaluation Rubric \& JSON Schema:}\\
Evaluate the template based on the rubric below. The \texttt{human\_review\_flag} should be \texttt{true} if any score is less than 4. The \texttt{explanation} should be a concise, one-sentence justification for the scores and the flag.\\[3pt]
\{\\
\hspace*{1em}"physical\_plausibility\_score": \textless integer, a score from 1--5 based on whether the problem respects the laws of physics and engineering\textgreater,\\
\hspace*{1em}"mathematical\_correctness\_score": \textless integer, a score from 1--5 based on whether the equations and calculations are accurate\textgreater,\\
\hspace*{1em}"pedagogical\_clarity\_score": \textless integer, a score from 1--5 based on whether the problem statement is clear, unambiguous, and solvable\textgreater,\\
\hspace*{1em}"confidence\_score": \textless integer, a score from 1--5 indicating your confidence in this evaluation\textgreater,\\
\hspace*{1em}"human\_review\_flag": \textless boolean, true if the template requires human inspection, otherwise false\textgreater,\\
\hspace*{1em}"explanation": "\textless string, a concise, one-sentence justification for the scores and flag\textgreater"\\
\}
\end{tcolorbox}

\begin{table*}[t]
    \centering
    \small
    \renewcommand{\arraystretch}{1.1}
    \setlength{\tabcolsep}{4pt}
    \begin{tabular}{l p{0.205\textwidth} p{0.205\textwidth} p{0.205\textwidth} p{0.205\textwidth}}
        \toprule
        \rowcolor{gray!10}
        \textbf{Branch} & \textbf{Wrong Constant} & \textbf{Wrong Unit Conversion} &
        \textbf{Flipped Sign} & \textbf{Wrong Printed Result} \\
        \midrule
        Chemical & van der Waals pressure: $R$ with two digits transposed & annulus flow rate: kPa to Pa by a factor of 100 & heat of reaction: Hess's law reversed & first-order batch reactor: one printed time 4\% too high \\
        Electrical & Coulomb's law: $\varepsilon_0$ ten times too large & wave parameters: MHz to Hz by a factor of $10^5$ & C/D and D/C conversion: a delay written as an advance & mean and variance: one printed mean 0.2 too high \\
        Civil & Manning discharge: $R^{3/4}$ for $R^{2/3}$ & cantilever deflection: m to mm by a factor of 100 & effective stress: $\sigma' = \sigma + u$ & beam reactions: one printed reaction 4\% too high \\
        Industrial & safety stock: a two-sided $z$ for a one-sided one & M/M/1 time in system: hours to minutes by a factor of 100 & $\bar{x}$ and $R$ chart limits: the lower limit as center plus $A_2\bar{R}$ & process capability: one printed $C_p$ 6\% too high \\
        Mechanical & torsional shear stress: $J$ with $\pi/4$ for $\pi/2$ & hydrostatic pressure: kPa to Pa by a factor of 100 & U-tube manometer: the pipe-fluid column's sign flipped & natural frequency: one printed $\omega_n$ 4\% too high \\
        \bottomrule
    \end{tabular}
    \caption{\textbf{Planted defects.} The 20 planted defects, four per branch, each written into a
    real template of the branch; every expert of the branch reviews all four.}
    \label{tab:planted_defects}
\end{table*}

\paragraph{Expert Certification.} Three domain experts per branch each review the 30 templates of
their branch mixed with four planted defects.
Each expert sees the items in an order of their own, under opaque codes, and knows that planted
items exist but not which ones.
The experts do not see the screen's results, and they do not discuss a template before all three
have submitted.
Each planted defect is a real template of the branch with one error written into it
(\autoref{tab:planted_defects}); a wrong constant, unit conversion, or sign is carried through the
computation, the printed steps, and the answer, whereas a wrong printed result changes a single
printed line.
The planted defects test whether an expert catches a wrong number, through a constant, a unit
conversion, a sign, or a printed result; whether a question states everything its answer depends
on is examined separately, in the domain experts' reading of six near-miss templates
(\autoref{appendix:error_analysis}).

\begin{figure*}[tp]
    \centering
    \definecolor{framegray}{HTML}{EBDFD0}
    \setlength{\fboxsep}{0pt}
    \setlength{\fboxrule}{0.5pt}
    {\small\textbf{(a)} Hand check}\par\smallskip
    \fcolorbox{framegray}{white}{%
        \includegraphics[width=0.985\textwidth]{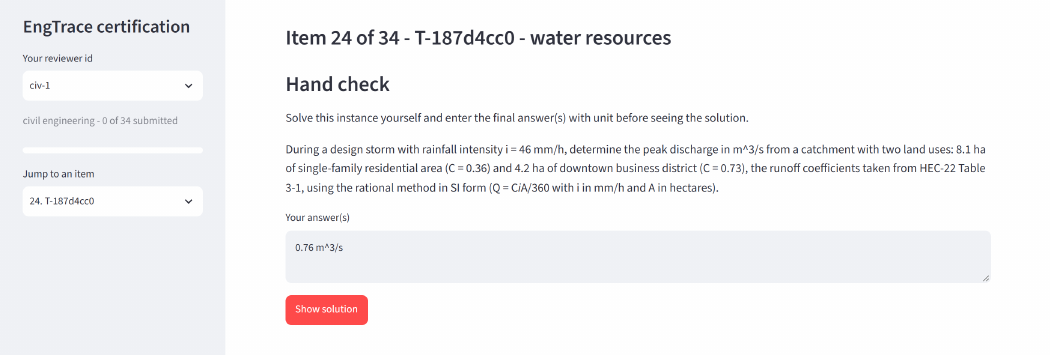}}
    \par\bigskip
    {\small\textbf{(b)} Review}\par\smallskip
    \fcolorbox{framegray}{white}{%
        \includegraphics[width=0.985\textwidth]{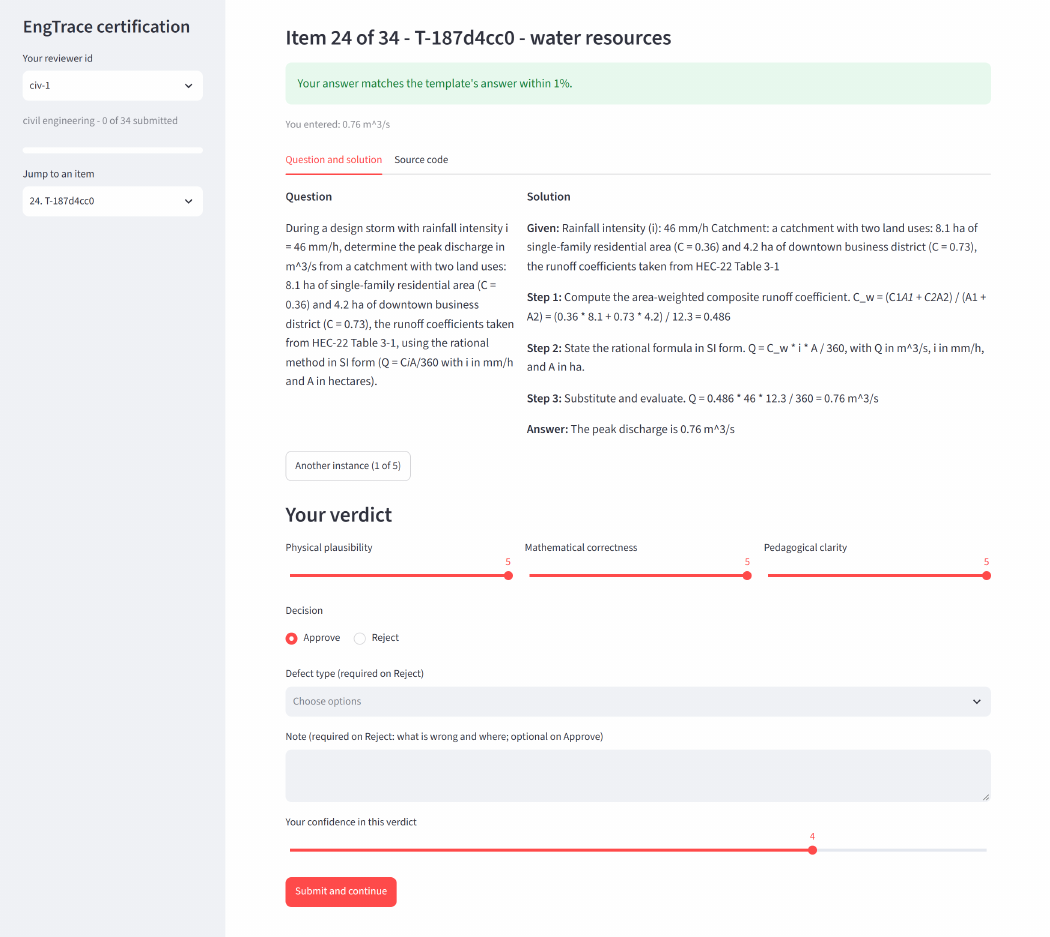}}
    \caption{\textbf{Certification app.}
    (a) The hand check: the expert enters the final answer with its unit before the solution is
    shown.
    (b) The review: the result of the hand check, the question beside the gold solution, the source
    code in a second tab, the template's other instances, and the verdict form.}
    \label{fig:certification_app}
\end{figure*}

\paragraph{Certification App.} The experts work in a \texttt{Streamlit} web app
(\autoref{fig:certification_app}) built around the protocol:
\begin{itemize}[nosep, leftmargin=*]
\item \textbf{Blind hand check:} For each item, the expert solves one instance by hand; the app
shows the solution only once the expert enters an answer.
\item \textbf{Automatic comparison:} The app reports whether the answer is within 1\% of the gold
value and, on a mismatch, asks the expert to find which side is right.
\item \textbf{Multi-instance review:} Beside the gold solution, the expert can read the template's
source code and step through its other four instances, all generated in advance.
\item \textbf{Structured verdict:} The expert scores physical plausibility, mathematical
correctness, and pedagogical clarity from 1 to 5, approves or rejects, rates their confidence
from 1 to 5, and can name defect types from a fixed list: physics or scenario implausible;
governing equation or formula; constant or table value; unit or conversion; sign or direction; a
step that does not follow; final answer or unit wrong; question ambiguous or unsolvable; and
wording or formatting.
\item \textbf{Required justification:} The app accepts a rejection only with a note on what is
wrong and where and at least one defect type.
\end{itemize}

\begin{table}[!htbp]
    \centering
    \small
    \setlength{\tabcolsep}{3.5pt}
    \begin{tabular}{l|ccc|ccc}
        \toprule
        \rowcolor{gray!10}
        \multirow{2}{*}{\textbf{Branch}} & \multicolumn{3}{c|}{\textbf{Approve or Reject}} &
        \multicolumn{3}{c}{\textbf{AC2 on Scores}} \\
        & $\kappa$ & \textbf{AC1} & \textbf{Agr.} &
        \textbf{Phys.} & \textbf{Math.} & \textbf{Ped.} \\
        \midrule
        Chemical & 0.492 & 0.776 & 77\% & 0.898 & 0.952 & 0.912 \\
        Electrical & 0.365 & 0.926 & 90\% & 0.941 & 0.989 & 0.976 \\
        Civil & -- & 1.000 & 100\% & 0.993 & 0.981 & 0.974 \\
        Industrial & -- & 1.000 & 100\% & 0.962 & 0.997 & 0.965 \\
        Mechanical & 0.627 & 0.792 & 80\% & 0.927 & 0.973 & 0.952 \\
        \textbf{All} & 0.589 & 0.914 & 89\% & 0.947 & 0.980 & 0.954 \\
        \bottomrule
    \end{tabular}
    \caption{\textbf{Expert agreement.} Agreement among the three experts of each branch in the
    first round: Fleiss' $\kappa$, Gwet's AC1, and percent agreement on the decision, and Gwet's AC2
    on the physical, mathematical, and pedagogical scores.
    --: undefined, since every decision of the branch is an approval.}
    \label{tab:expert_agreement}
\end{table}

\paragraph{Results.} Every expert rejects all four planted defects of their branch, 60 of 60
readings, one per expert and planted defect.
Each expert makes one hand check per item.
Of the 510 first-round hand checks, 504 have a number to compare; 459 of these match the gold value
within 1\%, and 44 of the 45 mismatches end in a rejection.
\autoref{tab:expert_agreement} gives the first-round agreement within each branch.
Fleiss' $\kappa$ is undefined for civil and industrial, where every decision is an approval, and is
deflated where nearly every template is approved~\citep{feinstein1990high}: in electrical, 5 of the
90 decisions on templates are rejections, so 90\% agreement gives a $\kappa$ of 0.365.
Gwet's AC1 and AC2 are not deflated in this way, so the table reports them beside $\kappa$.
In the first round, 43 of the 450 decisions on templates are rejections.
At least one expert rejects 22 templates (chemical 9, electrical 3, and mechanical 10), 15 of them
by a majority, and every civil and industrial expert approves all 30 templates of their branch.

A template that any expert rejects, or whose code changes, returns to the same three experts until
all three approve it.
It is revised where the rejection holds, and each expert hand-checks a new instance that none of
the three has seen.
Certification takes six rounds:
\begin{itemize}[nosep, leftmargin=*]
\item \textbf{Round 1:} All 150 templates and the 20 planted defects; 43 of the 450 decisions on
templates are rejections, and at least one expert rejects 22 templates (chemical 9, electrical 3,
and mechanical 10), 15 of them by a majority, while every civil and industrial expert approves all
30 templates of their branch.
\item \textbf{Round 2:} The 22 templates that at least one expert rejected; 10 of the 66 decisions
are rejections, on 5 of the 22 templates, and 66 of 66 hand checks match the gold value.
\item \textbf{Round 3:} The 5 templates rejected in round 2; all three experts approve all 5, and
13 of 15 hand checks match, the other two giving the gold value in pascals where the gold answer
is in kilopascals.
\item \textbf{Round 4:} Two templates whose sampled parameters are widened so that each supplies
15 distinct instances; all three experts approve both, and 6 of 6 hand checks match.
\item \textbf{Round 5:} Two templates whose questions state the system, the form of the equation,
and the property data on which the answer depends; one of the 6 decisions is a rejection, on the
physical plausibility of compressing organic compounds above their decomposition temperatures, and
6 of 6 hand checks match.
\item \textbf{Round 6:} The template rejected in round 5, with organic compounds held below their
decomposition temperatures; all three experts approve it, and 3 of 3 hand checks match.
\end{itemize}
Every template is thus approved unanimously in the latest round that reviewed it.
Its current code regenerates, byte for byte, the instances its experts reviewed, and the evaluation
set holds instances of these certified versions only.

\section{Scoring Details}
\label{appendix:scoring}

This section details the four checks of~\autoref{sec:evaluation};
\autoref{tab:scoring_settings} tests each setting on the expert study
of~\autoref{appendix:validation}, and~\autoref{box:worked_example} scores one response end to end.
Every check reads the response a model returns, not the reasoning text that some endpoints return
separately.

\subsection{Final-Answer Check}

The check reads the final answer from the 700 characters that follow the response's last
final-answer heading or, failing that, its answer marker (the first when several lie within 700
characters of one another, otherwise the last) or its last mention of ``answer'', and otherwise
from its last 700 characters; a response with neither a heading or marker nor a number or label
word in its last 700 characters has no readable final answer.
It reads scientific notation, LaTeX, fractions, multiples of $\pi$, and thousands separators as
numbers, but not a digit subscripted with an underscore, as in \texttt{p\_1}.
The targets are the milestones on the gold trace's final-answer line, the last one for a scalar or
array answer and every one otherwise; where the line states no milestone, the target is its last
value that the question does not give, so a vector or multipart answer is then checked on that
value alone.
A classification answer also takes as targets any label word that its gold line states, which the
last non-negated word of the same family (e.g., laminar or turbulent) in the response's final answer
must match, and any yes-or-no property, matched label by label.

\autoref{eq:match} uses the unit factors $10^{\pm 3}$, $10^{\pm 6}$, $10^{\pm 9}$, $10^{\pm 12}$,
100, 60, 3,600, and their inverses, and $\epsilon$ is the check's one fitted constant.
A response's last digit counts only when one unit of it is smaller than the value, and its term
$c\,u(\hat{y})$ in~\autoref{eq:match} counts for at most 1\% of the target; values also match in
absolute value, a question that prescribes its rounding requires the exact digits, and a scalar gold
answer stated in two units accepts either.
Every one of the 2,250 gold traces scores correct.

\paragraph{What Must Match.} Matching in absolute value leaves a target's sign unchecked, since sign
conventions differ (a deflection given as a negative displacement, a contraction as a negative
change), and a unit factor applies whatever unit is written beside a value, since answers often
state another unit (kilonewtons for newtons).
Each target is matched by any value in the final answer, not by the value written for that quantity,
and an array answer is matched by its last value.
Labels must match on classification items; on six templates of other kinds the question also asks
for a label (whether aliasing occurred, the limiting reactant, the signal type, the system type,
whether the flow is rotational, the direction of propagation), which no target checks.

\paragraph{Symbolic Answers.} Nine templates answer with an expression, such as a signal or a
velocity field, and the check scores them by the numbers the expression states.
On five of the nine templates with symbolic answers it also decides equivalence: it parses the gold
expression and every expression in the response's final answer with SymPy~\citep{meurer2017sympy},
and accepts the response when, at every sample point of the free variable, its value lies within
half a unit of the last digit of every decimal the gold writes, so an equal expression in another
form is correct and an answer coarser than the gold is not.
This step only raises an incorrect or partial verdict to correct and never lowers one.
Against domain experts' equivalence grades, it reaches precision 1.000 and recall 0.990 on the 264
verdicts the experts graded, which also served to settle the rule, and precision 1.000 and recall
0.840 on the 30 readings of these templates in the experts' readings of the evaluated models
(\autoref{appendix:validation} and~\autoref{appendix:error_analysis}), which the rule did not see.
The other four templates with symbolic answers stay scored by the numbers they state.

\paragraph{Prescribed Digits.} 17 templates prescribe the rounding of their answer and of the values
it depends on, as their procedures do in practice (an order quantity in whole units, a standard
normal value to four decimals), and the check then requires the prescribed digits: a value within
the tolerance but not at those digits is incorrect by the question's own terms.
We keep this requirement as part of the task; relaxing it to the tolerance raises 136 verdicts on 14
templates and keeps Kendall's $\tau$ with the ordering as scored at 0.917
(\autoref{tab:scoring_variants}).

\paragraph{Sensitivity of the Rule.} \autoref{tab:scoring_variants} re-applies the rule with one
clause changed at a time.
Turning off the absolute-value clause lowers 310 verdicts on 20 templates ($\tau = 0.782$); leaving
the last-digit term uncapped raises 63 verdicts on 35 templates ($\tau = 0.964$); and partial credit
in proportion to the parts matched, instead of one half, moves 90 verdicts on 10 templates
($\tau = 1.000$).
Of the 22,635 accepted numeric answers, 93.1\% lie within 0.2\% of their target, 6.5\% between
0.2\% and 1\%, which the last-digit terms admit, and 0.4\% further from it.
Scoring instead every stated part and label part as a target, accepting a unit factor only where the
written unit allows it and the absolute value only where the sign is a convention, judging a target
near zero without its own last-digit window, and reading the notation and label wording of the
discrepant readings (\autoref{appendix:readings}) as the domain experts read them moves no model's
FAC by more than 0.008; the only change in the ordering is that \texttt{GPT-5.4 mini} and
\texttt{Muse Glimmer 30B}, 0.0002 apart as scored, trade places ($\tau = 0.964$).
Requiring a correct rounding, within half a unit of the last digit rather than one unit, moves a
model's FAC by up to 0.026, with the same exchange of these two models.

\subsection{Milestones and the Judge}

\paragraph{Milestones.} We rerun each template at the instance's seed and, if the regenerated
question and gold trace equal the instance byte for byte, take as milestones the computed values
that the gold trace states within 0.5\% and the question does not give under any unit factor.
We exclude values within $10^{-9}$ of 0, 1, or 2, bookkeeping variables, the intermediate states of
iterative schemes, and the elements of containers such as lookup tables, and count a quantity
stored twice once.
Matching uses the unit factors above except $10^{\pm 12}$, and MC leaves out the 57 instances
without milestones.
Matching checks the value alone: not the quantity's name, the unit written beside it, the step it
appears in, or the order and dependencies among steps, so a value that coincides with a milestone
counts, and the chance floor of~\autoref{tab:coverage}, the same responses matched against a
sibling instance's milestones, measures how often that happens.
The judge sees each milestone's name and value and rules on the response's own route; it is the
one check that can credit a quantity stated in another form or excuse one the route does not need.

\paragraph{The Judge.} \texttt{MiMo-V2.5-Pro} (Xiaomi) receives one call per response in which
matching misses a milestone, with the prompt below, at temperature 0, with JSON output and up to
three attempts; it decides 294 of the 1,245 milestones of the expert study.
We credit only reached because the judge calls none of the 88 unstated values of
\autoref{tab:scoring_settings} reached but excuses a quarter of them as not needed.

\begin{tcolorbox}[
    colback=gray!5,
    colframe=gray!40,
    boxrule=0.4pt,
    arc=2pt,
    left=6pt,
    right=6pt,
    top=6pt,
    bottom=6pt,
    breakable
]
\footnotesize
\ttfamily
You are checking one student's worked solution to an engineering problem. \\

PROBLEM: \\
\{question\} \\

STUDENT SOLUTION: \\
\{trace\} \\

A correct solution to this problem obtains the intermediate quantities listed below.
An automatic checker could not find them in the student's solution by matching numbers.
For EACH one, decide: \\

- "REACHED": the student does obtain this quantity correctly - possibly in different units, a
different but equivalent form, or with different rounding. \\
- "NOT\_NEEDED": the student reaches a correct result by a valid route that does not pass through
this quantity at all. \\
- "MISSING": the student does not obtain this quantity, or obtains it incorrectly. \\

QUANTITIES: \\
\{quantities\} \\

Return a raw JSON object only: \\
\{"results": [\{"milestone": "\textless name\textgreater", "verdict": "REACHED" | "NOT\_NEEDED" |
"MISSING"\}]\}
\end{tcolorbox}

\begin{table}[t]
    \centering
    \footnotesize
    \setlength{\tabcolsep}{3pt}
    \begin{tabular}{l|l|c}
        \toprule
        \rowcolor{gray!10}
        \textbf{Check} & \textbf{Setting} & \textbf{Result} \\
        \midrule
        Final answer & $\epsilon$ fitted on each half & 0.915, 0.939 \\
        \midrule
        Milestones & \textbf{0.5\%, unit factors} & \textbf{0.705} \\
        & 0.2\%, unit factors & 0.697 \\
        & 1\%, unit factors & 0.653 \\
        & 2\%, unit factors & 0.568 \\
        & 0.5\%, no unit factors & 0.696 \\
        \midrule
        Judge & \textbf{reached only} & \textbf{0 of 88} \\
        & reached or not needed & 22 of 88 \\
        \midrule
        Arithmetic & \textbf{digits shown, as used} & \textbf{0.817 / 0.427} \\
        & digits shown alone & 0.581 / 0.646 \\
        & 0.1\% tolerance & 0.345 / 0.107 \\
        & 1\% tolerance & 0.171 / 0.039 \\
        \bottomrule
    \end{tabular}
    \caption{\textbf{Scoring settings.} Each check's setting, in bold where it has alternatives, on
    the expert study: the three-way agreement with the experts on the half of the responses that
    $\epsilon$ was not fitted on (fitted values 0.0015 and 0.0020); the separation between a
    response's coverage of its own milestones and of the other instances of its template; the
    unstated values credited among 88 milestone values multiplied by 1.37; and precision / recall on
    the incorrect steps inside correct-answer responses.}
    \label{tab:scoring_settings}
\end{table}

\subsection{Arithmetic Check}

The check evaluates each chain of numeric segments in a step, such as $a = b = c$, with
SymPy~\citep{meurer2017sympy}, skips symbolic segments, and flags a result that is not a correct
rounding, at its displayed precision, of the value recomputed from the response's displayed
operands: $0.92979^{2/3} = 0.95300$ is flagged (0.95263 at five decimals), whereas
$0.92979^{2/3} = 0.9526$ passes.
Unit conversions are compared after their factor, and no result is held to a precision that its
operands cannot determine.
It flags none of the 6,875 calculations it reads in the 2,250 gold traces.

\subsection{Judged Step Check}

The check sends the judge one call per response with the prompt below, at the provider's default
settings, reviewing every non-empty step that the arithmetic check does not flag, and flags a step
the judge labels a conceptual or calculation error.
Its flags are the judge's rulings, not verified errors: \autoref{appendix:validation} reports their
precision against the domain experts, and~\autoref{tab:judged_steps} their rates on the evaluated
models.

\begin{tcolorbox}[
    colback=gray!5,
    colframe=gray!40,
    boxrule=0.4pt,
    arc=2pt,
    left=6pt,
    right=6pt,
    top=6pt,
    bottom=6pt,
    breakable
]
\footnotesize
\ttfamily
You are an expert engineering professor acting as an automated evaluator.
Your task is to analyze the Student's reasoning steps against the Ground Truth steps. \\

QUESTION: \\
\{question\} \\

GROUND TRUTH CHAIN: \\
\{gold steps\} \\

STUDENT CHAIN: \\
\{response steps\} \\

TASK: \\
Analyze the Student Steps at these specific indices (0-based): \{indices\}. \\
Compare them logically to the Ground Truth. \\

CRITICAL RULES: \\
1. If the step is factually correct but takes a different path than the Ground Truth, you MUST use
"Alternative Correct". \\
2. Do not contradict the category definitions below. \\
3. The final output must be a raw JSON object only. \\

ERROR CATEGORIES: \\
- "Conceptual Error": The model applied the wrong scientific principle or formula. \\
- "Calculation Error": The model used the correct formula but made a mathematical mistake. \\
- "Alternative Correct": The model's step is valid and logically sound, but follows a different
method/phrasing. \\
- "Other": Nonsensical, irrelevant, hallucination, or formatting errors. \\

OUTPUT FORMAT: \\
Return a single JSON object containing a "results" list. Each entry must have exactly two keys:
"step\_index" and "category". \\
\{"results": [\{"step\_index": \{first index\}, "category": "Alternative Correct"\}]\}
\end{tcolorbox}

\subsection{Worked Example}

\autoref{box:worked_example} follows one wrong answer from the evaluation through every check, from
the final-answer match to its scores $v(r)$ and $m(r)$.
\autoref{box:coverage_cases} then shows what coverage does and does not measure: a correct
derivation that takes a valid route past two gold milestones and scores $m(r) = 1/3$, and a
planted misstated rule whose numbers are those of the right rule, which every deterministic check
passes and only some judges see.

\begin{figure*}[t]
\centering
\definecolor{wePanel}{HTML}{FAFAF4}
\definecolor{weFrame}{HTML}{EBDFD0}
\definecolor{wePill}{HTML}{F6DBE2}
\definecolor{weModule}{HTML}{FFF1CC}
\definecolor{weShadow}{HTML}{DCD4C6}
\definecolor{weAccent}{HTML}{8C6E62}
\tcbset{
  we pill/.style={on line, colback=wePill, colframe=wePill, boxrule=0pt, arc=3pt, boxsep=0pt,
    left=6pt, right=6pt, top=2.5pt, bottom=2.5pt, fontupper=\bfseries},
  we module/.style={sharp corners, colframe=weShadow, leftrule=0pt, toprule=0pt, rightrule=2pt,
    bottomrule=2pt, boxsep=0pt, left=10pt, right=10pt, top=8pt, bottom=8pt, before skip=6pt,
    after skip=14pt},
}
\begin{tcolorbox}[colback=wePanel, colframe=weFrame, boxrule=1.5pt, arc=6pt, left=12pt, right=12pt,
    top=12pt, bottom=12pt]
\small

\noindent\tcbox[we pill]{Question}\enspace{\color{weAccent}as posed; chemical engineering, Intermediate}
\begin{tcolorbox}[we module, colback=weModule]
Estimate the dynamic viscosity ($\mu$) of Sulfur Hexafluoride (SF$_{6}$) gas at a temperature of 430
K.\\[3pt]
Use the Chapman-Enskog equation for low-density gases:\\[3pt]
\hspace*{1.5em}$\mu$ = (2.6693e-6 * sqrt(M*T)) / ($\sigma$\^{}2 * $\Omega$\_$\mu$)\\[3pt]
You are given the following parameters for Sulfur Hexafluoride (SF$_{6}$):\\[1pt]
\hspace*{1.5em}--~Molar Mass (M) = 146.06 g/mol\\[1pt]
\hspace*{1.5em}--~Lennard-Jones diameter ($\sigma$) = 5.128 \AA{}\\[1pt]
\hspace*{1.5em}--~Collision Integral ($\Omega$\_$\mu$) = 1.1895\\[3pt]
Provide your answer in units of Pascal-seconds (Pa$\cdot$s).
\end{tcolorbox}

\noindent\tcbox[we pill]{Gold milestones}\enspace{\color{weAccent}($\mathcal{M}$, 7) and how each was
settled}
\begin{tcolorbox}[we module, colback=white]
\renewcommand{\arraystretch}{1.3}
\begin{tabular*}{\linewidth}{@{\extracolsep{\fill}}l r l l@{}}
\toprule
\textbf{Milestone} & \textbf{Gold value} & \textbf{Settled by} & \textbf{Outcome} \\
\midrule
\texttt{MT} & $62{,}805.8$ & matching & states $62.8058$, $c = 10^{3}$ \\
\texttt{T\_star} & $1.9361$ & judge & not needed \\
\texttt{denominator} & $31.28$ & judge & reached \\
\texttt{epsilon} & $222.1$ & judge & not needed \\
\texttt{numerator} & $0.00066896$ & judge & missing \\
\texttt{sigma2} & $26.2964$ & judge & reached \\
\texttt{viscosity} & $2.13862\times10^{-5}$ & judge & missing \\
\bottomrule
\end{tabular*}
\end{tcolorbox}

\noindent\begin{tabular}{@{}p{3.5cm}@{\hspace{10pt}}p{\dimexpr\linewidth-3.5cm-10pt\relax}@{}}
\tcbox[we pill]{Arithmetic check} &
The flagged step prints $(2.1155\times10^{-5}) / (3.128\times10^{-19}) \approx 6.762\times10^{-5}$\,Pa$\cdot$s;
the printed operands give $6.763\times10^{13}$, which does not round to the printed result at its last
digit ($1\times10^{-8}$), so the step is flagged (the one flag in 9 steps, 3 calculations parsed).
\\[10pt]
\tcbox[we pill]{Final answer} &
The response states $\hat{y} = 6.762\times10^{-5}$ against the target $y = 2.139\times10^{-5}$: at
$c = 1$ the gap is 216\% of $y$, and no other factor in $\mathcal{C}$ brings it within $\epsilon\,|y|$
or one unit of either last digit (\autoref{eq:match}), so the answer is incorrect and $v(r) = 0$.
\\[10pt]
\tcbox[we pill]{Coverage} &
$|\mathcal{R}(r)| = 1$ matched $+\,2$ ruled reached $= 3$ of $|\mathcal{M}| = 7$, so
$m(r) = 3/7 = 0.43$ (\autoref{eq:coverage}).
\\
\end{tabular}
\end{tcolorbox}
\caption{\textbf{A worked scoring example.} A wrong answer by \texttt{Qwen3-235B-2507} on an
         Intermediate chemical-engineering instance. Matching settles 1 of the 7 milestones under a
         unit factor, the judge rules the other 6, the arithmetic check flags 1 of the 9 steps, and
         the response scores $v(r) = 0$ and $m(r) = 3/7$.}
\label{box:worked_example}
\end{figure*}

\begin{figure*}[t]
\centering
\definecolor{wePanel}{HTML}{FAFAF4}
\definecolor{weFrame}{HTML}{EBDFD0}
\definecolor{wePill}{HTML}{F6DBE2}
\definecolor{weModule}{HTML}{FFF1CC}
\definecolor{weShadow}{HTML}{DCD4C6}
\definecolor{weAccent}{HTML}{8C6E62}
\tcbset{
  we pill/.style={on line, colback=wePill, colframe=wePill, boxrule=0pt, arc=3pt, boxsep=0pt,
    left=6pt, right=6pt, top=2.5pt, bottom=2.5pt, fontupper=\bfseries},
  we module/.style={sharp corners, colframe=weShadow, leftrule=0pt, toprule=0pt, rightrule=2pt,
    bottomrule=2pt, boxsep=0pt, left=10pt, right=10pt, top=8pt, bottom=8pt, before skip=6pt,
    after skip=12pt},
}
\begin{tcolorbox}[colback=wePanel, colframe=weFrame, boxrule=1.5pt, arc=6pt, left=12pt, right=12pt,
    top=12pt, bottom=12pt]
\small

\noindent\tcbox[we pill]{A.\enspace A valid route that bypasses milestones}

\smallskip
\noindent A mechanical-engineering question (Intermediate) gives a velocity field and asks for the
vorticity at a point:
\begin{tcolorbox}[we module, colback=weModule]
A 2D fluid flow is described by the velocity field:\\[3pt]
\hspace*{1.5em}u = -1xy\\[1pt]
\hspace*{1.5em}v = 1x\^{}2 - 2y\^{}2\\[3pt]
where u and v are in m/s, and x and y are in meters.\\[3pt]
Calculate the vorticity at the point (1.3, 1.1) m and determine if the flow is rotational or
irrotational at that point.
\end{tcolorbox}
\begin{tcolorbox}[we module, colback=white]
\renewcommand{\arraystretch}{1.3}
\begin{tabular*}{\linewidth}{@{\extracolsep{\fill}}l r l l@{}}
\toprule
\textbf{Milestone} & \textbf{Gold value} & \textbf{Settled by} & \textbf{Outcome} \\
\midrule
\texttt{du\_dy\_val} & $-1.3$ & judge & not needed \\
\texttt{dv\_dx\_val} & $2.6$ & judge & not needed \\
\texttt{zeta\_z} & $3.9$ & matching & states $3.9$, $c = 1$ \\
\bottomrule
\end{tabular*}
\end{tcolorbox}
\noindent The gold trace evaluates the two partial derivatives at the point; the response
differentiates symbolically, $\omega_z = 2x - (-x) = 3x$, and evaluates $3(1.3) = 3.9$ in one step,
so it never states \texttt{du\_dy\_val} and \texttt{dv\_dx\_val}. The answer is correct, $v(r) = 1$;
MC credits only reached, so $m(r) = 1/3$, and route-adjusted coverage, with the not-needed
milestones out of the denominator, is 1/1.

\medskip
\noindent{\color{weFrame}\rule{\linewidth}{1.2pt}}
\medskip

\noindent\tcbox[we pill]{B.\enspace Matched values behind a misstated rule}

\smallskip
\noindent A planted defect of the validation study: a clean response by \texttt{Claude Opus 4.7} to a
rectangular-channel discharge question (Manning's equation) with one step's rule text changed and no
digit changed.
\begin{tcolorbox}[we module, colback=weModule]
\renewcommand{\arraystretch}{1.3}
\begin{tabular}{@{}l l@{}}
{\color{weAccent}\textit{Untouched step:}} & Compute the hydraulic radius. R = A / P = 6.48 / 7.5 = 0.864 m \\
{\color{weAccent}\textit{Planted step:}} & Compute the hydraulic radius.
    {\setlength{\fboxsep}{1.5pt}\colorbox{wePill}{R = P / A}} = 6.48 / 7.5 = 0.864 m \\
\end{tabular}
\end{tcolorbox}
\noindent The hydraulic radius is $A/P$; the planted step states $P/A$ while its numbers still
compute $A/P$ ($6.48 / 7.5 = 0.864$\,m passes the arithmetic check), every milestone matches, and
the final answer is correct, so no deterministic check flags it. Shown the step alone,
\texttt{GPT-5} and \texttt{Grok 4.6} label it a conceptual error and \texttt{MiMo-V2.5-Pro} (the
judge of this paper) does not.
\end{tcolorbox}
\caption{\textbf{What coverage does and does not measure.} A: a correct, valid derivation that
         bypasses two of three gold milestones scores $m(r) = 1/3$, since Milestone Coverage credits
         the gold route. B: a misstated rule whose numbers are those of the right rule passes
         matching, the arithmetic check and the answer check; only a judge can see it, and not every
         judge does.}
\label{box:coverage_cases}
\end{figure*}

\section{Validation Against Expert Judgment}
\label{appendix:validation}

This section details the expert study behind~\autoref{sec:evaluation} and the domain experts'
readings of the evaluated models' responses; \autoref{tab:validation_results} collects the results.

\subsection{Expert Study}

The study draws four instances from each of 15 templates, 60 in all, one template per branch and
level: within each branch and level it takes the template whose answer kind is least represented so
far, breaking ties by a seeded hash, and it leaves out the four templates whose answers can be read
off the question's wording (\autoref{appendix:error_analysis}).
This rule favors symbolic, vector, and multipart answers, so only 3 of its templates have a single
scalar answer, against 89 of the 150, which makes it a hard test of answer checking.
Five LLMs outside the evaluated models (\texttt{GPT-5}~\citep{gpt5},
\texttt{Claude Opus 4.7}~\citep{claudeopus47}, \texttt{Gemini 3.1 Pro}~\citep{gemini31pro},
\texttt{DeepSeek-R1}~\citep{guo2025deepseekr1}, and
\texttt{Llama 3.1 70B}~\citep{grattafiori2024llama3}) answer each instance once, giving 300
responses.
Fifteen domain experts, three per branch, label each response of their branch behind an opaque code:
every step (correct, alternative correct, incorrect with an error type, or not a claim), every
milestone, the final answer (correct, partial, incorrect, or not stated), and the soundness of the
reasoning.
Three rounds follow: a written reason for each of the 1,042 incorrect step labels, a blind
re-labeling of seven responses per expert, and a blind adjudication of the 255 distinct steps on
which experts split.
The experts agree at Fleiss'~$\kappa$~\citep{fleiss1971} 0.781 on steps, 0.880 on milestones, and
0.966 on final answers, and an expert's two passes at Cohen's~$\kappa$~\citep{cohen1960} 0.828.

The study both guides and validates the evaluator.
Only $\epsilon$ is cross-fitted, fitted on one half of the responses and tested on the other; the
other settings of~\autoref{tab:scoring_settings} are compared with their alternatives on all 300
responses, so the agreement figures of~\autoref{tab:validation_results} are not an estimate on
unseen responses.
The experts' readings of the evaluated models' responses (\autoref{appendix:readings}) test the
evaluator on responses it was not designed on.

\subsection{Evaluator Designs}

We compare six designs on these labels.
\begin{itemize}
    \item \textbf{Step matching with in-family judges}: matches response steps to gold steps by
    numerical agreement and semantic similarity, sends unmatched steps to a panel of \texttt{GPT-5},
    \texttt{Claude Opus 4.5}~\citep{claudeopus45}, and \texttt{Gemini 3.1 Pro} when the answer is
    correct but fewer than 80\% of the gold steps match and for a random fifth of wrong answers, and
    scores an F1 over the steps; we run it with two and with three judges voting.
    The panel's Claude judge is \texttt{Claude Opus 4.5}, a different model from the study's
    \texttt{Claude Opus 4.7} but of the same family, which is what makes the panel in-family.
    \item \textbf{Step matching with outside judges}: the same, with \texttt{Grok 4.6},
    \texttt{MiniMax M3}, and \texttt{MiMo-V2.5-Pro}.
    \item \textbf{Process reward models}: three open-weights models that assign each step a reward,
    with a step below 0.5 counted as incorrect (\texttt{Qwen2.5-Math-PRM-72B},
    \texttt{Qwen2.5-Math-PRM-7B}~\citep{zhang2025prmlessons}, and
    \texttt{VersaPRM}~\citep{zeng2025versaprm}).
    The threshold of 0.5 is the models' default: choosing it on one half of the responses and
    testing on the other changes step F1 by $-0.006$, and inside correct answers no threshold
    reaches precision 0.80 on the held-out half.
    \item \textbf{Milestone matching alone}: credits each gold milestone that the response states
    within 0.5\% under a unit factor, in any step and order (\autoref{sec:evaluation}), with no
    judge.
    \item \textbf{Milestone matching with the arithmetic check}: milestone matching plus the
    arithmetic check, which recomputes each displayed calculation it can parse and flags a result
    that is not a correct rounding at its displayed precision.
    \item \textbf{Milestone matching with the judge}: milestone matching, with
    \texttt{MiMo-V2.5-Pro} ruling each milestone that matching misses reached, not needed, or
    missing, and only reached credited.
\end{itemize}
Our evaluator combines the last three with the final-answer check and the judged step check.

\paragraph{Main Findings.} Whether the answer is right nearly decides the experts' judgment of a
response's soundness (area under the receiver operating characteristic curve, AUROC, 0.974), so
the soundness judgment cannot tell reasoning evaluators apart, and we compare them on steps and
milestones, where matching with the judge agrees best with the experts' milestone labels.
Of the 178 incorrect steps behind correct answers, 175 are calculation slips, which rarely change
milestone coverage; the arithmetic check is there to find them.
The best reward model ranks steps well (AUROC 0.825), but three of four of its flags inside correct
answers are false, and outside judges barely move the step-matching score (0.379 with two in-family
judges, 0.378 with three outside ones).

\begin{table*}[t]
    \centering
    \small
    \begin{tabular}{l|l|l}
        \toprule
        \rowcolor{gray!10}
        \textbf{Component} & \textbf{Basis} & \textbf{Result} \\
        \midrule
        \multicolumn{3}{l}{\textit{Agreement with the expert labels on 300 responses}} \\
        Final-answer check & three-way, all 300 responses & 0.930 \\
        Final-answer check & 281 responses not marked partial & 0.986 (step matching 0.747) \\
        Milestone matching & P / R / F1 & 0.927 / 0.924 / 0.926 \\
        Milestone matching with the judge$^\dagger$ & P / R / F1 & 0.930 / 0.989 / 0.958 \\
        Arithmetic check & P / R, all steps; correct answers & 0.872 / 0.317; 0.817 / 0.427 \\
        Judged step and arithmetic checks$^\dagger$ & P / R, all steps; correct answers & 0.707 / 0.603; 0.703 / 0.360 \\
        Judged step check alone$^\dagger$ & P, all steps; correct answers & 0.640; 0.467 (7 of 15) \\
        Best process reward model & P, all steps; correct answers & 0.539; 0.246 \\
        \midrule
        \multicolumn{3}{l}{\textit{Planted defects: conceptual / arithmetic caught, untouched steps flagged}} \\
        Deterministic checks & conceptual & 0 of 60 \\
        \texttt{GPT-5} & LLM judge, one step & 20 of 60 / 43 of 60 / 0 of 120 \\
        \texttt{Claude Opus 4.5} & LLM judge, one step & 8 of 60 / 43 of 60 / 0 of 120 \\
        \texttt{MiMo-V2.5-Pro}$^\ddagger$ & LLM judge, one step & 16 of 51 / 43 of 60 / 0 of 111 \\
        \texttt{Grok 4.6} & LLM judge, one step & 22 of 60 / 52 of 60 / 0 of 120 \\
        Judged step check & conceptual, every unflagged step sent & 19 of 60 \\
        Step matching with in-family judges & conceptual, end to end & 9 of 60 \\
        \midrule
        \multicolumn{3}{l}{\textit{Readings of the evaluated models' responses}} \\
        Arithmetic flags & 189 decided flags & 171 real (precision 0.905) \\
        Final answer correct & 166 readings & 160 confirmed \\
        Final answer incorrect & 86 readings & 75 confirmed, 8 called correct \\
        Final answer partial & 36 readings & 18 called fully correct, 9 incorrect \\
        Judge reached & 98 readings & 83 confirmed \\
        Judge missing & 96 readings & 75 confirmed, 33 of them not needed by the route \\
        \bottomrule
    \end{tabular}
    \caption{\textbf{Validation results.} Top: each component against the domain experts' labels on
    the expert study (P precision, R recall; correct answers are responses with a correct final
    answer).
    Middle: planted defects caught and untouched steps flagged.
    Bottom: the domain experts' readings of the evaluated models' responses.
    $^\dagger$Measured on the milestone matches and arithmetic flags recorded when the study was
    scored (milestone F1 0.921 without the judge); the unmarked rows run each check as the
    evaluation runs it, so the two milestone F1 values, 0.926 and 0.958, are separate measurements
    and their difference is not the judge's contribution alone.
    $^\ddagger$17 of its 240 calls returned no reply after retries, so its counts rest on the 51
    conceptual defects and the 111 untouched steps with a reply.}
    \label{tab:validation_results}
\end{table*}

\subsection{Planted Defects}

Because the labels hold almost no conceptual error behind a correct answer, we plant one defect in
each of 120 clean responses, verified without the checker under test: 60 one-digit slips in a
displayed intermediate value and 60 conceptual defects that misstate a criterion, rule, or formula
without changing any digit.
Each LLM judge sees the planted step and, separately, the same step untouched, with the step prompt
of~\autoref{appendix:scoring}, and a defect counts as caught only when the judge flags the planted
step alone.
None of our deterministic checks flags a conceptual defect, and the four judges label at most about
a third of them an error, without false alarms (up to 33 of 60 when ``Other'' labels count); the
judged step check, which sends every unflagged step, catches more than step matching, whose panel
sees a conceptual defect no more often than the same step untouched.

\subsection{Judge Independence}

We use \texttt{MiMo-V2.5-Pro} because it shares no family with an evaluated model, as LLM judges can
favor their own family~\citep{panickssery2024selfrecognition, li2026preferenceleakage}, and catches
about as many planted defects as \texttt{GPT-5}.
Re-judging 201 sampled responses, 14 to 20 per model, with \texttt{Grok 4.6} from a third family
changes each model's coverage of them by $-0.018$ to $+0.019$, and every 95\% interval includes
zero.
In the step-matching panel, dropping a response model's own-family judge changes its score by at
most 0.006, as much as dropping one of the other two (the design detects changes of 0.003 to
0.025), yet each judge accepts 90\% to 97\% of the incorrect steps it judges outside
\texttt{Llama 3.1 70B}'s responses: an absence of self-preference does not make a judge accurate.

\subsection{Readings on the Evaluated Models}
\label{appendix:readings}

Because the study's responses come from models outside the eleven, domain experts also read samples
of the evaluated models' responses at the providers' default settings, without the model's name.
A domain expert reads up to 20 flagged calculations per model, drawn from all its answered
responses.
Two domain experts of the item's branch read each of 144 final-answer verdicts on the five models
strongest at these settings (83 correct, 43 incorrect, and 18 partial) and 97 judge rulings across
all eleven models (49 reached and 48 missing), and agree with each other on 97\% of answers and
88\% of milestones.
The readings cut both ways.
The experts call 18 of their 36 readings of partial verdicts fully correct and 9 incorrect, so
on these 18 partial verdicts scoring 0.5 is conservative on average, and in a third of their
readings of missing rulings the response's route does not need the milestone, so MC is
conservative for a response that takes another route; but 15 of the 98 readings of reached rulings
do not confirm them, and 8 of the 86 readings of incorrect verdicts call the answer correct.
The readings are reported per verdict category; the evaluator's accuracy per model is not measured.

\paragraph{By Answer Kind.} \autoref{tab:answer_kinds} gives the final-answer check's agreement with
the domain experts by answer kind.
On the study it matches the experts' verdict on every scalar, array and symbolic response; multipart
responses agree least, mostly on one template whose gold answer line, the check's target, states
only the last of the results its question asks for.
Of the 44 readings that differ from the check's verdict, the check credits less than the experts in
28, more in 12, and in the other 4 the experts differ from each other or give partial credit by
judgment.
The 18 readings that call a partial verdict fully correct concern a label stated with a hedge or
followed by the criteria that decide it (6), a power of ten written once for a whole vector or a
unit note read as a yes-or-no answer (8), and an equivalent form that the number rule cannot read
(4).
The 8 readings that call an incorrect verdict correct concern an array whose last listed number, the
check's target, is an index (4) and a label answered as ``yes'' or restated after the verdict (4).
The 9 readings that call a partial verdict incorrect concern a number matched in another part's role
(4), an answer that states two values for one quantity (2), a value within the tolerance but off in
the gold's printed digits (2), and a judgment of partial credit (1).
The 6 readings that call a correct verdict incorrect concern a value one unit off in its last digit
(4) and a gold whose last digit carries the rounding of its intermediate values, on which the two
experts differ (2); the 3 that give an incorrect verdict partial credit concern a right part of an
array scored on its last value (2) and a judgment of partial credit (1).
\autoref{tab:answer_cases} shows one discrepancy in each direction.

\begin{table}[t]
    \centering
    \small
    \setlength{\tabcolsep}{4pt}
    \begin{tabular}{@{}l r r r r@{}}
        \toprule
        \rowcolor{gray!10}
        \textbf{Answer kind} & \textbf{Study} & \textbf{Three-way} & \textbf{Not partial} & \textbf{Readings} \\
        \midrule
        Scalar & 60 & 1.000 & 1.000 & 139 of 146 \\
        Multipart & 80 & 0.800 & 0.985 & 50 of 60 \\
        Vector & 60 & 0.933 & 0.949 & 12 of 16 \\
        Array & 20 & 1.000 & 1.000 & 8 of 14 \\
        Symbolic & 40 & 1.000 & 1.000 & 31 of 34 \\
        Classification & 40 & 0.975 & 1.000 & 4 of 18 \\
        \midrule
        All & 300 & 0.930 & 0.986 & 244 of 288 \\
        \bottomrule
    \end{tabular}
    \caption{\textbf{Final-answer check by answer kind.} Study: the expert study's responses of each kind and
    the check's agreement with the experts' verdict, three-way and on the responses the experts did not call
    partial. Readings: the domain experts' readings of the evaluated models' verdicts that agree with the
    check; the sample is stratified by verdict, so these are counts, not rates for a kind.}
    \label{tab:answer_kinds}
\end{table}

\begin{table*}[t]
    \centering
    \small
    \renewcommand{\arraystretch}{1.15}
    \begin{tabular}{@{}p{0.2\linewidth} p{0.12\linewidth} p{0.12\linewidth} p{0.07\linewidth} p{0.09\linewidth} p{0.26\linewidth}@{}}
        \toprule
        \rowcolor{gray!10}
        \textbf{Question} & \textbf{Final answer} & \textbf{Gold} & \textbf{Check} & \textbf{Experts} & \textbf{Why they differ} \\
        \midrule
        Average time in the system of an M/M/1/K queue (industrial, Advanced) & 3.44 minutes & 3.45 minutes & correct & incorrect (both) & the rule accepts a value one unit off in its last digit; the unrounded value, $L/\lambda_{\mathrm{eff}}$, is 3.4475 minutes, which rounds to 3.45, not 3.44 \\
        Output signal after downsampling by $M = 3$ (electrical, Advanced) & $y[n] = \cos(\pi n/8)$ & $y[n] = \cos(15\pi n/8)$ & partial & correct (both) & the same sequence, since $15\pi/8 = 2\pi - \pi/8$; the check reads the stated numbers and finds 1 of its 3 targets \\
        \bottomrule
    \end{tabular}
    \caption{\textbf{One discrepancy in each direction.} Two final answers of the evaluated models on which
    the check's verdict differs from both domain experts' readings: it credits the first more and the second
    less than they do.}
    \label{tab:answer_cases}
\end{table*}

\autoref{tab:judged_steps} gives the judged step check's flag rates on the evaluated models.
On the expert study the check's precision is 0.640 over all steps and 0.467 inside correct answers
(7 of 15 flags), so about half of its flags on correct answers mark a real error.

\section{Evaluated Models}
\label{appendix:models}

One rule selects the evaluated models (\autoref{subsec:models}): no model whose responses we used to
develop or validate the evaluator, and no model we use as a judge.
The first condition excludes, among others, the five models of the expert study (\texttt{GPT-5},
\texttt{Claude Opus 4.7}, \texttt{Gemini 3.1 Pro}, \texttt{DeepSeek-R1}, and
\texttt{Llama 3.1 70B}); the second excludes every model we use as a judge (\texttt{GPT-5},
\texttt{Claude Opus 4.5}, \texttt{Gemini 3.1 Pro}, \texttt{Grok 4.6}, \texttt{MiniMax M3}, and
\texttt{MiMo-V2.5-Pro}).
\autoref{tab:models} lists the eleven evaluated models and the two flagship anchors with their
developing organization, size, and identifiers.

\begin{table*}[t]
\centering
\resizebox{1.0\textwidth}{!}{%
\renewcommand{\arraystretch}{1.1}
\begin{tabular}{lllll}
\toprule
Model & Organization & Size & Identifier & Weights \\
\noalign{\vskip 0.5ex}\hdashline\noalign{\vskip 0.5ex}

\multicolumn{5}{l}{\textbf{Open-Weights LLMs}} \\
\noalign{\vskip 0.5ex}\hdashline\noalign{\vskip 0.5ex}
\texttt{DeepSeek V4.1 Flash} & DeepSeek & 552B (8B/16B active) & \texttt{deepseek/deepseek-v4.1-flash} & \texttt{deepseek-ai/DeepSeek-V4.1-Flash} \\
\texttt{Gemma 4 26B} & Google DeepMind & 25.2B (3.8B active) & \texttt{google/gemma-4-26b-a4b-it} & \texttt{google/gemma-4-26B-A4B-it} \\
\texttt{GLM-5.3} & Z.ai & 753B & \texttt{z-ai/glm-5.3} & \texttt{zai-org/GLM-5.3} \\
\texttt{GLM-5.3-Flash} & Z.ai & 320B (18B active) & \texttt{z-ai/glm-5.3-flash} & \texttt{zai-org/GLM-5.3-Flash} \\
\texttt{gpt-oss-20b} & OpenAI & 21B (3.6B active) & \texttt{openai/gpt-oss-20b} & \texttt{openai/gpt-oss-20b} \\
\texttt{Kimi K3} & Moonshot AI & 2.8T (104B active) & \texttt{moonshotai/kimi-k3} & \texttt{moonshotai/Kimi-K3} \\
\texttt{Muse Glimmer 30B} & Meta & 30B & \texttt{meta/muse-glimmer-30b} & \texttt{meta-models/Muse-Glimmer-30B} \\
\texttt{Qwen3-235B-2507} & Alibaba Cloud & 235B (22B active) & \texttt{qwen/qwen3-235b-a22b-2507} & \texttt{Qwen/Qwen3-235B-A22B-Instruct-2507} \\

\noalign{\vskip 0.5ex}\hdashline\noalign{\vskip 0.5ex}

\multicolumn{5}{l}{\textbf{Closed LLMs}} \\
\noalign{\vskip 0.5ex}\hdashline\noalign{\vskip 0.5ex}
\texttt{Claude Sonnet 5} & Anthropic & N/A & \texttt{anthropic/claude-sonnet-5} & -- \\
\texttt{Gemini 3.1 Flash-Lite} & Google DeepMind & N/A & \texttt{google/gemini-3.1-flash-lite} & -- \\
\texttt{GPT-5.4 mini} & OpenAI & N/A & \texttt{openai/gpt-5.4-mini} & -- \\

\noalign{\vskip 0.5ex}\hdashline\noalign{\vskip 0.5ex}

\multicolumn{5}{l}{\textbf{Flagship Anchors}} \\
\noalign{\vskip 0.5ex}\hdashline\noalign{\vskip 0.5ex}
\texttt{DeepSeek V4 Pro} & DeepSeek & 1.6T (49B active) & \texttt{deepseek/deepseek-v4-pro} & \texttt{deepseek-ai/DeepSeek-V4-Pro} \\
\texttt{GPT-5.4} & OpenAI & N/A & \texttt{openai/gpt-5.4} & -- \\

\bottomrule
\end{tabular}
}
\caption{\textbf{Evaluated models.} The eleven models of~\autoref{sec:experiments}, grouped by
access, and the two flagship anchors run on the 450-instance subset.
Size is the developer's stated parameter count, with the parameters active per token in
parentheses for mixture-of-experts models: \texttt{DeepSeek V4.1 Flash} activates 8B in prefill and
16B in decoding, and for \texttt{GLM-5.3}, whose card states no active count, we give the parameter
count of its released weights; closed models do not disclose their size.
Identifier is the model's identifier on OpenRouter, through which we query every model; Weights is
the Hugging Face repository of an open-weights model.}
\label{tab:models}
\end{table*}

\section{Experimental Setup}
\label{appendix:setup}

\subsection{Decoding and Token Use}
\label{appendix:decoding}

We query every model through OpenRouter with the prompt (\autoref{appendix:prompt}) and the output
ceiling, and set no temperature or top-p; the three models whose endpoints offer reasoning but leave
it off by default run at medium effort, and every other model decodes at its provider's defaults.
The ceiling is 32,768 tokens, except 16,384 for \texttt{Muse Glimmer 30B}, whose provider caps its
output there.
Providers serve every open-weights model at 8-bit floating-point precision or higher, and the
responses of one model can come from several providers.
A request that fails at the provider is sent again, and every model returns one response, possibly
empty, for each of the 2,250 instances.
\texttt{Qwen3-235B-2507} has no thinking mode, and \texttt{Gemma 4 26B},
\texttt{Gemini 3.1 Flash-Lite}, and \texttt{GPT-5.4 mini} return no reasoning tokens at their
providers' defaults; the other seven models return them on nearly every response.
\autoref{tab:decoding} gives each model's output ceiling, number of providers, and token use.

\paragraph{Matched Settings.} At matched settings, the configuration the paper reports, the three
models that return no reasoning tokens at their defaults but whose endpoints offer a reasoning
setting, \texttt{Gemma 4 26B}, \texttt{Gemini 3.1 Flash-Lite}, and \texttt{GPT-5.4 mini}, answer
every instance with reasoning at
medium effort and the same prompt and output ceiling, and every one of these responses carries
reasoning tokens; with reasoning on, \texttt{Gemma 4 26B} leaves 3 of its 2,250 responses empty at
the output ceiling.
Their responses at the providers' defaults (6,750, with no reasoning tokens) are the comparison
of~\autoref{tab:matched}.
The other eight models have the same responses at both settings: seven reason by default, and the
endpoint of \texttt{Qwen3-235B-2507} offers no reasoning setting.
Its sibling with a thinking mode, \texttt{Qwen3-235B-A22B-Thinking-2507}, is a different model and
is not evaluated.

\paragraph{Serving Endpoints.} \autoref{tab:providers} gives, for every model that more than one
endpoint served, the responses each endpoint served and its difference from the model's mean on the
templates both cover.
Over the 21 endpoints matched on at least 20 templates, these differences lie between $-0.025$ and
$+0.021$; the endpoints that served or matched fewer templates are marked in the table, since their
differences rest on too few templates to read.

\begin{table*}[t]
\centering
\small
\renewcommand{\arraystretch}{1.1}
\begin{tabular}{lrrrrr}
\toprule
Model & Output ceiling & Providers & Completion tokens & Reasoning tokens & Responses with \\
 & & & median / p90 & median / p90 & reasoning tokens \\
\noalign{\vskip 0.5ex}\hdashline\noalign{\vskip 0.5ex}

\multicolumn{6}{l}{\textbf{Open-Weights LLMs}} \\
\noalign{\vskip 0.5ex}\hdashline\noalign{\vskip 0.5ex}
\texttt{DeepSeek V4.1 Flash} & 32,768 & 8 & 3,060 / 10,296 & 2,380 / 9,300 & 100.0\% \\
\texttt{Gemma 4 26B}$^{\ddagger}$ & 32,768 & 2 & 5,556 / 11,174 & 4,754 / 9,905 & 100.0\% \\
\texttt{GLM-5.3} & 32,768 & 8 & 3,270 / 17,954 & 2,473 / 16,298 & 99.8\% \\
\texttt{GLM-5.3-Flash} & 32,768 & 11 & 2,500 / 11,328 & 1,785 / 10,247 & 99.9\% \\
\texttt{gpt-oss-20b} & 32,768 & 3 & 1,795 / 6,164 & 993 / 4,987 & 100.0\% \\
\texttt{Kimi K3} & 32,768 & 2 & 2,192 / 8,277 & 1,495 / 7,121 & 100.0\% \\
\texttt{Muse Glimmer 30B} & 16,384 & 1 & 2,988 / 6,436 & 1,824 / 3,853 & 100.0\% \\
\texttt{Qwen3-235B-2507} & 32,768 & 3 & 1,013 / 2,300 & 0 / 0 & 0.0\% \\

\noalign{\vskip 0.5ex}\hdashline\noalign{\vskip 0.5ex}

\multicolumn{6}{l}{\textbf{Closed LLMs}} \\
\noalign{\vskip 0.5ex}\hdashline\noalign{\vskip 0.5ex}
\texttt{Claude Sonnet 5} & 32,768 & 1 & 1,804 / 4,819 & 718 / 3,387 & 99.6\% \\
\texttt{Gemini 3.1 Flash-Lite}$^{\ddagger}$ & 32,768 & 2 & 1,485 / 2,426 & 812 / 1,373 & 100.0\% \\
\texttt{GPT-5.4 mini}$^{\ddagger}$ & 32,768 & 1 & 1,256 / 4,317 & 516 / 3,433 & 100.0\% \\

\noalign{\vskip 0.5ex}\hdashline\noalign{\vskip 0.5ex}

\multicolumn{6}{l}{\textbf{At the providers' defaults}} \\
\noalign{\vskip 0.5ex}\hdashline\noalign{\vskip 0.5ex}
\texttt{Gemma 4 26B} & 32,768 & 3 & 797 / 1,434 & 0 / 0 & 0.0\% \\
\texttt{Gemini 3.1 Flash-Lite} & 32,768 & 1 & 653 / 1,031 & 0 / 0 & 0.0\% \\
\texttt{GPT-5.4 mini} & 32,768 & 2 & 632 / 1,198 & 0 / 0 & 0.0\% \\

\bottomrule
\end{tabular}
\caption{\textbf{Decoding and token use.} Per model over its 2,250 responses at matched settings:
the output ceiling, the number of providers that served it, the median and 90th-percentile (p90)
completion and reasoning tokens as the provider reports them, and the share of responses with any
reasoning tokens; the bottom block gives the three models that return no reasoning tokens at their
providers' defaults on their responses there. $^{\ddagger}$Reasoning at medium effort.}
\label{tab:decoding}
\end{table*}

\begin{table*}[t]
\centering
\footnotesize
\renewcommand{\arraystretch}{1.1}
\begin{tabular}{l l r r r r r}
\toprule
\rowcolor{gray!10}
\textbf{Model} & \textbf{Endpoint} & \textbf{Responses} & \textbf{FAC} & \textbf{\makecell{No readable\\answer}} & \textbf{\makecell{Matched\\difference}} & \textbf{\makecell{Templates\\matched}} \\
\midrule
\texttt{DeepSeek V4.1 Flash} & CoreWeave & 782 & 0.992 & 0.1\% & $0.000$ & 141 \\
 & Parasail & 466 & 0.992 & 0.0\% & $-0.002$ & 138 \\
 & Morph & 456 & 0.991 & 0.0\% & $+0.002$ & 133 \\
 & DeepInfra & 411 & 0.984 & 0.5\% & $0.000$ & 78 \\
 & Makora & 87 & 1.000 & 0.0\% & $+0.003$ & 43 \\
 & InferenceNet$^{\S}$ & 30 & 0.800 & 16.7\% & -- & 0 \\
 & Novita$^{\S}$ & 16 & 1.000 & 0.0\% & $+0.016$ & 7 \\
 & GMICloud$^{\S}$ & 2 & 1.000 & 0.0\% & $0.000$ & 2 \\
\texttt{Kimi K3} & Morph$^{\S}$ & 2,231 & 0.980 & 0.7\% & $-0.017$ & 7 \\
 & BaseTen$^{\S}$ & 19 & 1.000 & 0.0\% & $+0.017$ & 7 \\
\texttt{GLM-5.3-Flash} & GMICloud & 981 & 0.976 & 1.5\% & $-0.003$ & 145 \\
 & Novita & 419 & 0.962 & 3.1\% & $+0.003$ & 120 \\
 & Parasail & 220 & 0.964 & 3.6\% & $+0.002$ & 75 \\
 & Sail Research & 201 & 0.965 & 2.0\% & $-0.022$ & 114 \\
 & Phala & 166 & 0.991 & 0.6\% & $+0.004$ & 74 \\
 & StreamLake & 80 & 0.950 & 3.8\% & $-0.025$ & 35 \\
 & Z.AI & 79 & 0.994 & 0.0\% & $+0.004$ & 48 \\
 & NextBit$^{\S}$ & 43 & 0.965 & 2.3\% & $+0.031$ & 13 \\
 & AtlasCloud$^{\S}$ & 42 & 1.000 & 0.0\% & $+0.014$ & 15 \\
 & SiliconFlow$^{\S}$ & 18 & 1.000 & 0.0\% & $0.000$ & 5 \\
 & Morph$^{\S}$ & 1 & 1.000 & 0.0\% & $0.000$ & 1 \\
\texttt{GLM-5.3} & Baidu & 918 & 0.956 & 3.6\% & $-0.006$ & 86 \\
 & Sail Research & 739 & 0.964 & 3.0\% & $-0.004$ & 77 \\
 & Morph & 257 & 0.940 & 5.4\% & $+0.021$ & 123 \\
 & Novita & 167 & 0.943 & 5.4\% & $-0.004$ & 41 \\
 & AtlasCloud & 137 & 0.927 & 7.3\% & $-0.010$ & 54 \\
 & AkashML$^{\S}$ & 15 & 0.267 & 73.3\% & $-0.020$ & 5 \\
 & BaseTen$^{\S}$ & 15 & 1.000 & 0.0\% & $0.000$ & 5 \\
 & GMICloud$^{\S}$ & 2 & 1.000 & 0.0\% & $0.000$ & 2 \\
\texttt{Gemma 4 26B}$^{\ddagger}$ & Io Net & 2,194 & 0.943 & 0.1\% & $-0.003$ & 36 \\
 & DekaLLM & 56 & 0.911 & 1.8\% & $+0.003$ & 36 \\
\texttt{Gemini 3.1 Flash-Lite}$^{\ddagger}$ & Google AI Studio$^{\S}$ & 2,132 & 0.913 & 0.0\% & $+0.001$ & 18 \\
 & Google$^{\S}$ & 118 & 0.975 & 0.0\% & $-0.001$ & 18 \\
\texttt{Qwen3-235B-2507} & GMICloud & 2,192 & 0.899 & 0.0\% & $-0.010$ & 35 \\
 & Parasail & 54 & 0.907 & 0.0\% & $+0.006$ & 35 \\
 & Nebius$^{\S}$ & 4 & 1.000 & 0.0\% & $+0.182$ & 1 \\
\texttt{gpt-oss-20b} & Darkbloom$^{\S}$ & 2,240 & 0.808 & 2.6\% & $-0.228$ & 9 \\
 & DekaLLM$^{\S}$ & 9 & 0.889 & 0.0\% & $+0.239$ & 8 \\
 & DeepInfra$^{\S}$ & 1 & 1.000 & 0.0\% & $+0.143$ & 1 \\
\bottomrule
\end{tabular}
\caption{\textbf{Serving endpoints.} The endpoints behind the responses of the eight models that
more than one endpoint served: the responses each served, their FAC and their share with no readable
answer, and the template-matched difference, the endpoint's mean minus the model's mean on the same
templates, over the templates both cover. The raw FAC per endpoint is not comparable across
endpoints, since OpenRouter assigns instances to endpoints unevenly; $^{\S}$marks the 18 endpoints
that served or matched fewer than 20 templates, whose difference rests on too few templates to read.
$^{\ddagger}$Responses with reasoning at medium effort; a model that one endpoint served is not
listed.}
\label{tab:providers}
\end{table*}

\subsection{Inference Prompt and Output Processing}
\label{appendix:prompt}

To ensure consistent and parsable outputs across all models, we use a standardized zero-shot prompt,
sent as a single user message with the instance's question in place of \texttt{\{question\}}.
This prompt explicitly instructs the model to structure its response with specific headings and to
format the final result for automated extraction.

\begin{quote}
\small
\begin{verbatim}
You are an expert engineer. Solve the following
problem by providing a detailed, structured
solution. Use the exact headings and
formatting provided below.

## Given
List all known variables and their
values with units.

## Find
State the variable(s) to be calculated.

## Formulae
Write down all necessary governing equations
before substituting any values.

## Solution
Provide a step-by-step calculation. Each step
must start on a new line and be formatted exactly
as '**Step X:**', where X is the step number.
Show the substitution of values into
the formulae clearly.

## Final Answer
State the final numerical result with its units
in the format: **Answer:** [value] [units]

Problem:
{question}
\end{verbatim}
\end{quote}

Each check reads the response as~\autoref{appendix:scoring} describes:
\begin{itemize}
    \item \textbf{Final answer}: the check reads the answer after the last final-answer heading or
    the \texttt{**Answer:**} marker, and still reads a response that has neither.
    \item \textbf{Milestones and arithmetic}: milestone matching searches the whole response, in any
    order, and the arithmetic check reads every calculation the response displays.
    \item \textbf{Judged steps}: the judged step check splits the response into steps the same way
    as in the expert study.
\end{itemize}

\subsection{Statistical Protocol}
\label{appendix:protocol}

The template is the unit of analysis (\autoref{subsec:setup}): every template has 15 instances, so
Final Answer Accuracy over the 2,250 instances equals the mean of the 150 template means.
On this unit, we estimate and test each quantity as follows:
\begin{itemize}
    \item \textbf{Intervals}: every 95\% interval is a percentile bootstrap that resamples templates
    10,000 times, and a bounded change also carries its 90\% interval.
    \item \textbf{Two models}: we compare them on the per-template differences of their scores, with
    a paired bootstrap interval and a sign-flip permutation test with 100,000 sign flips.
    Strict accuracy, which scores a partial answer 0, and McNemar's exact test on the paired
    verdicts check the comparisons of Final Answer Accuracy; a Wilcoxon signed-rank test checks
    those of Milestone Coverage, which use the 148 templates with milestones.
    \item \textbf{Level gaps}: a model's level gap is its mean on the 58 Easy templates minus its
    mean on the 34 Advanced ones; we test it with Welch's $t$-test on the template
    means~\citep{welch1947generalization}, because scores vary more across Advanced templates for
    nine of the eleven models, and its interval resamples templates within each level.
    \item \textbf{Orderings}: we compare two orderings of the models with Kendall's $\tau$ and a
    bootstrap interval.
    \item \textbf{Multiple tests}: Holm's correction~\citep{holm1979simple} applies within each
    family of tests, such as the 55 pairs of models for each measure or the eleven level gaps.
    \item \textbf{Providers' defaults}: the three models' default responses, with the other eight
    models, form a second family of 55 pairs, tested and corrected in the same way, and a model's
    change between the two settings is paired by instance.
    \item \textbf{Detectable differences}: at 80\% power and a two-sided level of 0.05, a paired
    comparison detects $2.80\,s/\sqrt{n}$, with $s$ the standard deviation of the $n$ per-template
    differences, and a level gap $2.80\,\sigma\sqrt{1/58 + 1/34}$, with $\sigma$ the pooled
    standard deviation of the template means within a level.
    \item \textbf{Bounded changes}: a change counts as bounded when its 90\% interval lies within
    $\pm 0.05$, which is two one-sided tests at 5\%.
\end{itemize}

\section{Full Results}
\label{appendix:results}

This section holds the tests and measures behind~\autoref{subsec:overall_performance}
and~\autoref{subsec:difficulty_level}.
Intervals resample templates, every family of tests is Holm-corrected, and each FAC comparison and
level gap that does not hold comes with the smallest difference the design detects at 80\% power
(\autoref{appendix:protocol}).

\paragraph{Pairwise Comparisons.} The final answer separates most pairs of models but few at the
top, and MC separates fewer pairs, none that the final answer does not.
Of the 55 pairs, 36 differ on FAC after Holm correction under the sign-flip test over templates:
three inside the first five (\texttt{DeepSeek V4.1 Flash} from \texttt{GLM-5.3-Flash},
\texttt{DeepSeek V4.1 Flash} from \texttt{GPT-5.4 mini}, and \texttt{Claude Sonnet 5} from
\texttt{GPT-5.4 mini}), none between \texttt{Muse Glimmer 30B} and the first five, 27 of the 30
between the first six and the last five, and six inside the last five.
Every non-significant FAC difference lies below what its pair detects (0.013 to 0.039).
On MC, 15 pairs differ, each of them also on FAC: the three highest models,
\texttt{Claude Sonnet 5}, \texttt{GLM-5.3-Flash}, and \texttt{Kimi K3}, do not separate, nor do
\texttt{Claude Sonnet 5} and \texttt{DeepSeek V4.1 Flash}, which FAC does not separate either;
Kendall's $\tau$~\citep{kendall1938} between the FAC and MC orderings is 0.709 (95\% interval 0.345
to 0.818).
Strict FAC, which scores a partial answer 0, separates 29 pairs and agrees with FAC on 48 of the 55;
a Wilcoxon test~\citep{wilcoxon1945} separates 17 on MC; and McNemar's exact
test~\citep{mcnemar1947}, which treats the 2,250 instances as independent, separates 45.

\paragraph{Matched Settings.} Reasoning raises all three models run with it (Holm-adjusted $p$
$<$0.001), and for \texttt{GPT-5.4 mini} and \texttt{Gemma 4 26B} the 90\% interval lies wholly
above $+0.05$, the margin of~\autoref{appendix:protocol}.
It also reorders the models: at the providers' defaults the final answer sorts them into two tiers,
an upper six each of which differs from each of a lower five, while at matched settings, the
configuration the evaluation reports, \texttt{GPT-5.4 mini} joins the first five
(\autoref{tab:matched}; settings in~\autoref{appendix:decoding}).
\autoref{appendix:experiments} reads these gains beside the reasoning-effort experiment, and the
expert readings of wrong answers (\autoref{appendix:error_analysis}) are of the responses at the
providers' defaults, so they do not describe the three models at matched settings.

\paragraph{Consistency Within a Template.} The two lowest models, \texttt{Qwen3-235B-2507} and
\texttt{gpt-oss-20b}, solve some but not all instances of more single-path templates than any other
model: they often solve a derivation on one instance and fail it on another.
We count, per model, the templates whose instances the model solves on all, on some, and on none,
separately for the 58 templates whose instances all follow one derivation and for the 92 others.
Across the models, the share of templates solved on every instance runs from 34\% to 93\% on the
single-path templates and from 29\% to 88\% on the others, and the share solved on some but not
all from 7\% to 59\% and from 12\% to 70\%.
No model fails every instance of more than four of the 58 single-path templates or two of the 92
others; the counts show how consistent a model is within a template, not why.

\begin{table*}[t]
\centering
\footnotesize
\renewcommand{\arraystretch}{1.1}
\adjustbox{max width=\textwidth}{%
\begin{tabular}{l c c c c l c r c r c}
\toprule
\rowcolor{gray!10}
 & \multicolumn{2}{c}{\textbf{Providers' defaults}} & \multicolumn{3}{c}{\textbf{Matched settings}} & \multicolumn{5}{c}{\textbf{Paired change, reasoning minus default}} \\
\cmidrule(lr){2-3}\cmidrule(lr){4-6}\cmidrule(lr){7-11}
\textbf{Model} & \textbf{FAC} & \textbf{Tier} & \textbf{FAC} & \textbf{Tier} & \textbf{Setting} & \textbf{\makecell{Change\\(95\% interval)}} & \textbf{$p$ (Holm)} & \textbf{90\% interval} & \textbf{\makecell{MC\\change}} & \textbf{\makecell{Empty responses,\\default / reasoning}} \\
\midrule
\texttt{DeepSeek V4.1 Flash} & 0.988 {\scriptsize\textcolor{gray}{(0.979--0.995)}} & a & 0.988 {\scriptsize\textcolor{gray}{(0.979--0.995)}} & a & reasons by default & -- & -- & -- & -- & -- \\
\texttt{Claude Sonnet 5} & 0.987 {\scriptsize\textcolor{gray}{(0.980--0.993)}} & ab & 0.987 {\scriptsize\textcolor{gray}{(0.980--0.993)}} & ab & reasons by default & -- & -- & -- & -- & -- \\
\texttt{Kimi K3} & 0.980 {\scriptsize\textcolor{gray}{(0.967--0.990)}} & ab & 0.980 {\scriptsize\textcolor{gray}{(0.967--0.990)}} & abc & reasons by default & -- & -- & -- & -- & -- \\
\texttt{GLM-5.3-Flash} & 0.972 {\scriptsize\textcolor{gray}{(0.953--0.987)}} & b & 0.972 {\scriptsize\textcolor{gray}{(0.953--0.987)}} & bc & reasons by default & -- & -- & -- & -- & -- \\
\texttt{GPT-5.4 mini}$^{\ddagger}$ & 0.852 {\scriptsize\textcolor{gray}{(0.810--0.890)}} & de & 0.969 {\scriptsize\textcolor{gray}{(0.958--0.980)}} & cde & effort medium & $+0.118$ (0.082--0.156) & $<$0.001 & 0.088--0.149 & $+0.058$ & 0 / 0 \\
\texttt{Muse Glimmer 30B} & 0.969 {\scriptsize\textcolor{gray}{(0.949--0.985)}} & abc & 0.969 {\scriptsize\textcolor{gray}{(0.949--0.985)}} & abcd & reasons by default & -- & -- & -- & -- & -- \\
\texttt{GLM-5.3}$^{\dagger}$ & 0.950 {\scriptsize\textcolor{gray}{(0.921--0.975)}} & c & 0.950 {\scriptsize\textcolor{gray}{(0.921--0.975)}} & def & reasons by default & -- & -- & -- & -- & -- \\
\texttt{Gemma 4 26B}$^{\ddagger}$ & 0.872 {\scriptsize\textcolor{gray}{(0.835--0.905)}} & d & 0.942 {\scriptsize\textcolor{gray}{(0.916--0.964)}} & ef & effort medium & $+0.070$ (0.049--0.092) & $<$0.001 & 0.053--0.088 & $+0.008$ & 0 / 3 \\
\texttt{Gemini 3.1 Flash-Lite}$^{\ddagger}$ & 0.878 {\scriptsize\textcolor{gray}{(0.837--0.915)}} & d & 0.917 {\scriptsize\textcolor{gray}{(0.884--0.945)}} & fg & effort medium & $+0.038$ (0.023--0.056) & $<$0.001 & 0.025--0.053 & $+0.016$ & 0 / 0 \\
\texttt{Qwen3-235B-2507}$^{\ast}$ & 0.899 {\scriptsize\textcolor{gray}{(0.869--0.926)}} & d & 0.899 {\scriptsize\textcolor{gray}{(0.869--0.926)}} & g & none offered & -- & -- & -- & -- & -- \\
\texttt{gpt-oss-20b} & 0.808 {\scriptsize\textcolor{gray}{(0.764--0.849)}} & e & 0.808 {\scriptsize\textcolor{gray}{(0.764--0.849)}} & h & reasons by default & -- & -- & -- & -- & -- \\
\bottomrule
\end{tabular}
}
\caption{\textbf{Matched settings.} FAC at the providers' default settings and at matched settings,
with the tier letters of each configuration's own 55-pair family (Holm-corrected sign-flip tests
over templates; models sharing a letter do not differ). Setting: what the matched configuration
runs. For the three models run with reasoning on, the change from their default rows over the same
2,250 instances, paired by instance: with its 95\% interval, the Holm-adjusted $p$ of the sign-flip
test over templates, the 90\% interval read against the $\pm 0.05$ margin, the change in MC, and the
empty responses in each configuration. Kendall's $\tau$ between the two orderings is 0.709 (95\%
interval 0.636 to 0.818). $^{\ast}$No reasoning setting offered. $^{\ddagger}$Run with reasoning at
medium effort.}
\label{tab:matched}
\end{table*}

\paragraph{Scoring Rules.} The tolerance, and which answers and templates count, barely move the
ordering.
Halving the answer tolerance swaps one pair of models and doubling it swaps two, and no swapped pair
differs after correction ($\tau = 0.964$ and $0.927$ against the ordering as scored).
Requiring a correct rounding at the displayed precision rather than accepting one unit either way,
crediting a requested quantity stated in the body but not on the answer line, and leaving out the
four templates answerable from their wording or the nine with symbolic answers move no model's FAC
by more than 0.028 and keep $\tau$ at 0.964 or above.
The ordering changes most when only readable answers count ($\tau = 0.636$), chiefly because
\texttt{GLM-5.3}'s 99 empty responses then drop out.
\autoref{tab:scoring_variants} changes one clause of the final-answer rule at a time, and
\autoref{appendix:scoring} reads it.

\begin{table*}[t]
\centering
\footnotesize
\renewcommand{\arraystretch}{1.1}
\adjustbox{max width=\textwidth}{%
\begin{tabular}{l r c c c c r r r r}
\toprule
\rowcolor{gray!10}
 & & \multicolumn{4}{c}{\textbf{FAC under the variant (verdicts up/down)}} & \multicolumn{4}{c}{\textbf{Relative error of accepted answers}} \\
\cmidrule(lr){3-6}\cmidrule(lr){7-10}
\textbf{Model} & \textbf{\makecell{FAC\\as scored}} & \textbf{\makecell{Absolute-value\\clause off}} & \textbf{\makecell{Last-digit term\\not capped}} & \textbf{\makecell{Prescribed digits\\relaxed}} & \textbf{\makecell{Proportional\\partial credit}} & \textbf{$\le$0.2\%} & \textbf{0.2--1\%} & \textbf{1--5\%} & \textbf{$>$5\%} \\
\midrule
\texttt{DeepSeek V4.1 Flash} & 0.988 & 0.985 (0/8) & 0.988 (1/0) & 0.992 (8/0) & 0.988 (2/2) & 2,073 & 78 & 1 & 5 \\
\texttt{Claude Sonnet 5} & 0.987 & 0.979 (0/24) & 0.987 (1/0) & 0.992 (11/0) & 0.987 (3/2) & 2,067 & 86 & 3 & 5 \\
\texttt{Kimi K3} & 0.980 & 0.964 (0/42) & 0.982 (6/0) & 0.983 (6/0) & 0.980 (0/6) & 2,047 & 89 & 2 & 5 \\
\texttt{GLM-5.3-Flash} & 0.972 & 0.942 (0/76) & 0.974 (5/0) & 0.974 (4/0) & 0.972 (2/2) & 2,041 & 93 & 4 & 3 \\
\texttt{GPT-5.4 mini} & 0.969 & 0.964 (0/15) & 0.970 (1/0) & 0.976 (15/0) & 0.969 (2/2) & 1,914 & 184 & 4 & 8 \\
\texttt{Muse Glimmer 30B} & 0.969 & 0.967 (0/6) & 0.974 (13/0) & 0.971 (5/0) & 0.969 (2/1) & 1,829 & 294 & 3 & 3 \\
\texttt{GLM-5.3} & 0.950 & 0.924 (0/69) & 0.951 (4/0) & 0.950 (0/0) & 0.950 (4/2) & 1,988 & 95 & 3 & 3 \\
\texttt{Gemma 4 26B} & 0.942 & 0.937 (0/17) & 0.943 (3/0) & 0.950 (18/0) & 0.943 (9/1) & 1,947 & 79 & 3 & 9 \\
\texttt{Gemini 3.1 Flash-Lite} & 0.917 & 0.913 (0/13) & 0.918 (3/0) & 0.923 (14/0) & 0.916 (2/6) & 1,919 & 72 & 2 & 5 \\
\texttt{Qwen3-235B-2507} & 0.899 & 0.898 (0/4) & 0.900 (4/0) & 0.907 (18/0) & 0.897 (5/26) & 1,812 & 100 & 4 & 6 \\
\texttt{gpt-oss-20b} & 0.808 & 0.793 (0/36) & 0.815 (22/0) & 0.824 (37/0) & 0.808 (7/2) & 1,439 & 298 & 5 & 5 \\
\midrule
Kendall's $\tau$ with FAC as scored & & 0.782 & 0.964 & 0.917 & 1.000 & & & & \\
\bottomrule
\end{tabular}
}
\caption{\textbf{Scoring-rule variants.} The final-answer rule re-applied to every response of the
evaluation with one clause changed at a time: the absolute-value clause off, the response's
last-digit term not capped at one hundredth of the target, the exact-digit requirement of the
prescribing templates relaxed to the tolerance, and partial credit in proportion to the parts
matched instead of one half. Each cell gives FAC under the variant and the verdicts that rise and
fall; the last row Kendall's $\tau$ of the variant's ordering with the ordering as scored. Right:
the accepted answers by their relative error to the target.}
\label{tab:scoring_variants}
\end{table*}

\paragraph{Coverage and What Marks a Wrong Answer.} For every model, wrong answers state far more
gold milestones than the chance floor, and for most models a milestone the judge rules missing marks
most of them (\autoref{tab:coverage}).
The flag rates, arithmetic and judged, do not rank the models: the two checks together recall fewer
than half of the flawed steps inside correct answers (\autoref{appendix:validation}), the precision
of the arithmetic flags varies across models, and the arithmetic check reads more calculations where
a model writes more arithmetic out.
Nor does MC reward writing more: with template fixed effects, its slope on the number of values a
response displays is barely positive (\autoref{tab:coverage_variants}), and within each model
coverage falls slightly as a response displays more values (Spearman's
$\rho$~\citep{spearman1904} $-0.21$ to $-0.07$), consistent with lower coverage on templates with
more steps ($\rho$ $-0.09$ to $-0.28$ between a template's mean coverage and its number of steps).

Under the four definitions of~\autoref{tab:coverage_variants}, what separates
\texttt{Claude Sonnet 5} from \texttt{DeepSeek V4.1 Flash} is how closely a response follows the
gold derivation's route, not whether it reaches the quantities its own route needs: the two differ
by matching alone and on intermediate milestones only, but not as scored or once the milestones the
judge rules not needed leave the denominator.
The judge raises every model's coverage by 0.019 to 0.045 over matching alone and narrows the
differences between models: matching alone separates 20 of the 55 pairs, MC as scored 15, and a
second judge on a sample shifts a model's coverage by $-0.018$ to $+0.019$
(\autoref{appendix:validation}).
A mechanical check classifies every arithmetic flag on a correct answer: 12 of the 3,077 (0.4\%;
at most 2.5\% for any model) are carried precision, a printed result that is a correct rounding of
the value recomputed from the unrounded values earlier in the response, and of the 171 slips the
domain expert confirmed, none of the 168 it can classify is.

\begin{table*}[t]
\centering
\footnotesize
\renewcommand{\arraystretch}{1.1}
\adjustbox{max width=\textwidth}{%
\begin{tabular}{l r r r r r c r r c c}
\toprule
\rowcolor{gray!10}
 & \multicolumn{6}{c}{\textbf{Wrong answers}} & \multicolumn{2}{c}{\textbf{All responses}} & \multicolumn{2}{c}{\textbf{Correct answers}} \\
\cmidrule(lr){2-7}\cmidrule(lr){8-9}\cmidrule(lr){10-11}
\textbf{Model} & \textbf{Number} & \textbf{\makecell{MC by\\matching}} & \textbf{\makecell{Chance\\floor}} & \textbf{\makecell{All\\reached}} & \textbf{\makecell{Milestone\\missing}} & \textbf{\makecell{Any step\\flagged}} & \textbf{\makecell{Judge-decided\\milestones}} & \textbf{\makecell{Steps flagged\\per response}} & \textbf{\makecell{Judged step\\flags}} & \textbf{\makecell{Arithmetic flag\\precision ($n$)}} \\
\midrule
\texttt{DeepSeek V4.1 Flash} & 12 & 0.667 & 0.141 & 0.083 & 0.833 & 0.417 {\scriptsize\textcolor{gray}{(0.000--0.636)}} & 0.169 & 0.03 & 0.010 {\scriptsize\textcolor{gray}{(0.005--0.015)}} & 1.000 (6) \\
\texttt{Claude Sonnet 5} & 20 & 0.850 & 0.150 & 0.550 & 0.200 & 0.650 {\scriptsize\textcolor{gray}{(0.412--0.875)}} & 0.125 & 0.28 & 0.054 {\scriptsize\textcolor{gray}{(0.041--0.068)}} & 0.950 (20) \\
\texttt{Kimi K3} & 20 & 0.584 & 0.113 & 0.200 & 0.636 & 0.545 {\scriptsize\textcolor{gray}{(0.350--0.765)}} & 0.150 & 0.05 & 0.017 {\scriptsize\textcolor{gray}{(0.011--0.024)}} & 1.000 (4) \\
\texttt{GLM-5.3-Flash} & 12 & 0.378 & 0.134 & 0.083 & 0.750 & 0.333 {\scriptsize\textcolor{gray}{(0.077--0.667)}} & 0.126 & 0.07 & 0.032 {\scriptsize\textcolor{gray}{(0.023--0.042)}} & 0.632 (19) \\
\texttt{GPT-5.4 mini} & 61 & 0.640 & 0.137 & 0.311 & 0.459 & 0.770 {\scriptsize\textcolor{gray}{(0.646--0.885)}} & 0.184 & 0.27 & 0.115 {\scriptsize\textcolor{gray}{(0.091--0.140)}} & -- \\
\texttt{Muse Glimmer 30B} & 33 & 0.583 & 0.178 & 0.364 & 0.588 & 0.529 {\scriptsize\textcolor{gray}{(0.208--0.743)}} & 0.167 & 0.15 & 0.059 {\scriptsize\textcolor{gray}{(0.046--0.074)}} & 0.800 (20) \\
\texttt{GLM-5.3} & 8 & 0.486 & 0.146 & 0.500 & 0.500 & 0.250 {\scriptsize\textcolor{gray}{(0.000--0.625)}} & 0.107 & 0.06 & 0.026 {\scriptsize\textcolor{gray}{(0.016--0.036)}} & 0.900 (20) \\
\texttt{Gemma 4 26B} & 116 & 0.548 & 0.099 & 0.181 & 0.724 & 0.853 {\scriptsize\textcolor{gray}{(0.731--0.935)}} & 0.186 & 0.35 & 0.074 {\scriptsize\textcolor{gray}{(0.053--0.097)}} & -- \\
\texttt{Gemini 3.1 Flash-Lite} & 174 & 0.515 & 0.116 & 0.161 & 0.750 & 0.875 {\scriptsize\textcolor{gray}{(0.803--0.927)}} & 0.187 & 0.37 & 0.093 {\scriptsize\textcolor{gray}{(0.068--0.121)}} & -- \\
\texttt{Qwen3-235B-2507} & 173 & 0.586 & 0.175 & 0.249 & 0.631 & 0.864 {\scriptsize\textcolor{gray}{(0.776--0.930)}} & 0.158 & 0.86 & 0.107 {\scriptsize\textcolor{gray}{(0.084--0.131)}} & 0.950 (20) \\
\texttt{gpt-oss-20b} & 335 & 0.455 & 0.117 & 0.146 & 0.777 & 0.916 {\scriptsize\textcolor{gray}{(0.871--0.950)}} & 0.238 & 1.01 & 0.206 {\scriptsize\textcolor{gray}{(0.177--0.237)}} & 0.950 (20) \\
\bottomrule
\end{tabular}
}
\caption{\textbf{Milestone Coverage behind wrong answers, and the step flags.} Per model, the
readable wrong answers with milestones: how many; MC by matching alone (with the judge:
\autoref{tab:coverage_variants}); the chance floor, the same responses matched against a sibling
instance's milestones; and the share that reach every milestone. Over the answered wrong answers:
the share with a milestone the judge rules missing, and the share with at least one step either
check flags. Over all answered responses: the share of milestones the judge rather than the matcher
decides, and the steps either check flags per response. Over correct answers: the share with at
least one step the judge flags, and the precision of the arithmetic flags a domain expert read, with
their number (171 of 189 were slips overall; the expert read responses at the providers' default
settings, so the column is empty for \texttt{GPT-5.4 mini}, \texttt{Gemma 4 26B}, and
\texttt{Gemini 3.1 Flash-Lite}, run here with reasoning at medium effort). The two shares of
responses with a flagged step carry 95\% intervals over templates. The judged step check runs the
judge at the provider's default settings, so its flags are judged, not verified: in the expert
study, 7 of the 15 flags it raises inside correct answers are errors (precision 0.47), and together
with the arithmetic check it finds 0.36 of the flawed steps there at precision 0.70
(\autoref{tab:validation_results}). The 57 instances without milestones are left out of the coverage
columns.}
\label{tab:coverage}
\label{tab:judged_steps}
\end{table*}

\begin{table*}[t]
\centering
\footnotesize
\renewcommand{\arraystretch}{1.1}
\adjustbox{max width=\textwidth}{%
\begin{tabular}{l c r c c r c c r c c r c }
\toprule
\rowcolor{gray!10}
 & \multicolumn{3}{c}{\textbf{As scored}} & \multicolumn{3}{c}{\textbf{Matching alone}} & \multicolumn{3}{c}{\textbf{Route-adjusted}} & \multicolumn{3}{c}{\textbf{Intermediate only}} \\
\cmidrule(lr){2-4}\cmidrule(lr){5-7}\cmidrule(lr){8-10}\cmidrule(lr){11-13}
\textbf{Model} & \textbf{All} & \textbf{Wrong} & \textbf{Tier} & \textbf{All} & \textbf{Wrong} & \textbf{Tier} & \textbf{All} & \textbf{Wrong} & \textbf{Tier} & \textbf{All} & \textbf{Wrong} & \textbf{Tier} \\
\midrule
\texttt{DeepSeek V4.1 Flash} & 0.899 {\scriptsize\textcolor{gray}{(0.874--0.922)}} & 0.667 & abc & 0.861 {\scriptsize\textcolor{gray}{(0.831--0.890)}} & 0.667 & bcd & 0.982 {\scriptsize\textcolor{gray}{(0.974--0.988)}} & 0.711 & a & 0.822 {\scriptsize\textcolor{gray}{(0.778--0.864)}} & 0.644 & bc \\
\texttt{Claude Sonnet 5} & 0.922 {\scriptsize\textcolor{gray}{(0.902--0.942)}} & 0.869 & a & 0.903 {\scriptsize\textcolor{gray}{(0.880--0.925)}} & 0.850 & a & 0.984 {\scriptsize\textcolor{gray}{(0.978--0.990)}} & 0.935 & ab & 0.859 {\scriptsize\textcolor{gray}{(0.817--0.897)}} & 0.834 & a \\
\texttt{Kimi K3} & 0.912 {\scriptsize\textcolor{gray}{(0.888--0.934)}} & 0.584 & ab & 0.877 {\scriptsize\textcolor{gray}{(0.850--0.902)}} & 0.584 & bc & 0.973 {\scriptsize\textcolor{gray}{(0.961--0.983)}} & 0.627 & abc & 0.847 {\scriptsize\textcolor{gray}{(0.806--0.885)}} & 0.646 & ab \\
\texttt{GLM-5.3-Flash} & 0.912 {\scriptsize\textcolor{gray}{(0.886--0.936)}} & 0.458 & abc & 0.891 {\scriptsize\textcolor{gray}{(0.864--0.916)}} & 0.378 & ab & 0.966 {\scriptsize\textcolor{gray}{(0.947--0.981)}} & 0.524 & abc & 0.859 {\scriptsize\textcolor{gray}{(0.818--0.898)}} & 0.462 & ab \\
\texttt{GPT-5.4 mini} & 0.892 {\scriptsize\textcolor{gray}{(0.866--0.917)}} & 0.689 & bc & 0.847 {\scriptsize\textcolor{gray}{(0.815--0.877)}} & 0.640 & d & 0.972 {\scriptsize\textcolor{gray}{(0.962--0.981)}} & 0.819 & abc & 0.817 {\scriptsize\textcolor{gray}{(0.771--0.860)}} & 0.711 & bc \\
\texttt{Muse Glimmer 30B} & 0.897 {\scriptsize\textcolor{gray}{(0.871--0.922)}} & 0.653 & abc & 0.862 {\scriptsize\textcolor{gray}{(0.832--0.890)}} & 0.583 & bcd & 0.962 {\scriptsize\textcolor{gray}{(0.944--0.977)}} & 0.681 & abc & 0.830 {\scriptsize\textcolor{gray}{(0.787--0.871)}} & 0.788 & ab \\
\texttt{GLM-5.3} & 0.900 {\scriptsize\textcolor{gray}{(0.866--0.930)}} & 0.596 & abc & 0.881 {\scriptsize\textcolor{gray}{(0.846--0.912)}} & 0.486 & abcd & 0.946 {\scriptsize\textcolor{gray}{(0.916--0.971)}} & 0.596 & bc & 0.852 {\scriptsize\textcolor{gray}{(0.807--0.893)}} & 0.825 & ab \\
\texttt{Gemma 4 26B} & 0.884 {\scriptsize\textcolor{gray}{(0.856--0.910)}} & 0.569 & bc & 0.850 {\scriptsize\textcolor{gray}{(0.819--0.880)}} & 0.548 & cd & 0.959 {\scriptsize\textcolor{gray}{(0.944--0.973)}} & 0.631 & c & 0.813 {\scriptsize\textcolor{gray}{(0.770--0.853)}} & 0.603 & bc \\
\texttt{Gemini 3.1 Flash-Lite} & 0.881 {\scriptsize\textcolor{gray}{(0.851--0.908)}} & 0.540 & c & 0.855 {\scriptsize\textcolor{gray}{(0.824--0.886)}} & 0.515 & bcd & 0.953 {\scriptsize\textcolor{gray}{(0.935--0.969)}} & 0.603 & c & 0.815 {\scriptsize\textcolor{gray}{(0.770--0.856)}} & 0.574 & abc \\
\texttt{Qwen3-235B-2507} & 0.892 {\scriptsize\textcolor{gray}{(0.867--0.916)}} & 0.616 & bc & 0.866 {\scriptsize\textcolor{gray}{(0.836--0.894)}} & 0.586 & bcd & 0.956 {\scriptsize\textcolor{gray}{(0.942--0.969)}} & 0.675 & c & 0.834 {\scriptsize\textcolor{gray}{(0.792--0.873)}} & 0.673 & ab \\
\texttt{gpt-oss-20b} & 0.812 {\scriptsize\textcolor{gray}{(0.775--0.847)}} & 0.465 & d & 0.790 {\scriptsize\textcolor{gray}{(0.752--0.826)}} & 0.455 & e & 0.872 {\scriptsize\textcolor{gray}{(0.840--0.902)}} & 0.509 & d & 0.759 {\scriptsize\textcolor{gray}{(0.711--0.805)}} & 0.534 & c \\
\bottomrule
\end{tabular}
}
\caption{\textbf{Milestone Coverage under four definitions.} Per model: as scored (the judge's
rulings added to the matcher's), by matching alone, route-adjusted (milestones the judge rules not
needed leave the denominator), and intermediate only (the milestones that state the final answer
leave the set). All: the mean over all responses with milestones, with its 95\% interval over
templates; wrong: the mean over wrong answers; tier: models sharing a letter do not differ after
Holm's correction over the pairwise sign-flip tests of that definition. Intermediate only: the
answer milestones are those whose value equals a number of the answer check's targets under its unit
factors; once they leave the set, 502 instances have no milestone left (57 have none at all).
\texttt{Claude Sonnet 5} against \texttt{DeepSeek V4.1 Flash}: Holm-adjusted $p$ 0.069 as scored,
0.001 by matching alone, 1.000 route-adjusted, 0.032 intermediate only. The slope of MC on the
number of values a response displays, with template fixed effects, is $+0.0020$ (MC per 10 numeric
values shown; 95\% interval 0.0005 to 0.0041).}
\label{tab:coverage_variants}
\end{table*}

\paragraph{Difficulty Level and Depth.} The level gap is model-specific, and for \texttt{GLM-5.3} it
comes from empty responses: over the responses with a readable answer, its gap is $-0.004$
(Holm-adjusted Welch $p$ 1.000).
We read Welch's test, since the permutation of level labels assumes equal spread across levels and
the Advanced level, with fewer templates, varies more than the Easy level for nine of the eleven
models; under that permutation the gap holds for four models (\autoref{tab:level_gap}).
The share scored 0 counts responses with no readable answer as well as wrong answers, so the last
two columns of~\autoref{tab:level_gap} relate a wrong answer to the depth of the gold derivation
over readable responses only, with the answer kind as a covariate: the logistic slope holds after
Holm correction for three of the eleven models.

\begin{table*}[t]
\centering
\footnotesize
\renewcommand{\arraystretch}{1.1}
\adjustbox{max width=\textwidth}{%
\begin{tabular}{l r r r c r r r r r r r r c r}
\toprule
\rowcolor{gray!10}
 & \multicolumn{3}{c}{\textbf{FAC by level}} & \multicolumn{4}{c}{\textbf{Easy minus Advanced}} & \multicolumn{5}{c}{\textbf{Share scored 0, by gold milestones}} & \multicolumn{2}{c}{\textbf{Logistic slope}} \\
\cmidrule(lr){2-4}\cmidrule(lr){5-8}\cmidrule(lr){9-13}\cmidrule(lr){14-15}
\textbf{Model} & \textbf{Easy} & \textbf{Intermediate} & \textbf{Advanced} & \textbf{Gap (95\% interval)} & \textbf{Welch $p$} & \textbf{\makecell{Permutation\\$p$}} & \textbf{Detectable} & \textbf{1} & \textbf{2} & \textbf{3} & \textbf{4--5} & \textbf{6+} & \textbf{\makecell{Per milestone\\(95\% interval)}} & \textbf{$p$ (Holm)} \\
\midrule
\texttt{DeepSeek V4.1 Flash} & 0.995 & 0.986 & 0.979 & $+0.016$ ($-0.001$ to 0.041) & 0.791 & 0.341 & 0.024 & 0.000 & 0.000 & 0.000 & 0.006 & 0.032 & $+0.504$ (0.218--0.791) & 0.005 \\
\texttt{Claude Sonnet 5} & 0.993 & 0.983 & 0.984 & $+0.008$ ($-0.007$ to 0.024) & 1.000 & 1.000 & 0.022 & 0.000 & 0.017 & 0.003 & 0.013 & 0.011 & $+0.023$ ($-0.194$ to 0.241) & 1.000 \\
\texttt{Kimi K3} & 0.992 & 0.985 & 0.953 & $+0.039$ ($-0.002$ to 0.094) & 0.791 & 0.238 & 0.055 & 0.000 & 0.006 & 0.014 & 0.028 & 0.025 & $+0.168$ ($-0.031$ to 0.367) & 0.493 \\
\texttt{GLM-5.3-Flash} & 0.994 & 0.970 & 0.939 & $+0.055$ (0.012--0.113) & 0.368 & 0.021 & 0.057 & 0.003 & 0.017 & 0.011 & 0.022 & 0.061 & $+0.189$ ($-0.110$ to 0.487) & 0.862 \\
\texttt{GPT-5.4 mini} & 0.982 & 0.953 & 0.975 & $+0.006$ ($-0.013$ to 0.028) & 1.000 & 1.000 & 0.029 & 0.007 & 0.022 & 0.014 & 0.048 & 0.034 & $+0.137$ (0.005--0.268) & 0.288 \\
\texttt{Muse Glimmer 30B} & 0.976 & 0.968 & 0.958 & $+0.019$ ($-0.026$ to 0.067) & 1.000 & 1.000 & 0.066 & 0.003 & 0.017 & 0.038 & 0.030 & 0.047 & $+0.072$ ($-0.159$ to 0.302) & 1.000 \\
\texttt{GLM-5.3} & 0.984 & 0.966 & 0.865 & $+0.120$ (0.030--0.223) & 0.216 & 0.020 & 0.110 & 0.003 & 0.013 & 0.027 & 0.057 & 0.112 & $-0.028$ ($-0.303$ to 0.247) & 1.000 \\
\texttt{Gemma 4 26B} & 0.967 & 0.934 & 0.912 & $+0.055$ ($-0.003$ to 0.116) & 0.528 & 0.341 & 0.083 & 0.017 & 0.041 & 0.046 & 0.042 & 0.104 & $+0.226$ (0.059--0.394) & 0.065 \\
\texttt{Gemini 3.1 Flash-Lite} & 0.959 & 0.928 & 0.826 & $+0.132$ (0.049--0.225) & 0.062 & 0.009 & 0.111 & 0.007 & 0.039 & 0.057 & 0.098 & 0.152 & $+0.233$ (0.103--0.363) & 0.005 \\
\texttt{Qwen3-235B-2507} & 0.917 & 0.890 & 0.885 & $+0.031$ ($-0.035$ to 0.097) & 1.000 & 1.000 & 0.098 & 0.034 & 0.056 & 0.073 & 0.085 & 0.121 & $+0.132$ (0.005--0.260) & 0.288 \\
\texttt{gpt-oss-20b} & 0.900 & 0.768 & 0.719 & $+0.181$ (0.070--0.299) & 0.035 & 0.009 & 0.149 & 0.061 & 0.091 & 0.190 & 0.236 & 0.254 & $+0.211$ (0.121--0.301) & $<$0.001 \\
\bottomrule
\end{tabular}
}
\caption{\textbf{Difficulty level and trace depth.} Left: FAC over the 58 Easy, 58 Intermediate, and
34 Advanced templates. Middle: the Easy mean minus the Advanced mean, with its 95\% interval, the
Holm-adjusted $p$ of Welch's $t$-test and of a permutation of level labels, and the smallest gap the
design detects at 80\% power. Right: the share of instances scored 0, empty responses included, by
the number of milestones in the gold trace (293, 462, 368, 542, and 528 instances). Far right: over
readable responses only (empty responses and responses that state no answer are left out), the slope
of a template-clustered logistic model of a wrong answer on the milestone count, with the answer
kind as a covariate, and its Holm-adjusted $p$ over the eleven models; the slope holds after
correction for three of them.}
\label{tab:level_gap}
\end{table*}

\section{Branch and Domain Performance}
\label{appendix:branch_domain}

\begin{figure}[t]
    \centering
    \includegraphics[width=\columnwidth]{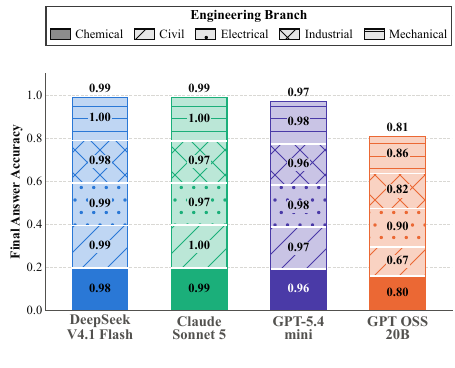}
    \caption{\textbf{Final Answer Accuracy by engineering branch for four representative models.} One
    stacked bar per model, for \texttt{DeepSeek V4.1 Flash} (first on FAC), \texttt{Claude Sonnet 5}
    (first on MC), \texttt{GPT-5.4 mini} (closed, fifth on FAC), and \texttt{gpt-oss-20b} (GPT OSS 20B
    in the figure). Each of the five branches holds 30 of the 150 templates, so a segment is that
    branch's share of the model's FAC: the number inside it is the branch's FAC, and its height is a
    fifth of that. The segments add up to the model's FAC, printed above the bar; color marks the model
    and hatching the branch.}
    \label{fig:branch_bars}
\end{figure}

\begin{figure}[t]
    \centering
    \includegraphics[width=\columnwidth]{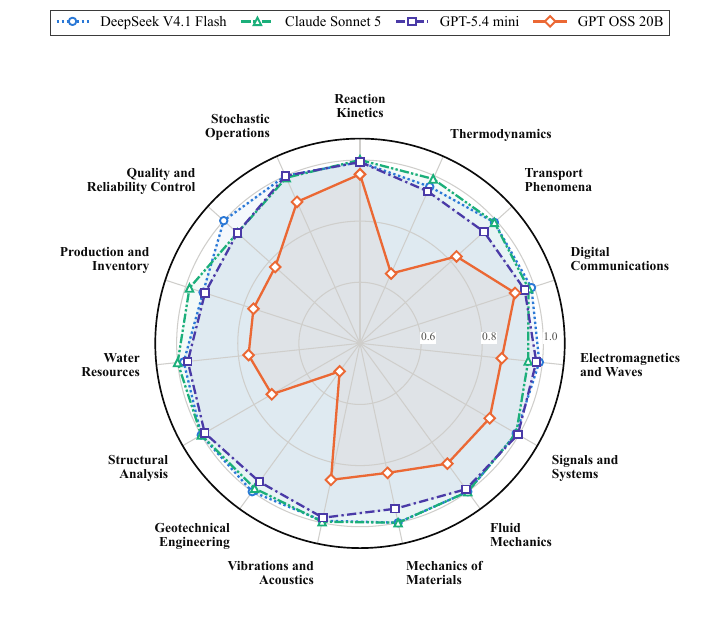}
    \caption{\textbf{Final Answer Accuracy by domain for four representative models.} The instance mean
    over each of the 15 domains, in the order of their branches around the rim, for the same four
    models; the radial axis starts at 0.4. The domain means carry no interval and no test
    (\autoref{tab:by_domain_kind}). In the figure, \texttt{gpt-oss-20b} is labeled GPT OSS 20B.}
    \label{fig:domain_radar}
\end{figure}

The branch means carry no order of difficulty: the lowest branch differs by model, chemical for five
models, industrial for four, and civil for one, with \texttt{DeepSeek V4.1 Flash} tied between
chemical and industrial.
Within a model, a single branch pair differs after correction.
This section gives, for~\autoref{subsec:branch_domain}, every model's branch and domain means
(\autoref{tab:branch_domain}) and draws those of four representative models:
\texttt{DeepSeek V4.1 Flash}, first on FAC and open-weights; \texttt{Claude Sonnet 5}, first on MC
and closed; \texttt{GPT-5.4 mini}, closed and fifth on FAC; and \texttt{gpt-oss-20b}, open-weights
and last on FAC.
In~\autoref{fig:branch_bars}, \texttt{DeepSeek V4.1 Flash}, \texttt{Claude Sonnet 5}, and
\texttt{GPT-5.4 mini} stay close to their FAC in every branch, while \texttt{gpt-oss-20b} varies far
more and is lowest in civil engineering.
In~\autoref{fig:domain_radar}, the same three stay above 0.93 in every domain, each lowest in an
industrial one, while \texttt{gpt-oss-20b} falls to 0.51 in geotechnical engineering.

\begin{table*}[t]
\centering
\footnotesize
\renewcommand{\arraystretch}{1.1}
\adjustbox{max width=\textwidth}{%
\begin{tabular}{l r r r r r r r r r r r }
\toprule
\rowcolor{gray!10}
\textbf{Branch or domain (templates)} & \rotatebox{90}{\texttt{DeepSeek V4.1 Flash}} & \rotatebox{90}{\texttt{Claude Sonnet 5}} & \rotatebox{90}{\texttt{Kimi K3}} & \rotatebox{90}{\texttt{GLM-5.3-Flash}} & \rotatebox{90}{\texttt{GPT-5.4 mini}} & \rotatebox{90}{\texttt{Muse Glimmer 30B}} & \rotatebox{90}{\texttt{GLM-5.3}} & \rotatebox{90}{\texttt{Gemma 4 26B}} & \rotatebox{90}{\texttt{Gemini 3.1 Flash-Lite}} & \rotatebox{90}{\texttt{Qwen3-235B-2507}} & \rotatebox{90}{\texttt{gpt-oss-20b}} \\
\midrule
\rowcolor{gray!10}\textit{Chemical} (30) & 0.980 & 0.993 & 0.960 & 0.960 & 0.961 & 0.963 & 0.900 & 0.896 & 0.846 & 0.869 & 0.798 \\
\quad Reaction kinetics (10) & 0.993 & 1.000 & 0.993 & 1.000 & 0.993 & 1.000 & 0.993 & 0.960 & 0.903 & 0.980 & 0.953 \\
\quad Thermodynamics (12) & 0.961 & 0.989 & \textbf{0.917} & 0.906 & 0.944 & 0.922 & \textbf{0.767} & \textbf{0.806} & \textbf{0.778} & 0.833 & 0.650 \\
\quad Transport phenomena (8) & 0.992 & 0.992 & 0.983 & 0.992 & 0.946 & 0.979 & 0.983 & 0.950 & 0.875 & \textbf{0.783} & 0.825 \\
\rowcolor{gray!10}\textit{Civil} (30) & 0.993 & 0.996 & 0.982 & 0.982 & 0.971 & 0.962 & 0.973 & 0.960 & 0.918 & 0.940 & 0.671$^{\P}$ \\
\quad Geotechnical engineering (10) & 1.000 & 0.987 & 1.000 & 0.993 & 0.960 & 0.927 & 0.993 & 0.953 & 0.973 & 0.947 & \textbf{0.513} \\
\quad Structural analysis (10) & 1.000 & 1.000 & 0.993 & 1.000 & 0.987 & 0.987 & 1.000 & 0.973 & 0.987 & 0.947 & 0.733 \\
\quad Water resources (10) & 0.980 & 1.000 & 0.953 & 0.953 & 0.967 & 0.973 & 0.927 & 0.953 & 0.793 & 0.927 & 0.767 \\
\rowcolor{gray!10}\textit{Electrical} (30) & 0.990 & 0.974 & 0.984 & 0.981 & 0.981 & 0.989 & 0.961 & 0.960 & 0.947 & 0.911 & 0.897 \\
\quad Digital communications (10) & 0.990 & 0.980 & 0.987 & 0.980 & 0.967 & 1.000 & 0.947 & 0.970 & 0.947 & 0.897 & 0.933 \\
\quad Electromagnetics and waves (10) & 0.990 & 0.953 & 0.977 & 0.977 & 0.980 & 0.987 & 0.943 & 0.910 & 0.927 & 0.890 & 0.867 \\
\quad Signals and systems (10) & 0.990 & 0.990 & 0.990 & 0.987 & 0.997 & 0.980 & 0.993 & 1.000 & 0.967 & 0.947 & 0.890 \\
\rowcolor{gray!10}\textit{Industrial} (30) & 0.980 & 0.973 & 0.980 & 0.942 & 0.958 & 0.938 & 0.918 & 0.920 & 0.893 & 0.871 & 0.816 \\
\quad Production and inventory (10) & \textbf{0.940} & 0.987 & 0.967 & \textbf{0.887} & \textbf{0.933} & \textbf{0.880} & 0.880 & 0.907 & 0.920 & 0.847 & 0.767 \\
\quad Quality and reliability control (10) & 1.000 & \textbf{0.940} & 0.980 & 0.947 & 0.940 & 0.933 & 0.907 & 0.860 & 0.813 & 0.787 & 0.773 \\
\quad Stochastic operations (10) & 1.000 & 0.993 & 0.993 & 0.993 & 1.000 & 1.000 & 0.967 & 0.993 & 0.947 & 0.980 & 0.907 \\
\rowcolor{gray!10}\textit{Mechanical} (30) & 0.998 & 0.999 & 0.996 & 0.996 & 0.976 & 0.993 & 0.998 & 0.974 & 0.980 & 0.904 & 0.859 \\
\quad Fluid mechanics (10) & 1.000 & 1.000 & 1.000 & 0.997 & 0.990 & 0.997 & 1.000 & 0.973 & 0.977 & 0.913 & 0.887 \\
\quad Mechanics of materials (10) & 1.000 & 1.000 & 0.993 & 0.993 & 0.953 & 0.997 & 1.000 & 0.963 & 0.977 & 0.943 & 0.833 \\
\quad Vibrations and acoustics (10) & 0.993 & 0.997 & 0.993 & 0.997 & 0.983 & 0.987 & 0.993 & 0.987 & 0.987 & 0.857 & 0.857 \\
\midrule
Detectable branch difference & 0.037 & 0.030 & 0.055 & 0.080 & 0.049 & 0.081 & 0.123 & 0.107 & 0.136 & 0.129 & 0.188 \\
\bottomrule
\end{tabular}
}
\caption{\textbf{Final Answer Accuracy by branch and by domain.} Models in order of FAC. Shaded
rows: the mean of the branch's 30 template means; below each, its domains' instance means, which
carry no interval or test; template counts in parentheses. Bold: each model's lowest domain.
$^{\P}$The one branch pair that differs within a model after Holm correction (Welch's $t$-test over
the model's ten pairs): \texttt{gpt-oss-20b}'s electrical above its civil. Last row: the smallest
branch difference 30 templates detect at 80\% power, per model.}
\label{tab:branch_domain}
\label{tab:by_branch}
\label{tab:by_domain_kind}
\end{table*}

\section{Paraphrase Test}
\label{appendix:paraphrase}

Because every instance of a template shares its wording, a model could recognize a problem from the
wording alone and reproduce a method it has memorized for it.
This section gives the design and the results of the paraphrase test
of~\autoref{subsec:overall_performance}.

\paragraph{Design.} We select three of every template's 15 instances (450 instances) and paraphrase
each once with \texttt{Mistral Large 3}~\citep{mistral2025mistral3} (the writer), an LLM from a
family outside both the evaluated models and the judge, keeping every number, unit, symbol,
formula, acronym, and labeled part in place.
A paraphrase must pass two checks:
\begin{itemize}[nosep, leftmargin=*]
\item \emph{Scripted:} the numbers as a multiset, every technical token, the part labels, a word
similarity to the original of at most 0.75, the length, and no preamble, with three attempts per
instance and the original's notation restored where the writer used symbols the original lacks.
\item \emph{Expert:} a domain expert of the instance's branch keeps a pair only if it is the same
problem with the same answer and adds nothing.
\end{itemize}
Of the 450 instances, 314 pass the scripted checks and the domain experts keep 275 pairs on 114
templates; the templates left out are mostly formula-bound ones whose technical tokens the writer
does not keep in place.
The expert step is needed: the experts reject 39 pairs that pass every scripted check, mostly for
a dropped or altered qualifier that changes the physics or the asked quantity.
The eleven models answer the kept paraphrases under the evaluation's prompt, settings, and scoring.

\paragraph{Results.} On the 275 kept pairs, no change in FAC holds after correction.
Read against the $\pm 0.05$ margin, the 90\% interval lies within it for every model but
\texttt{Qwen3-235B-2507} and \texttt{gpt-oss-20b}, the two models whose paired differences the
design measures least precisely: the smallest difference it detects at 80\% power is 0.052 and
0.079 for them and at most 0.044 for the others (\autoref{tab:paraphrase}).
MC changes by at most 0.023 in either direction, so the derivation is at least as robust to the
wording as the answer.
The paraphrases reorder the models within the range sampling noise produces, though at its high end:
their ordering agrees with the originals' at Kendall's $\tau$ 0.722 (95\% interval 0.449 to 0.844),
while two disjoint draws of 275 original instances with the same template mix agree with a median
$\tau$ of 0.807 (quartiles 0.748 to 0.860; 5th percentile 0.647).
For the same two models, the paraphrase change also exceeds decoding noise: of the four models
decoded three more times on 300 instances (\texttt{Gemma 4 26B}, \texttt{Gemini 3.1 Flash-Lite},
\texttt{Qwen3-235B-2507}, and \texttt{gpt-oss-20b}), FAC varies by a standard deviation of 0.003 to
0.008 across repeats, and the paraphrase change lies beyond the spread of the repeats' changes for
\texttt{Qwen3-235B-2507} and \texttt{gpt-oss-20b} and within it for the other two.
The test bounds memorized wording, not memorized methods, since the numbers and the problem
structure are kept, and it covers only the 114 templates that can be paraphrased without loss.

\begin{table*}[t]
\centering
\footnotesize
\renewcommand{\arraystretch}{1.1}
\begin{tabular}{l c c r r}
\toprule
\rowcolor{gray!10}
\textbf{Model} & \textbf{FAC change (95\% interval)} & \textbf{90\% interval} & \textbf{$p$ (Holm)} & \textbf{MC change} \\
\midrule
\texttt{DeepSeek V4.1 Flash} & $-0.013$ ($-0.029$ to 0.000) & $-0.026$ to $-0.002$ & 1.000 & $-0.009$ \\
\texttt{Claude Sonnet 5} & $-0.011$ ($-0.025$ to 0.000) & $-0.022$ to $-0.003$ & 1.000 & $-0.005$ \\
\texttt{Kimi K3} & $+0.002$ ($-0.005$ to 0.011) & $-0.004$ to 0.009 & 1.000 & $-0.005$ \\
\texttt{GLM-5.3-Flash} & $-0.011$ ($-0.029$ to 0.005) & $-0.026$ to 0.004 & 1.000 & $-0.015$ \\
\texttt{GPT-5.4 mini} & $+0.005$ ($-0.012$ to 0.024) & $-0.009$ to 0.021 & 1.000 & $+0.023$ \\
\texttt{Muse Glimmer 30B} & $+0.025$ (0.004--0.051) & 0.007--0.046 & 0.527 & $-0.007$ \\
\texttt{GLM-5.3} & $+0.002$ ($-0.018$ to 0.022) & $-0.015$ to 0.019 & 1.000 & $+0.001$ \\
\texttt{Gemma 4 26B} & $0.000$ ($-0.033$ to 0.028) & $-0.026$ to 0.023 & 1.000 & $-0.009$ \\
\texttt{Gemini 3.1 Flash-Lite} & $-0.013$ ($-0.031$ to 0.004) & $-0.028$ to 0.000 & 1.000 & $-0.006$ \\
\texttt{Qwen3-235B-2507} & $-0.036$ ($-0.080$ to 0.005) & $-0.073$ to $-0.002$ & 1.000 & $-0.015$ \\
\texttt{gpt-oss-20b} & $+0.018$ ($-0.031$ to 0.067) & $-0.023$ to 0.059 & 1.000 & $-0.021$ \\
\bottomrule
\end{tabular}
\caption{\textbf{Change under paraphrase.} Paraphrase minus original on the 275 expert-kept pairs
over 114 templates, positive when the paraphrase scores higher: the change in FAC with its 95\%
interval, its 90\% interval read against the $\pm 0.05$ margin, the Holm-adjusted $p$ of the
sign-flip test over templates, and the change in MC.}
\label{tab:paraphrase}
\end{table*}

\section{Further Experiments: Reasoning Effort, Flagship Anchors, Open Book, and Open Tool}
\label{appendix:experiments}

Four experiments extend the evaluation, each answering one question it leaves open, and
\autoref{subsec:reasoning_equations_tools} reads them.
Each runs on the originals of the 450-instance subset (three instances per template) with the
evaluation's prompt, output ceiling, and scoring, and changes one thing.
Where a model answers both with and without the change, we pair its two responses to each instance
(\autoref{tab:experiments}) and read the 90\% interval against the $\pm 0.05$ margin, beside a
sign-flip test over templates, Holm-corrected within the experiment, and the smallest change the
experiment detects at 80\% power.
Apart from that one change, the experiments run at the providers' defaults, as do the responses
they are paired with.

\paragraph{Reasoning Effort.} Do the two closed models that return no reasoning tokens at their
providers' defaults, \texttt{GPT-5.4 mini} and \texttt{Gemini 3.1 Flash-Lite}, score differently
when they reason?
\texttt{GPT-5.4 mini} does, in its derivation as well as its answer, and
\texttt{Gemini 3.1 Flash-Lite} changes by less than the experiment detects.
Each answers the subset with the serving API's reasoning-effort parameter set to medium and nothing
else changed.
The setting lifts \texttt{GPT-5.4 mini}'s FAC beyond the margin, raises its MC by 0.057 (95\%
interval 0.031 to 0.085), and lowers its arithmetic flags on correct answers; its largest point
change in FAC is on Advanced templates, though the intervals of the three levels overlap.
Over all 2,250 instances, the matched settings (\autoref{appendix:decoding}) give the same
comparison for these two models and \texttt{Gemma 4 26B} (\autoref{tab:matched}).
There, the FAC gain of \texttt{Gemini 3.1 Flash-Lite} holds, though it is smaller than the subset
experiment detects.
The derivation changes chiefly for \texttt{GPT-5.4 mini}, whose MC rises by about as much as on the
subset, while that of the other two moves by at most 0.016.

\paragraph{Flagship Anchors.} Does a flagship exceed the first five, or is the ceiling an artifact
of evaluating none?
The flagships do not score clearly above the first five.
\texttt{GPT-5.4} and \texttt{DeepSeek V4 Pro}, the two flagships that pass the selection rule
(\autoref{appendix:models}), answer the same subset at their providers' defaults, and
\texttt{GPT-5.4}, whose default returns no reasoning tokens, also with reasoning at medium effort;
an anchor is a reference point outside the pairwise tests, so no test compares it with the
evaluated models.
\texttt{DeepSeek V4 Pro} scores 0.968 and \texttt{GPT-5.4} 0.951 at its default and 0.997 with
reasoning, against 0.967 to 0.990 for the first five on the same instances; the highest anchor
run lies inside the interval of \texttt{DeepSeek V4.1 Flash} on these instances, and on Advanced
templates the anchors score 0.89 to 1.00 and the first five 0.94 to 0.99.
Reasoning lifts a flagship as it lifts \texttt{GPT-5.4 mini}: with it, \texttt{GPT-5.4} gains over
its own default and sheds most of its arithmetic flags (\autoref{tab:experiments}).

\paragraph{Open Book.} Does recall of the governing equations limit the scores?
Recall is part of what limits the weakest model, \texttt{gpt-oss-20b}, and supplying the equations
leaves \texttt{Claude Sonnet 5} and \texttt{GPT-5.4 mini} within the margin.
For the 135 templates whose code states their governing equations, the question carries them as a
handbook block, one per line, with no data tables or constants (405 instances); three models
answer: \texttt{Claude Sonnet 5} from the first five, and the closed \texttt{GPT-5.4 mini} and the
open-weights \texttt{gpt-oss-20b}, which score lower at the providers' defaults.
The equations lift \texttt{gpt-oss-20b}, partly because fewer of its responses run out of room (the
change on the 388 instances it answered in both runs is $+0.044$), and MC rises for all three,
consistent with the equations prompting the responses to state the intermediates.
The experiment supplies equations and no data tables, so it bounds what formula recall adds and not
what a property lookup would.

\paragraph{Open Tool.} Does arithmetic execution limit the scores?
An offered interpreter moves neither \texttt{Claude Sonnet 5} nor \texttt{GPT-5.4 mini} beyond the
margin in FAC or MC, though \texttt{GPT-5.4 mini}'s arithmetic flags on correct answers fall.
The model may call a Python interpreter with NumPy, SciPy, and SymPy up to 8 times before one final
turn without it; each script runs isolated, with no credentials, a 20 s limit, and its output
truncated at 4,000 characters, behind a filter that refuses file-system, network, process, and
introspection access.
Only these two closed models answer, since no endpoint serving \texttt{gpt-oss-20b} at 8-bit
precision or higher handles its tool calls reliably.
Whether to call the interpreter is the model's choice, \texttt{Claude Sonnet 5} on 66\% of the
instances and \texttt{GPT-5.4 mini} on 18\%, so the experiment bounds what an offered tool adds, not
what required use would.

\begin{table*}[t]
\centering
\footnotesize
\renewcommand{\arraystretch}{1.1}
\adjustbox{max width=\textwidth}{%
\begin{tabular}{l r r r c r r c r c}
\toprule
\rowcolor{gray!10}
\textbf{Model} & \textbf{Instances} & \textbf{\makecell{FAC,\\evaluation}} & \textbf{\makecell{FAC,\\experiment}} & \textbf{\makecell{Change\\(95\% interval)}} & \textbf{$p$ (Holm)} & \textbf{Detectable} & \textbf{90\% interval} & \textbf{\makecell{MC\\change}} & \textbf{\makecell{Arithmetic flags,\\evaluation / experiment}} \\
\midrule
\rowcolor{gray!10}\multicolumn{10}{l}{\textit{Reasoning at medium effort}} \\
\texttt{GPT-5.4 mini} & 450 & 0.862 & 0.964 & $+0.102$ (0.063--0.143) & $<$0.001 & 0.058 & 0.070--0.137 & $+0.057$ & 0.125 / 0.067 \\
\texttt{Gemini 3.1 Flash-Lite} & 450 & 0.884 & 0.898 & $+0.013$ ($-0.014$ to 0.042) & 0.407 & 0.040 & $-0.010$ to 0.038 & $+0.017$ & 0.086 / 0.075 \\
\rowcolor{gray!10}\multicolumn{10}{l}{\textit{Open book: governing equations supplied}} \\
\texttt{gpt-oss-20b} & 405 & 0.810 & 0.878 & $+0.068$ (0.031--0.106) & 0.001 & 0.054 & 0.037--0.100 & $+0.068$ & 0.165 / 0.179 \\
\texttt{GPT-5.4 mini} & 405 & 0.853 & 0.837 & $-0.016$ ($-0.047$ to 0.016) & 0.629 & 0.044 & $-0.042$ to 0.010 & $+0.032$ & 0.123 / 0.161 \\
\texttt{Claude Sonnet 5} & 405 & 0.980 & 0.989 & $+0.009$ ($-0.002$ to 0.021) & 0.629 & 0.017 & 0.000--0.020 & $+0.028$ & 0.146 / 0.133 \\
\rowcolor{gray!10}\multicolumn{10}{l}{\textit{Open tool: Python interpreter offered}} \\
\texttt{GPT-5.4 mini} & 450 & 0.862 & 0.859 & $-0.003$ ($-0.040$ to 0.036) & 0.909 & 0.054 & $-0.034$ to 0.029 & $+0.010$ & 0.125 / 0.094 \\
\texttt{Claude Sonnet 5} & 450 & 0.982 & 0.992 & $+0.010$ (0.000--0.023) & 0.252 & 0.017 & 0.001--0.021 & $+0.002$ & 0.136 / 0.128 \\
\rowcolor{gray!10}\multicolumn{10}{l}{\textit{\texttt{GPT-5.4} with reasoning at medium effort, against its default}} \\
\texttt{GPT-5.4} & 450 & 0.951 & 0.997 & $+0.046$ (0.021--0.074) & $<$0.001 & 0.039 & 0.024--0.069 & $+0.012$ & 0.117 / 0.027 \\
\bottomrule
\end{tabular}
}
\caption{\textbf{The four experiments against the evaluation.} Per model and experiment: the
instances covered; FAC in the evaluation and under the experiment on those instances, paired by
instance; the change with its 95\% interval, the Holm-adjusted $p$ of the sign-flip test over
templates, the smallest change the experiment detects at 80\% power, and the 90\% interval read
against the $\pm 0.05$ margin; the change in MC; and the share of correct-answer responses with an
arithmetic flag in the evaluation and under the experiment (unpaired). The experiments and their
evaluation column are at the providers' default settings on the same instances (for
\texttt{GPT-5.4 mini} and \texttt{Gemini 3.1 Flash-Lite} without reasoning); the full-set change
with reasoning on is in \autoref{tab:matched}.}
\label{tab:experiments}
\end{table*}

\section{Error Analysis}
\label{appendix:error_analysis}

What kind of error produces a wrong answer, and are the first five's remaining incorrect verdicts
errors at all?
This section gives the taxonomy, the sample, and the readings
behind~\autoref{subsec:error_analysis}, then the near misses and the incorrect verdicts that
concern the answer's form rather than its value.

\paragraph{Taxonomy.} The six categories follow the stages of an engineering derivation, from the
most to the least fundamental:
\begin{enumerate}[nosep, leftmargin=*]
\item \emph{Hallucination:} a value, constant, or equation that exists in no engineering reference.
\item \emph{Setup or assumption:} the problem is framed wrongly before any equation applies, such as
a wrong boundary condition or an ignored constraint.
\item \emph{Formula or principle:} the setup is right, but the governing equation or principle is
wrong.
\item \emph{Unit or dimension:} a missing conversion, mixed unit systems, or a value in the wrong
unit.
\item \emph{Sign or direction:} the wrong sign or physical direction for a quantity.
\item \emph{Calculation:} every earlier stage is right, and the arithmetic or algebra is wrong.
\end{enumerate}
A domain expert answers the six questions in this order and stops at the first yes, so the label is
the most fundamental error and not a downstream symptom; two further options are no error (the
answer is correct, or the question admits it) and incomplete.

\paragraph{Sample and Readings.} We read 137 wrong answers from four models that span the FAC order
(\texttt{Claude Sonnet 5}, \texttt{GPT-5.4 mini}, \texttt{Gemma 4 26B}, and \texttt{gpt-oss-20b}),
each a readable response at the providers' default settings whose final answer the check scores
incorrect; we draw up to 40 per model at random, across levels in proportion to the model's wrong
answers and spread over its templates, 77 templates in all.
Three domain experts of the wrong answer's branch, from the fifteen who certified the templates,
read each full response without the model's name.
Unit and dimension errors are nearly absent and sign errors rare, in two models only; every
hallucination reading is \texttt{gpt-oss-20b}'s; and agreement, high for every model, is lowest for
\texttt{gpt-oss-20b}, whose errors spread across the conceptual categories (\autoref{tab:errors}).
The categories describe sampled wrong answers, not rates over all responses, and they describe the
responses at the providers' default settings, not those of the three models at matched settings.

\begin{table*}[t]
\centering
\footnotesize
\setlength{\tabcolsep}{4pt}
\renewcommand{\arraystretch}{1.1}
\adjustbox{max width=\textwidth}{%
\begin{tabular}{l r r r r r r r }
\toprule
\rowcolor{gray!10}
 & \multicolumn{4}{c}{\textbf{By model: readings (majority labels)}} & \multicolumn{3}{c}{\textbf{By level: readings (share / share without no-error readings)}} \\
\cmidrule(lr){2-5}\cmidrule(lr){6-8}
\textbf{Category} & \texttt{Claude Sonnet 5} & \texttt{GPT-5.4 mini} & \texttt{Gemma 4 26B} & \texttt{gpt-oss-20b} & \textbf{Easy} & \textbf{Intermediate} & \textbf{Advanced} \\
\midrule
Hallucination & 0 (0) & 0 (0) & 0 (0) & 7 (3) & 2 (3\% / 3\%) & 5 (3\% / 3\%) & 0 (0\% / 0\%) \\
Setup or assumption & 9 (3) & 5 (2) & 7 (2) & 16 (6) & 2 (3\% / 3\%) & 26 (13\% / 14\%) & 9 (7\% / 7\%) \\
Formula or principle & 6 (2) & 16 (5) & 20 (7) & 31 (9) & 2 (3\% / 3\%) & 33 (17\% / 18\%) & 38 (28\% / 29\%) \\
Unit & 0 (0) & 0 (0) & 0 (0) & 1 (0) & 0 (0\% / 0\%) & 0 (0\% / 0\%) & 1 (1\% / 1\%) \\
Sign or direction & 0 (0) & 5 (2) & 6 (2) & 0 (0) & 0 (0\% / 0\%) & 6 (3\% / 3\%) & 5 (4\% / 4\%) \\
Calculation & 36 (12) & 88 (29) & 84 (28) & 53 (18) & 69 (88\% / 92\%) & 113 (58\% / 62\%) & 79 (57\% / 60\%) \\
No error & 6 (2) & 3 (1) & 0 (0) & 12 (4) & 3 (4\% / --) & 12 (6\% / --) & 6 (4\% / --) \\
\midrule
Wrong answers in the sample, by level & 1, 13, 5 & 8, 17, 14 & 9, 16, 14 & 8, 19, 13 & 26 & 65 & 46 \\
Readings & 57 & 117 & 117 & 120 & 78 & 195 & 138 \\
Fleiss' $\kappa$ & 1.000 & 0.917 & 0.962 & 0.835 & & &  \\
\bottomrule
\end{tabular}
}
\caption{\textbf{The domain experts' readings of the wrong answers.} The sample holds 137 wrong
answers of four models (19 from \texttt{Claude Sonnet 5}, 39 from \texttt{GPT-5.4 mini}, 39 from
\texttt{Gemma 4 26B}, and 40 from \texttt{gpt-oss-20b}); three domain experts of the answer's branch
read each (411 readings). Left: readings per category and, in parentheses, the wrong answers whose
majority label is the category. Right: readings per category by level, with their share of the
level's readings and their share once the no-error readings are removed, so the error mix can be
read apart from the share of answers the experts found correct. Below: the sample by level, the
readings per column, and Fleiss' $\kappa$ over the three experts per model (0.912 overall).
Categories run from the most to the least fundamental; no domain expert used the incomplete option.}
\label{tab:errors}
\label{tab:error_by_level}
\end{table*}

\paragraph{Near Misses.} On each of six templates where many wrong answers land between 0.2\% and
5\% of the gold value, the near miss shown to the domain experts is the model's error, not the
wording's.
Three domain experts of each template's branch read these templates
(\texttt{annulus\_flowrate}, \texttt{best\_hydraulic\_rectangular\_section},
\texttt{manning\_rectangular\_discharge}, \texttt{pfr\_volume\_changing\_rate},
\texttt{pitzer\_correlation\_z}, and \texttt{vdw\_solve\_for\_volume}), and for each, all three
answer that the question has one correct answer and that the response shown, within 5\% of the
gold value, is not correct.

\paragraph{Incorrect Verdicts and the Answer's Form.} About a quarter of the first five's incorrect
verdicts fall on three templates, and most of these lie within 0.2\% of a target whose digits the
question prescribes: of the 126, 18 fall on \texttt{qr\_policy\_one\_iteration}, 9 on
\texttt{single\_sampling\_oc\_point}, and 7 on \texttt{adiabatic\_flame\_temperature}; on the first
two, 24 of the 27 lie that close.
Two sources of incorrect verdicts concern the answer's form rather than its value.
First, 17 templates prescribe the digits of the answer, so a value within the tolerance but not at
the prescribed digits is incorrect by the question's own terms (240 incorrect verdicts across the
eleven models, 108 of them within 0.2\% of the target on 11 of these templates).
Second, four of the nine templates with symbolic answers are scored by the numbers they state, while
the other five also accept an equivalent expression (\autoref{appendix:scoring}); the first five's
symbolic incorrect verdicts all fall on one of these five, \texttt{ber\_estimation\_mary}.
Changing one clause of the rule at a time (the absolute-value clause, the last-digit term, the
prescribed digits, or proportional partial credit) moves a model's FAC by at most 0.030 and keeps
Kendall's $\tau$ with the ordering as scored at 0.782 or above (\autoref{tab:scoring_variants}).
Four templates whose answers can be read off the question's wording stay in the evaluation set,
since leaving them out moves FAC by at most 0.002: \texttt{levenspiel\_plot\_interpretation},
\texttt{line\_balancing\_heuristic}, \texttt{system\_properties\_memory\_causality}, and
\texttt{system\_property\_linearity}.

\end{document}